\RequirePackage[svgnames,table]{xcolor}
\PassOptionsToPackage{authoryear,round}{natbib}
\documentclass[10pt,letterpaper]{mystyle}
\usepackage[utf8]{inputenc}
\usepackage{natbib}
\setcitestyle{authoryear,round,citesep={;},aysep={,},yysep={;}}
\usepackage{url}
\usepackage{graphicx}
\usepackage{booktabs}
\usepackage{array}
\usepackage{amsmath}
\usepackage{amssymb}
\usepackage{xcolor}
\usepackage{placeins}
\usepackage{float}
\usepackage{listings}
\ifdefined\XeTeXversion
\newcommand{\aedpromptfont}{\font\aedpromptmono="[lmmono10-regular.otf]" at 7pt\aedpromptmono}
\else
\newcommand{\aedpromptfont}{\ttfamily\scriptsize}
\fi
\usepackage{titletoc}
\titlecontents{section}[1.8em]
  {\vspace{2pt}}
  {\contentslabel{1.8em}}
  {\hspace*{-1.8em}}
  {\titlerule*[0.5pc]{.}\contentspage}
\titlecontents{subsection}[4.3em]
  {}
  {\contentslabel{2.5em}}
  {\hspace*{-2.5em}}
  {\titlerule*[0.5pc]{.}\contentspage}
\usepackage{tikz}
\usepackage{pgfplots}
\pgfplotsset{compat=1.17}
\definecolor{phink}{RGB}{176,88,0}
\definecolor{phbg}{RGB}{255,244,229}

\usepackage[most]{tcolorbox}
\definecolor{rqink}{RGB}{28,74,112}
\definecolor{rqtint}{RGB}{236,242,247}
\definecolor{tkink}{RGB}{120,78,10}
\definecolor{tktint}{RGB}{252,246,234}
\newcommand{\rqlabel}[1]{\textbf{\textcolor{rqink}{#1.}}}
\newtcolorbox{rqbox}{enhanced,colback=rqtint,colframe=rqink,boxrule=0pt,leftrule=2.2pt,arc=0pt,
  left=6pt,right=6pt,top=3pt,bottom=3pt,before skip=5pt,after skip=5pt}
\newtcolorbox{rqblock}{enhanced,colback=rqtint,colframe=rqtint,boxrule=0pt,arc=0pt,left=4pt,right=4pt,
  top=2pt,bottom=2pt,before skip=4pt,after skip=4pt,fontupper=\itshape}
\newcommand{\researchq}[2]{\begin{rqblock}\rqlabel{#1}~#2\end{rqblock}}
\newtcolorbox{takeaway}{enhanced,colback=tktint,colframe=tkink,boxrule=0pt,leftrule=2.2pt,arc=0pt,
  left=6pt,right=6pt,top=2pt,bottom=2pt,before skip=5pt,after skip=4pt,fontupper=\small}

\newcommand{\cmark}{\ensuremath{\checkmark}}
\newcommand{\xmark}{\ensuremath{\times}}
\newcommand{\pmark}{\ensuremath{\circ}}

\newcommand{\dataset}{\textsc{AED}}

\newcommand{\numrows}{50,000}
\title{Agent Error Dataset: Scaling \numrows{} Error--Diagnosis Pairs for Failure Analysis and Error-Aware Post-Training}

\author{Kunlun Zhu, Xuyan Ye, Yibo Li, Cheng Qian, Beibin Li, Heng Ji\\
{\small Apodex}}

\runningtitle{Agent Error Dataset}

\input{title_branding}
\definecolor{aedblue}{RGB}{37,79,119}
\definecolor{aedteal}{RGB}{35,110,105}
\definecolor{aedamber}{RGB}{142,94,32}
\definecolor{aedcoral}{RGB}{164,66,57}
\definecolor{aedpurple}{RGB}{100,78,143}
\newtcolorbox{recoveryphase}[2]{enhanced,
  colback=#1!3!white,colframe=#1!50!white,colbacktitle=#1!10!white,
  coltitle=#1!90!black,title={#2},fonttitle=\small\bfseries,
  fontupper=\small\raggedright,boxrule=.4pt,leftrule=2.2pt,arc=1.1mm,
  left=7pt,right=7pt,top=4pt,bottom=4pt,before skip=5pt,after skip=5pt}
\newtcolorbox{artifactbox}[2][aedblue]{enhanced,breakable,
  colback=#1!3!white,colframe=#1!55!white,colbacktitle=#1!9!white,
  coltitle=#1!90!black,title={#2},fonttitle=\small\bfseries,
  fontupper=\small\raggedright,boxrule=.45pt,leftrule=2pt,arc=1.2mm,
  left=7pt,right=7pt,top=5pt,bottom=5pt,
  before skip=8pt,after skip=8pt,pad at break*=2mm}
\newcommand{\casefield}[1]{\par\smallskip\textbf{#1}\enspace}
\newcolumntype{P}[1]{>{\raggedright\arraybackslash}p{#1}}
\ifdefined\XeTeXversion
\renewcommand{\aedpromptfont}{\font\aedpromptmono="[lmmono10-regular.otf]" at 8pt\aedpromptmono}
\else
\renewcommand{\aedpromptfont}{\ttfamily\footnotesize}
\fi
\makeatletter
\let\aedplainclassz\@classz
\@ifundefined{insert@pcolumn}{\let\insert@pcolumn\insert@column}{}
\makeatother
\usepackage{colortbl}
\makeatletter
\let\aedcolorclassz\@classz
\let\@classz\aedplainclassz
\newcommand{\aedactivatecolumns}{\let\@classz\aedcolorclassz}
\makeatother
\definecolor{aedtablehead}{HTML}{E7F0F7}
\definecolor{aedtablerow}{HTML}{F3F7FA}
\definecolor{aedtablerule}{HTML}{7894A9}
\definecolor{aedtablelabel}{HTML}{315B82}
\definecolor{aedfigurelabel}{HTML}{287D7A}
\newif\ifaedsupplement
\newif\ifaedtableactive
\apptocmd{\appendix}{\global\aedsupplementtrue\aedactivatecolumns}{}%
  {\PackageError{aed-appendix}{Cannot activate supplement styling}{Check the appendix command.}}
\AtBeginEnvironment{table}{\ifaedsupplement
  \aedtableactivetrue\arrayrulecolor{aedtablerule}\fi}
\makeatletter
\AtEndEnvironment{table}{\ifaedsupplement
  \global\@rowcolorsfalse\arrayrulecolor{black}\fi}
\newcommand{\aedtabularcolors}{\ifaedtableactive
  \rowcolors{1}{aedtablerow}{white}%
  \def\@oddrowcolor{\ifnum\rownum=1
    \gdef\CT@row@color{\CT@color{aedtablehead}}\else
    \gdef\CT@row@color{\CT@color{aedtablerow}}\fi}\fi}
\AtBeginEnvironment{tabular}{\aedtabularcolors}
\AtBeginEnvironment{tabular*}{\ifaedsupplement\global\@rowcolorsfalse\fi}
\let\aedoriginaltablelabel\fnum@table
\let\aedoriginalfigurelabel\fnum@figure
\renewcommand{\fnum@table}{\ifaedsupplement
  \textcolor{aedtablelabel}{\aedoriginaltablelabel}\else\aedoriginaltablelabel\fi}
\renewcommand{\fnum@figure}{\ifaedsupplement
  \textcolor{aedfigurelabel}{\aedoriginalfigurelabel}\else\aedoriginalfigurelabel\fi}
\makeatother

\definecolor{aedreaderblue}{HTML}{254F77}
\definecolor{aedreaderviolet}{HTML}{62419A}
\definecolor{aedreadercyan}{HTML}{087FA6}
\definecolor{aedstatink}{HTML}{116B78}
\DeclareRobustCommand{\aedstat}[1]{{\color{aedstatink}\bfseries #1}}
\definecolor{aedscaleink}{HTML}{315B82}
\definecolor{aedlearnink}{HTML}{705586}
\DeclareRobustCommand{\aedscale}[1]{{\color{aedscaleink}\bfseries #1}}
\DeclareRobustCommand{\aedlearn}[1]{{\color{aedlearnink}\bfseries #1}}
\definecolor{aedmarkyes}{HTML}{087F73}
\definecolor{aedmarkno}{HTML}{BD5444}
\definecolor{aedmarkpartial}{HTML}{A87312}
\renewcommand{\cmark}{\textcolor{aedmarkyes}{\ensuremath{\checkmark}}}
\renewcommand{\xmark}{\textcolor{aedmarkno}{\ensuremath{\times}}}
\renewcommand{\pmark}{\textcolor{aedmarkpartial}{\ensuremath{\circ}}}
\hypersetup{colorlinks=true,linkcolor=aedreaderblue,
  citecolor=aedreaderblue,urlcolor=aedreaderblue}

\renewtcolorbox{rqbox}{enhanced,colback=aedreadercyan!4!white,
  colframe=aedreadercyan!70!black,boxrule=0pt,leftrule=2.2pt,arc=0pt,
  left=6pt,right=6pt,top=3pt,bottom=3pt,before skip=5pt,after skip=5pt}

\titleformat{\paragraph}[runin]
  {\normalsize\bfseries\color{aedreaderblue}}{}{0pt}{#1}

\definecolor{resulttint}{RGB}{237,242,247}
\definecolor{resultink}{HTML}{254F77}
\newcommand{\resultsetup}{\renewcommand{\arraystretch}{1.14}\setlength{\tabcolsep}{3pt}}
\newcommand{\resultgroup}[2]{\multicolumn{#1}{@{}l@{}}{\colorbox{resulttint}{%
  \makebox[\dimexpr\linewidth-2\fboxsep\relax][l]{\strut\textcolor{resultink}{\textbf{#2}}}}}\\}

\newtcolorbox{resulttakeaway}[1]{enhanced,colback=aedreaderviolet!2!white,
  colframe=aedreaderviolet!65!white,colbacktitle=aedreaderviolet!6!white,
  coltitle=aedreaderviolet,title={Takeaway: #1},
  fonttitle=\bfseries\small,fontupper=\small,boxrule=.45pt,leftrule=2pt,
  arc=1pt,left=6pt,right=6pt,top=3pt,bottom=3pt,toptitle=3pt,bottomtitle=0pt,
  before skip=6pt,after skip=6pt}

\newcommand{\humanPreferredStepAgree}{85.5}
\newcommand{\humanPreferredStepPairs}{69}
\newcommand{\humanStepAgree}{92.0}
\newcommand{\humanStepPairs}{50}

\definecolor{aedprojectblue}{HTML}{087FA6}
\hypersetup{
  pdftitle={Agent Error Dataset: Scaling 50,000 Error-Diagnosis Pairs for Failure Analysis and Error-Aware Post-Training},
  pdfauthor={Kunlun Zhu, Xuyan Ye, Yibo Li, Beibin Li, Heng Ji},
  pdfsubject={Agent failure analysis and error-aware post-training},
  pdfkeywords={agent failures, diagnosis, post-training, AED}}
\ifdefined\XeTeXversion
\fi
\makeatletter
\renewcommand{\maketitle}{%
  \par\vspace*{-12mm}
  \begin{tcolorbox}[titlebox,left=5mm,right=5mm,top=5mm,bottom=4mm,
    before skip=0pt,after skip=10pt]
    {\centering\fontsize{21}{24}\selectfont\@title\par}
    \vspace{9pt}
    {\centering\@author\par}
    \vspace{8pt}
    {\absfont\theabstract\par}
    \vspace{8pt}
    \noindent\begin{minipage}[c]{.79\linewidth}
      \small Code and data will be released at
      \href{https://agenterrordata.apodex.com}{\textcolor{aedprojectblue}{agenterrordata.apodex.com}}.
    \end{minipage}\hfill
    \begin{minipage}[c]{.19\linewidth}
      \raggedleft\includegraphics[width=2.8cm]{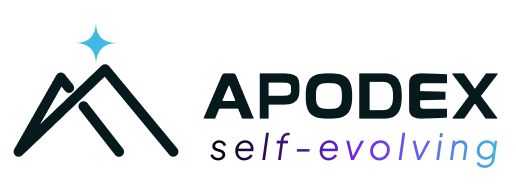}
    \end{minipage}\par
  \end{tcolorbox}%
  \thispagestyle{aedfirstpage}}
\makeatother
\fancypagestyle{aedfirstpage}{%
  \fancyhf{}%
  \fancyfoot[R]{\footerfont\thepage}}

\begin{document}

\begin{abstract}
An unsuccessful LLM agent rollout contains more information than its final reward:
the observations available to the agent, the actions it chose, and the
environment's responses. Reusing this experience for learning requires
identifying a decision to revise and testing a concrete alternative.
We introduce the \textbf{Agent Error Dataset} (\dataset{}), comprising
\aedscale{50{,}228} error--diagnosis pairs from $9{,}961$ source tasks across
\aedscale{33} environments, \aedscale{19} harness families and $23$ policy
models in text-based agent systems.
We retain source traces and execution metadata to support cross-setting
failure analysis and re-diagnosis without repeating the original rollout.
Our five-stage \textbf{Agentic Error-to-Training} (AET) pipeline collects
natural failures, generates diagnoses and proposed corrections, and checks
them against recorded evidence.
Where replay is supported, we compare corrections with original-action retries
from the same checkpoint under matched execution settings.
We then construct separate training views for diagnosis and actor recovery.
\par\smallskip
Across $3{,}062$ matched replay pairs, first-proposal corrections raise verifier
pass rates from $18.4\%$ to $51.1\%$, a gain of \aedstat{32.7} percentage points.
Using a separately frozen diagnosis release, full-diagnosis fine-tuning on
$1{,}656$ source tasks raises Qwen3-8B's exact-step agreement with internal
teacher labels from $47.2\%$ to \aedlearn{63.6}\%, averaged over three seeds
on a $943$-case holdout. The strongest prompted reference in this comparison
scores $54.7\%$; mean agreement improves at each of four increasing training-set
sizes. In a single-seed comparison of actor-training recipes, action-only repair training
scores \aedlearn{6.67} percentage points higher on WebShop-lite than
success-only training.
\end{abstract}

\maketitle
\begingroup
\makeatletter
\renewcommand{\thefootnote}{}
\renewcommand{\@makefntext}[1]{%
  \noindent\makebox[0pt][r]{\hspace{0.4em}}#1}
\footnotetext{We gratefully acknowledge Mr. Tianqiao Chen, the project lead of this work. We sincerely thank him for the vision, guidance, and support that made
this work possible.}
\makeatother
\endgroup

\begin{figure}[H]
\centering
\includegraphics[width=\linewidth]{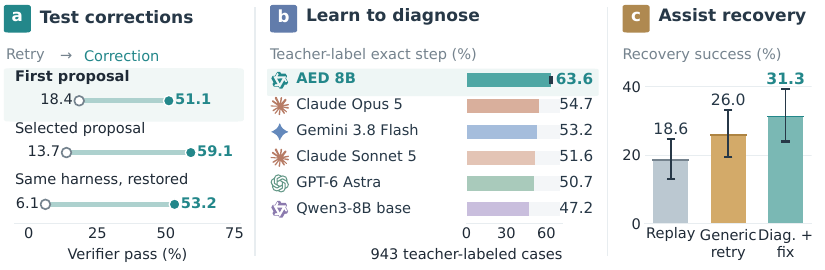}
\caption{\textbf{Failure reuse: correction, diagnosis learning and recovery.}
(a) Executed corrections outperform their matched original-action controls.
(b) AED-trained Qwen3-8B leads the shown references in internal exact-step teacher-label agreement.
(c) Supplied-location assistance improves recovery over replay.
Panels use separate cohorts; protocols and uncertainty:
Section~\ref{sec:experiments} and Appendices~\ref{app:results-detail}, \ref{app:located-cohort}.}
\label{fig:headline-results}
\vspace{-4pt}
\end{figure}

\clearpage
\section{Introduction}
\label{sec:intro}

Recent frontier large language models (LLMs), including GPT-6 Astra and Claude
Fable 5.1, support complex reasoning, coding and long-horizon agentic
work~\citep{openai2026astra,anthropic2026fable}.
These capabilities renew interest in how far language-model systems can generalize
beyond their training tasks~\citep{feng2024howfar}.
Open-weight Qwen3.8-Flash-Next, DeepSeek-V4.1-Flash and Kimi K3 broaden
access to capable agent models~\citep{qwen3.8flashnext,deepseekai2026deepseekv41flash,moonshot2026kimik3}.

These advances extend language-model capabilities~\citep{zhao2026surveylargelanguagemodels}
to interactive software and web tasks~\citep{swebench,webshop}.
Agent harnesses connect these models to executable tools and persistent state.
OpenHands provides a composable software-agent SDK~\citep{openhands_sdk};
$\pi$ agent combines a terminal agent loop with extensible tools, model-provider
interfaces and resumable sessions~\citep{piagent}.
Through these systems, agents inspect repositories, run commands and revise
solutions using tool feedback~\citep{10.1145/3704435,liu2025advanceschallengesfoundationagents}.
Understanding failures therefore requires examining the model together with
the execution system in which it acts.

Agent post-training improves task performance by learning from interaction.
AgentTuning uses reward-filtered trajectories for supervised
fine-tuning~\citep{zeng2024agenttuning}, while RAGEN and AgentRL extend policy
optimization to multi-turn environment feedback~\citep{ragen,agentrl}.
Yet an outcome reward alone does not specify which decision to revise or what
action should replace it. Failed rollouts preserve what the agent observed,
which actions it chose, and how the environment responded. We can turn these
rollouts into training examples by identifying where the agent went wrong,
explaining why, and proposing what it should do instead. This motivates a
dataset of agent failures with diagnoses and corrections across diverse
execution settings.

\begin{rqbox}
\emph{How should we analyze and transform failed agent experience so that
models can learn to recognize, avoid, and repair errors across execution settings?}
\end{rqbox}

Existing work has advanced several aspects of learning from agent failures.
MAST~\citep{mast} characterizes failure modes, and
Who\&When~\citep{whoandwhen} identifies responsible agents and decisive error
steps. These annotations support failure analysis, but do not directly specify
the actions an agent should learn instead. AgenTracer~\citep{agentracer} goes
further by constructing attribution data and training a diagnostic model, while
AgentDebug~\citep{agentdebug} uses corrective feedback to recover from failures.
Their principal learning and recovery objectives, however, differ from training
an acting policy on corrected behavior. Across these resources, differences in
execution settings, annotation targets, and correction evidence also make it
difficult to study diagnosis and action learning on a common collection of
failures. A broader resource is therefore needed to connect natural failures
with explanations, proposed corrections, and execution evidence, and to support
training for both diagnosis and action.

\begin{figure}[t]
\centering
\includegraphics[width=\linewidth]{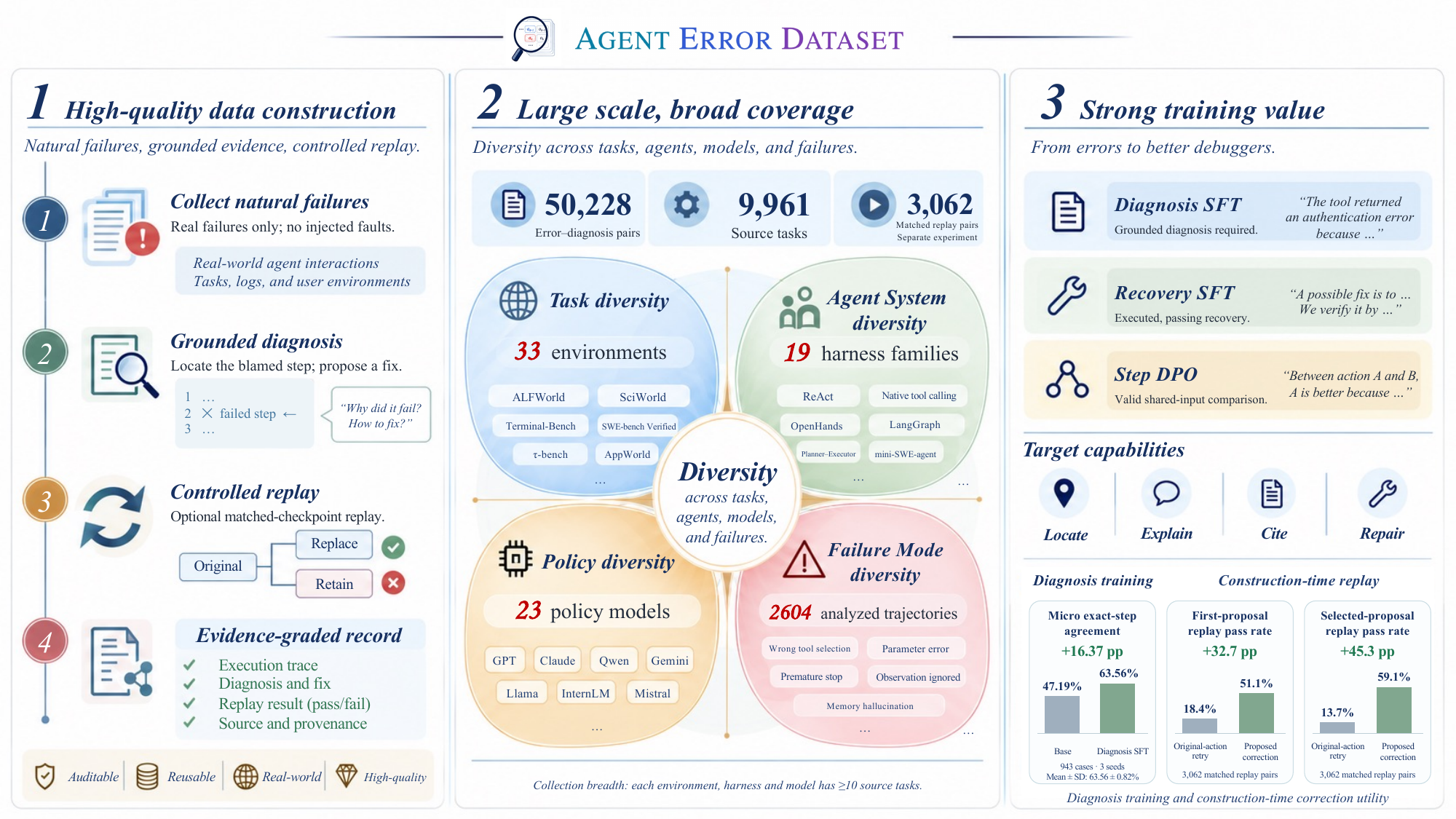}
\caption{\textbf{AET construction and training views.}
Natural failures yield diagnoses, corrections and optional replay evidence.
Coverage, historical error profiles and experiments use separate cohorts.
Replay bars use first/selected proposals ($n=3{,}062$ each) and restored-harness
selected proposals ($n=871$), each with its own control; they measure correction utility.
``Step DPO'' denotes an offline action-preference pilot (Appendix~\ref{app:method-details}).}
\label{fig:overview}
\label{fig:intuition}
\end{figure}

We introduce the \textbf{Agent Error Dataset} (\dataset{}), a collection of
\aedscale{50{,}228} error--diagnosis pairs spanning \aedscale{33} environments,
\aedscale{19} harness families and $23$ policy models.
This breadth supports failure analysis across text-based execution settings;
stored traces also support re-diagnosis without new rollouts.
Our Agentic Error-to-Training (AET) pipeline links diagnoses and proposed
corrections to matched replay outcomes where supported, then constructs
objective-specific training views (Figure~\ref{fig:overview}).
The collection counts pairs, while experiments use a separately frozen
diagnosis release and distinct replay and actor cohorts
(Section~\ref{sec:corpus}).

Our contributions connect \textbf{failure analysis, correction testing and
post-training}.
\textbf{(i) A scaled collection of natural agent failures} retains provenance
across execution settings, supporting analysis of how failures vary with
the model and harness (Section~\ref{sec:corpus}).
\textbf{(ii) AET, a pipeline for reusing failed experience}, links diagnoses,
corrections and optional controlled replay to separate training views.
First-proposal corrections improve matched-replay pass rates by
\aedstat{32.7} percentage points over original-action retries
(Section~\ref{sec:direct}).
\textbf{(iii) An empirical study of error-aware post-training} shows that
full-diagnosis SFT on $1{,}656$ source tasks raises internal exact-step teacher-label agreement from
$47.2\%$ to \aedlearn{63.6}\% across three seeds.

\section{Related Work}
\label{sec:related}

\begin{table}[t]
\centering
\scriptsize
\setlength{\tabcolsep}{2.4pt}
\caption{\textbf{Failure-record resources for LLM agents.}
Errors: natural (nat.), injected (inj.) or synthetic (syn.).
Envs/Harn./Models: environments, harnesses and generating models.
Records: each work's units (AED: error--diagnosis pairs).
Step: localization; Evid.: trace evidence; Exec.\ fix: executed correction;
Ctrl.: same-checkpoint original-action control. Train: debugger (D), agent (A);
Agree.: human agreement (AED: raw step, $59/69$; metrics differ).
\mbox{\pmark{} partial}; ``--'' not stated.
AED counts use one audited collection snapshot.
$^\ddagger$Counts include multimodal traces; AED is text-only.
Agreement definitions and details: Appendix~\ref{app:rw-table-notes}.}
\label{tab:labels}
\resizebox{\linewidth}{!}{\begin{tabular}{@{}l l r r r r c c c c c l@{}}
\toprule
Work & Errors & Envs & Harn. & Models & Records & Step & Evid. & Exec.\ fix & Ctrl. & Train & Agree. \\
\midrule
Who\&When~\citep{whoandwhen}      & nat.   & 2  & 2  & -- & 184 & \cmark & \xmark & \xmark & \xmark & -- & -- \\
Who\&When Pro$^\ddagger$~\citep{whowhenpro} & inj. & 26 & 15 & -- & 12{,}326 & \cmark & \xmark & \xmark & \xmark & -- & $\kappa$ .73 \\
AgenTracer~\citep{agentracer} & nat.+inj. & 6 & 6 & -- & ${>}$2{,}000 & \cmark & \xmark & \pmark & \xmark & D & -- \\
AEGIS~\citep{aegis}               & inj.   & 6  & 6  & 1  & 9{,}533 & -- & \xmark & \xmark & \xmark & D & $\kappa$ .81 \\
MAST~\citep{mast}                 & nat.   & 8  & 7  & 6  & 1{,}642 & \xmark & \xmark & \pmark & \xmark & -- & $\kappa$ .88 \\
TRAIL~\citep{trail}               & nat.   & 2  & 2  & 2  & 841 & \cmark & \cmark & \xmark & \xmark & -- & -- \\
AgentErrorBench~\citep{agentdebug}& nat.   & 3  & -- & -- & 200 & \cmark & \xmark & \cmark & \xmark & -- & $\kappa$ .55 \\
TraceElephant~\citep{traceelephant}& nat.  & 3  & 3  & 1  & 220 & \cmark & \xmark & -- & -- & -- & $\alpha$ .64 \\
TrajErrBench~\citep{trajdebug}    & nat.   & 2  & -- & 8  & 486 & \cmark & \cmark & \pmark & \xmark & -- & $\kappa$ .91/.67 \\
AgentRx~\citep{agentrx}           & nat.   & 4  & -- & -- & 170 & \cmark & \xmark & -- & -- & -- & $\kappa$ .89 \\
STeP~\citep{step}                 & syn.   & 3  & -- & 1  & 708 & \cmark & \xmark & \cmark & \xmark & A & -- \\
\midrule
Agent Error Dataset (AED) & nat. & 33 & 19 & 23 & 50{,}228 & \cmark & \cmark & \pmark & \pmark & D, A & $\humanPreferredStepAgree\%$ \\
\bottomrule
\end{tabular}}

\end{table}

\paragraph{Attribution and failure analysis.}
Understanding failures requires connecting recurring error patterns to decisions
within individual runs. MAST and AdaMAST organize these patterns into
taxonomies~\citep{mast,adamast}, while attribution researchers identify responsible
agents and error locations~\citep{whoandwhen,trail,agentrx}, including in long
traces and spans~\citep{telbench,trajaudit,traceelephant,agentlocate}.
To scale attribution supervision, AgenTracer trains localization models,
Who\&When Pro expands fault injection, and AEGIS verifies injected failures
before training attribution~\citep{agentracer,whowhenpro,aegis}.
For recovery, the question extends to whether a diagnosed error persists and
whether a correction helps: TrajDebug tracks error resolution and introduces
TrajErrBench~\citep{trajdebug}, while AgentDebug and AgentDebugX connect diagnosis
to corrective execution~\citep{agentdebug,agentdebugx}; CUADebug applies diagnosis
and repair to screenshot-based computer use~\citep{cuadebug}.
In AED, we use exploratory error groupings for cross-setting analysis, and link natural failures to diagnoses, proposed
corrections and available replay evidence to support post-training
(Table~\ref{tab:labels}).

\paragraph{Learning from unsuccessful experience.}
Prior work reuses failed actions, segments and goals~\citep{eef,agenther,cso,fate},
constructs reflective recovery trajectories~\citep{agentr,step,failedaction},
and guides later behavior with critiques~\citep{reflexion,score}.
Other resources retain executable environments and verifiers~\citep{toucan,swegym,swesmith};
ADP unifies trajectory formats~\citep{adp}. On a separate replay-supported subset,
we compare corrections with same-checkpoint original-action retries
(Table~\ref{tab:labels}). Appendix~\ref{app:rw-table-notes} details
resource scopes; Appendix~\ref{app:adp-relation} tests trajectory export, not
full-record interoperability.

\section{Agent Error Dataset}
\label{sec:method}

Each record links a failed trace to a diagnosis and any correction-test outcomes.
The available evidence determines which training views it can support.

\subsection{The data record}

We count a run as \emph{failed} when its environment adapter reports an unsuccessful
terminal outcome within the allowed budget. This collection-level outcome does not
locate an error: attributing an \emph{agent error} requires a trace-supported,
avoidable decision. Appendix~\ref{app:failure-definition} distinguishes task
failures, recovered tool errors and ungradable runs.

A record links a failed action--observation trace to one diagnosis, its proposed correction,
review decisions and available replay branches. It also records the policy, harness, debugger
and information used in construction, alongside intended uses and limitations following
dataset documentation practice~\citep{gebru2021datasheets}. Multiple proposals for one failure remain distinct
attempts, grouped by source task for splitting and analysis. Diagnoses are free text rather than
assignments to a fixed error taxonomy; error modes are induced afterwards without changing the
underlying labels (Appendix~\ref{app:atlas-exploratory}). Appendix~\ref{app:schema} gives the
schema and a worked example.

\subsection{AET: a five-stage data generation pipeline}
\label{sec:generation}

AET connects five stages (Figure~\ref{fig:overview}).
\textbf{(1) Collect} natural failures across environments, harnesses and policies,
preserving action--observation traces and screening out known infrastructure or grader
faults from agent-error targets.
\textbf{(2) Diagnose} the error location and responsible agent, with a trace-cited
explanation and proposed correction, following AgentDebug~\citep{agentdebug}
through \mbox{AgentDebugX}~\citep{agentdebugx}.
\textbf{(3) Ground} the diagnosis in the student-visible trace through structural
and semantic checks, recording accepted and rejected proposals.
\textbf{(4) Replay}, where supported, tests the correction against a fresh
original-action retry from the same checkpoint under matched policy, harness,
budget and verifier settings. Both outcomes are retained; recovery supports a
correction under those conditions, without establishing a unique root cause.
\textbf{(5) Build views} for diagnosis SFT, recovery SFT and action preferences,
each with its own evidence and split requirements (Section~\ref{sec:train}).
Diagnosis records require no replay; recovery targets require an executed,
passing continuation, and preferences require a valid shared-input comparison.
Stored failures can receive additional diagnoses without new rollouts, while
records that fail a view's admission rules remain available for collection
audits. Appendix~\ref{app:method-details} gives the stage contracts and
Appendix~\ref{app:quality-rubric} the review rubric.

\subsection{Collection scope and training subsets}
\label{sec:corpus}
\begin{figure}[t]
\centering
\includegraphics[width=\linewidth]{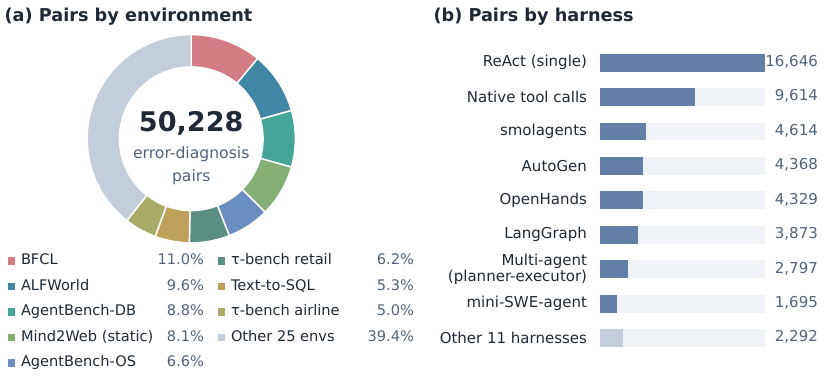}
\caption{\textbf{Error--diagnosis pair composition, current audit.}
Both panels include all $50{,}228$ source-linked pairs.
(a) Environment shares. (b) Harness counts; small sources are grouped.
Across this cohort, $33$ environments and $19$ harness families meet the
ten-source-task floor.
These are collection shares, not failure rates or training-admission rates
(Appendix~\ref{sec:scope}); the full distributions appear in Appendix~\ref{app:corpus-full-distribution}.}
\label{fig:corpus}
\end{figure}

We check source linkage, error-step presence and trace citations before counting a pair.
The environment inventory includes benchmarks, synthetic tasks and simplified ports;
it is not a count of public benchmarks.
An error--diagnosis pair joins one failed execution with one recorded diagnosis.
Different model, seed or temperature rollouts can contribute distinct failed runs;
additional diagnoses of one run add pairs, not runs. Source-task grouping keeps
related executions together for splitting and uncertainty estimates.
Collection membership does not imply training eligibility or a verified repair.
Figure~\ref{fig:corpus} shows $50{,}228$ pairs linked to $38{,}278$ stored source-trace
blobs and $9{,}961$ source tasks ($1.31$ pairs per blob).
Blob identities do not establish a count of independent executions.
Both panels use this index; Appendix~\ref{sec:scope} distinguishes
collection counts from training subsets.

The separately frozen diagnosis release used in Section~\ref{sec:experiments} contains
$4{,}319$ rows over $2{,}309$ source tasks in $15$ environments, each environment produced by up to
nine harness families and ten policy models (Table~\ref{tab:release-scale}).
The collection, replay cohort and objective-specific training subsets have different admission rules.

Appendix~\ref{app:collection-analysis} analyzes source coverage and diagnosis
multiplicity on this same collection index.

\begin{figure}[t]
\centering
\includegraphics[width=\linewidth]{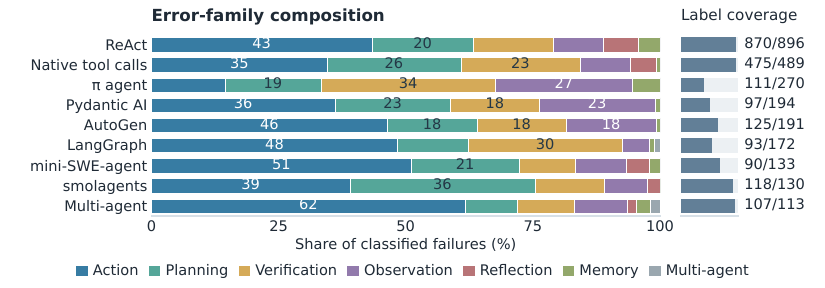}
\caption{\textbf{Failure profiles across harnesses.}
Historical machine labels: $2{,}092$ of $2{,}604$ trajectories classified.
The nine displayed groups contain $2{,}588$ trajectories ($2{,}086$ classified);
$16$ trajectories fall outside these groups.
Bars show conditional family shares; coverage retains abstentions.
Task mixtures differ across harnesses, so this is an exploratory subset comparison,
not a causal effect or the full-collection distribution
(Appendix~\ref{app:failure-composition}).}
\label{fig:taxonomy-harness}
\begin{resulttakeaway}{Exploratory error profiles differ across harnesses}
Action-error labels lead in $8$ of the $9$ shown harnesses,
including the planner--executor harness ($61.7\%$).
The $\pi$ agent instead concentrates labels on verification and observation ($61.3\%$).
These exploratory profiles motivate error-type coverage checks alongside
environment counts.
\end{resulttakeaway}
\end{figure}

\paragraph{Failure profiles across harnesses.}
We also examine an earlier, trajectory-deduplicated taxonomy study
(Figure~\ref{fig:taxonomy-harness}). The differences motivate checking error-type
coverage alongside environment counts when selecting training examples.
Appendix~\ref{app:failure-composition} identifies this historical subset and
its sampling and labeling limitations.

\section{Training Views}
\label{sec:train}
AED constructs learning examples from failures by selecting the visible history
and target responses, then applies standard post-training objectives.
For example $i$, let $x_i$ be the visible context, $y_i$ the target response,
$m_{ik}$ its binary token mask and $M_i=\sum_k m_{ik}$ the number of target tokens.
Real examples have $M_i>0$; actor padding contributes zero loss and zero tokens.
With model $\pi_\theta$ and optimizer window $\mathcal B$, the evaluated SFT losses are
\begin{equation}
\begin{aligned}
\ell_i(\theta)&=-\sum_k m_{ik}\log\pi_\theta(y_{ik}\mid x_i,y_{i,<k}),\\
\mathcal L_{\mathrm{diag}}(\theta)&=\frac{1}{|\mathcal B|}\sum_{i\in\mathcal B}\frac{\ell_i(\theta)}{M_i},
\qquad
\mathcal L_{\mathrm{actor}}(\theta)=\frac{\sum_{i\in\mathcal B}\ell_i(\theta)}{\sum_{i\in\mathcal B}M_i}.
\end{aligned}
\label{eq:sft}
\end{equation}
Only target assistant tokens receive loss, including the actor end-of-turn token;
context and tool observations receive no loss. Diagnosis uses per-example means;
actor normalization spans all ranks and accumulation steps.

\paragraph{Diagnosis.}
Diagnosis pairs the failed trace $x_i=\tau_i$ with
$y_i=d_i=(t_i^*,u_i^*,e_i,a_i^*)$: attributed step, responsible agent,
explanation with evidence and proposed correction. Compact targets retain
only attribution, optionally with a short rationale.
Inputs exclude construction-only verifier context, successful references and
replay outcomes; admission does not require successful replay.

\paragraph{Prevention and recovery.}
Each preventive example pairs pre-error history with an executed passing action,
$(x_i,y_i)=(h_{<t},a_t^+)$, using the inference chat prefix.
The post-error variant adds the original erroneous action and its rejection as masked context,
then supervises the correction, with or without a diagnosis-derived reflection.
Only state-preserving rejections enter post-error context; other repair examples
use pre-error history. Neither form includes the later failed suffix.
Section~\ref{sec:g3} specifies the success/repair mixtures.

\paragraph{Outcome and preference views.}
Replay retains corrected-action and original-action outcomes, including negative
and zero contrasts. A preference pair requires executed alternatives from the same
state and a positive outcome contrast under the construction protocol.
We evaluate SFT; the offline action-DPO~\citep{dpo} pilot establishes no recovery
benefit (Appendix~\ref{app:dpo-pairs}).
Appendix~\ref{app:learning-objectives} gives the loss definitions and implementation checks.

\section{Experiments}
\raggedbottom
\label{sec:experiments}\label{sec:produced}
\label{sec:repair}\label{sec:exp3}\label{sec:located}\label{sec:exp1}\label{sec:exp2}
We evaluate diagnosis production, correction utility, diagnosis learning and actor
training recipes. Appendix~\ref{app:experiment-claims} specifies comparison scope.

\paragraph{Setup.}\label{sec:setup}
Trainable models start from Qwen3-8B~\citep{qwen3}; each study uses a frozen,
objective-specific population.
Internal splits hold out source tasks; we audit public overlap below.
Missing or unparseable responses count as misses. Public comparisons share
visible inputs and scorers within each protocol; exact-step scores use
task-family macro or case-level micro averages. Seed SD measures run variability;
paired task-family intervals measure evaluation sampling. Actors measure
initial-state task success. Appendices~\ref{app:g2-designs}
and~\ref{app:result-protocols} specify checkpoint selection, budgets, prompts and admission.

\subsection{Diagnosis production and correction utility}
\label{sec:direct}
The citation-first judge produces trace-cited diagnoses for $93.5\%$ of the
common failure pool, compared with $48.4$--$65.8\%$ for the evaluated
alternatives (Table~\ref{tab:pipeline-main}b in the appendix). This measures
production yield, not independent label accuracy; rendering and teacher access
differ across configurations.

First-proposal corrections raise verifier success from $18.4\%$ to $51.1\%$,
a $32.7$-point gain (task-clustered $95\%$ interval: $28.4$--$37.0$) over
original-action retries without selecting among repeated proposals
(Figure~\ref{fig:headline-results}a).
Appendix~\ref{app:concise-result-details} retains paired uncertainty, costs,
the search-selected contrast and restored-harness sensitivity;
Appendix~\ref{app:failure-record-insights} analyzes paired outcomes and revisions.

In a separate supplied-location study, diagnosis-guided continuation scores
$31.3\%$, versus $26.0\%$ for generic reconsideration and $18.6\%$ for replay
(Figure~\ref{fig:headline-results}c). Its paired gain over generic reconsideration
remains uncertain (Appendix~\ref{app:located-cohort}). Both studies test correction
at a supplied location; neither tests autonomous detection or actor post-training.
The replay cohort does not join the frozen diagnosis release. Matched-size filtering
establishes no learning benefit from stricter admission under the tested recipe
(Appendix~\ref{app:concise-result-details}).

\subsection{Can AED train a competitive failure-diagnosis model?}
\label{sec:public-attribution}\label{sec:g2}
\begin{figure}[t]
\centering
\includegraphics[width=\linewidth]{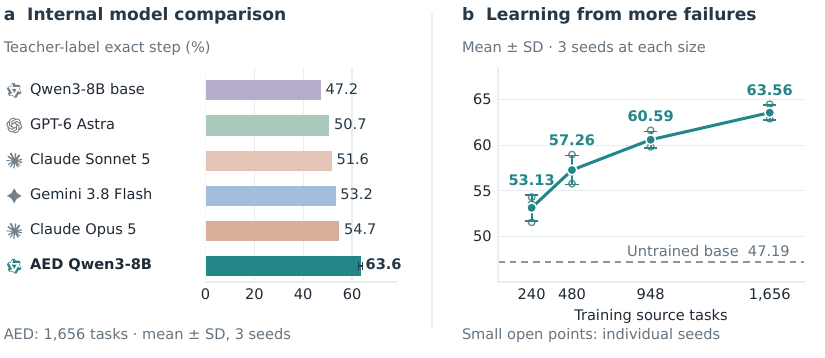}
\caption{\textbf{Diagnosis learning and training scale.}
(a) Exact-step agreement with internal teacher labels against prompted references.
(b) Four nested training sizes, each with three seeds: means $\pm$ sample SD,
individual runs (open points) and untrained base (dashed line).
All scores are case-level micro averages against recorded teacher labels on the same $943$ cases.
Checkpoint selection varies across sizes; protocols, score provenance and
complete results: Appendix~\ref{app:fullft-scaling} and Table~\ref{tab:training-results}.}
\label{fig:diagnosis-learning}
\end{figure}

\paragraph{Internal localization.}
Full-diagnosis SFT raises exact-step agreement at each of four nested training
sizes; even the smallest subset improves on the untrained base
(Figure~\ref{fig:diagnosis-learning}). We report case-level micro agreement because
the responsible-agent label is constant on this holdout.
The student learns from consensus-generated labels, whereas the references receive
no training on these labels. The comparison measures agreement with the recorded
annotations, including their conventions and defects, rather than a general
ranking of diagnostic ability. Appendix~\ref{app:arm-privilege} explains the
heterogeneous teachers and the recorded Gemini~3.6~Flash arbiter.

\paragraph{Independent labels and protocol sensitivity.}
Public benchmarks~\citep{whoandwhen,trajdebug} provide independent labels but can share tasks with training.
The $1{,}656$-task arm's three seeds improve on the base on Who\&When under the
unified protocol. These full-cohort scores include tasks shared with training.
After excluding flagged task overlap, its interval includes zero
(Table~\ref{tab:external-task-overlap}). Retraining after replacing flagged examples
retains a positive contrast under this protocol (Table~\ref{tab:public-retrain-complete}),
but does not establish broad transfer. Mean TrajErrBench accuracy remains below
base. We retain one checkpoint per seed across benchmarks (Appendix~\ref{app:external-transfer}).

Table~\ref{tab:public-main} tests the $948$-task, seed-$17$ student under a
different public protocol. Full-diagnosis training loses both responsible-agent
and exact-step accuracy on Who\&When; answer-format continuation recovers part of
the deficit but changes
training exposure as well as format. AgentErrorBench point estimates improve
without a clear paired advantage. Under the same frozen cases and first-call
contract (Table~\ref{tab:public-reference}), the answer-format student exceeds four
prompted frontier references on responsible-agent attribution in both hand-crafted
conditions, but not on the algorithm-generated conditions, exact step or
AgentErrorBench. Appendix~\ref{app:public-reference-protocol} explains the
student's reused first calls, the base run and the token-budget exclusion of two
further references. The larger arm also loses Who\&When accuracy under
benchmark-specific protocols, which change input rendering, answer schema and
step indexing as well as wording.

\begin{table}[t]
\centering\footnotesize\resultsetup
\caption{\textbf{Public attribution under a shared prompt.}
All-case percentages for the $948$-task, single-seed student and its
answer-format continuation. HC/AG: hand-crafted/algorithm-generated;
gold is the task answer. Budget-forced continuation is enabled.
Protocols and comparison scope:
Appendix~\ref{app:experiment-claims}.}
\label{tab:public-main}
\begin{tabular*}{\linewidth}{@{\extracolsep{\fill}}P{.36\linewidth}*{4}{>{\raggedleft\arraybackslash}p{.14\linewidth}}@{}}
\toprule
\resultgroup{5}{(a) Who\&When: agent / exact step (\%)}
Model or training resource & HC & HC + gold & AG & AG + gold \\
\midrule
Qwen3-8B base & $56.90/20.69$ & $51.72/18.97$ & $56.35/\mathbf{37.30}$ & $55.56/\mathbf{38.89}$ \\
\dataset{} diagnosis SFT & $50.00/15.52$ & $44.83/13.79$ & $23.81/15.87$ & $11.11/9.52$ \\
\quad $+$ answer-format diversity & $\mathbf{70.69}/\mathbf{24.14}$ & $\mathbf{68.97}/\mathbf{22.41}$ & $\mathbf{59.52}/34.92$ & $\mathbf{58.73}/34.13$ \\
\midrule
\resultgroup{5}{(b) AgentErrorBench: exact step / step+module (\%)}
Model or training resource & ALFWorld & WebShop & GAIA & Env.\ macro \\
\midrule
Qwen3-8B base & $12.00/1.00$ & $18.00/0.00$ & $20.00/4.00$ & $16.67/1.67$ \\
\dataset{} diagnosis SFT & $\mathbf{19.00}/\mathbf{5.00}$ & $\mathbf{20.00}/\mathbf{2.00}$ & $22.00/\mathbf{8.00}$ & $20.33/\mathbf{5.00}$ \\
\quad $+$ answer-format diversity & $18.00/\mathbf{5.00}$ & $18.00/0.00$ & $\mathbf{30.00}/4.00$ & $\mathbf{22.00}/3.00$ \\
\bottomrule
\end{tabular*}
\par\smallskip\raggedright\scriptsize Bold: highest displayed score per metric among the rows shown, including ties; not statistical significance.

\end{table}

\begin{table}[t]
\centering\footnotesize\resultsetup
\caption{\textbf{Frontier references under a shared prompt.}
All-case responsible-agent / exact-step accuracy (\%) on the same frozen cases, first call
only, with no continuation responses scored. Bold marks per-metric column maxima.
Student first calls come from the run in Table~\ref{tab:public-main}.
Run provenance, repeat variation and exclusions:
Appendix~\ref{app:public-reference-protocol}.}
\label{tab:public-reference}
\begin{tabular*}{\linewidth}{@{\extracolsep{\fill}}P{.22\linewidth}*{5}{>{\raggedleft\arraybackslash}p{.133\linewidth}}@{}}
\toprule
& \multicolumn{4}{c}{Who\&When} & \\
\cmidrule(lr){2-5}
Model or training resource & HC & HC + gold & AG & AG + gold & AEB \\
\midrule
\resultgroup{6}{Prompted references, same unified attribution prompt}
Gemini 3.7 Flash & $25.86/8.62$ & $25.86/10.34$ & $38.10/26.19$ & $38.89/27.78$ & $21.50/25.00$ \\
GPT-5.6 Sol & $51.72/\mathbf{25.86}$ & $58.62/\mathbf{34.48}$ & $\mathbf{67.46}/\mathbf{55.56}$ & $\mathbf{66.67}/\mathbf{49.21}$ & $18.50/25.50$ \\
Claude Opus 5 & $43.10/24.14$ & $50.00/22.41$ & $49.21/32.54$ & $48.41/33.33$ & $16.50/23.50$ \\
Claude Sonnet 5 & $56.90/20.69$ & $60.34/32.76$ & $65.87/49.21$ & $65.87/48.41$ & $\mathbf{23.00}/\mathbf{27.50}$ \\
\addlinespace[3pt]
\resultgroup{6}{Ours, same prompt and the same frozen cases}
Qwen3-8B base & $22.41/8.62$ & $17.24/8.62$ & $18.25/15.87$ & $19.05/13.49$ & $6.50/16.50$ \\
\quad $+$ \dataset{} diagnosis SFT, answer-format & $\mathbf{70.69}/24.14$ & $\mathbf{68.97}/22.41$ & $59.52/34.92$ & $58.73/34.13$ & $10.50/21.00$ \\
\bottomrule
\end{tabular*}

\end{table}

\paragraph{Matched resource comparison.}\label{sec:cross-resource}\label{sec:g1}
With a common base, compact-attribution recipe and matched source-task count,
AED and AgenTracer data each lead on a different test distribution
(Table~\ref{tab:cross-resource}). AgenTracer's largest advantage occurs on
injected errors, alongside task and annotation differences. Compact-target AED
does not beat the base internally. This compares training resources under one
student recipe, not AgenTracer's RL system~\citep{agentracer} or a controlled
ablation of full-diagnosis targets.

Overlap audits place the scaling gain outside identified high-overlap subsets,
but training and holdout share environments and many harness--policy combinations
(Appendix~\ref{app:overlap-sensitivity}).

\begin{resulttakeaway}{Internal agreement improves; broad transfer is environment dependent}
Full-diagnosis supervision improves agreement with the recorded internal labels.
External results depend on the benchmark, answer contract and task overlap;
resource rankings reverse across test distributions.
\end{resulttakeaway}

\subsection{Can repair supervision improve the acting policy?}
\label{sec:g3}
We compare success-only SFT with preventive, post-error action-only and reflective
repair supervision. The success-only control follows interaction
tuning~\citep{zeng2024agenttuning} using our pool, not AgentTuning's data.
All actors start held-out tasks without a test-time teacher. The single-seed
recipes differ in task pools, exposure and optimizer updates; their contrasts
measure the combined recipe, not repair supervision alone.

\begin{figure}[t]
\centering
\includegraphics[width=\linewidth]{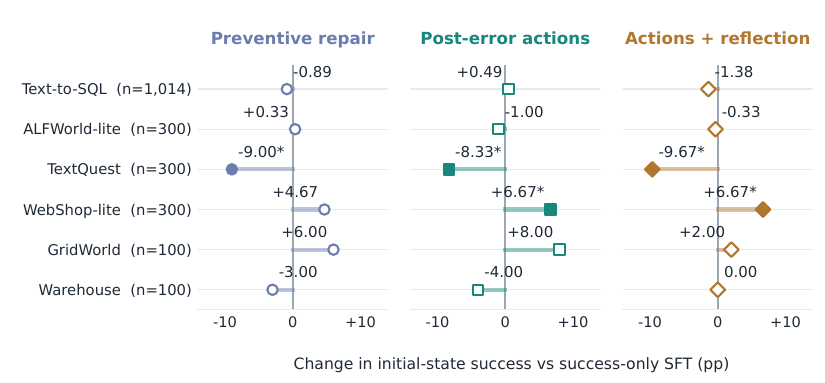}
\caption{\textbf{Repair-containing recipes: gains and losses across environments.}
All $18$ paired percentage-point contrasts against success-only SFT;
filled markers and $*$ indicate post-hoc Holm-adjusted significance across tasks,
conditional on one training seed. Task pools and exposure differ. Planned denominators include
run errors. Absolute scores and the untrained base:
Appendix~\ref{app:actor-full-table}.}
\label{fig:actor-contrasts}
\end{figure}

The evaluated repair-containing recipes score higher on WebShop-lite and lower on TextQuest
(Figure~\ref{fig:actor-contrasts}). Action-only and reflective WebShop gains and all TextQuest
losses survive the post-hoc multiplicity adjustment. Other differences are not
detected, which does not establish equivalence. Reflection adds no detected
benefit over action-only targets.

WebShop-lite gains accompany shorter episodes, while many newly lost TextQuest
tasks hit the step limit. These associations do not isolate a budget mechanism
from exposure or update-count differences
(Appendix~\ref{app:concise-result-details}).

\paragraph{Transfer scope.}
The six development environments test held-out tasks within actor-training
environments, outside the frozen core diagnosis corpus. On real ALFWorld and
ScienceWorld, the preventive arm loses to the untrained policy. Only that recipe
has this real-environment comparison; ScienceWorld omits six run errors from its
repair denominator (Appendix~\ref{app:concise-result-details}).

\begin{resulttakeaway}{Repair utility is environment-dependent}
The evaluated repair-containing recipes improve WebShop-lite. TextQuest and
real-environment losses limit the result to the tested recipes and environments.
\end{resulttakeaway}

\FloatBarrier
\section{Discussion and Conclusion}
\label{sec:discussion}
AED organizes failed experience for analysis, diagnosis and corrective supervision.
We find useful corrections and improved agreement with internal diagnostic labels,
with protocol-dependent public transfer and mixed actor outcomes.
The $50{,}228$-pair collection supports analysis across execution settings;
learning results concern smaller frozen populations. Historical supervision
omits policy context, and later checks do not validate it retroactively
(Appendix~\ref{app:limitations}). Correction utility, label validity and learned
recovery each require their own evidence.

\clearpage
\subsubsection*{Ethics statement}
We construct AED from public benchmarks, synthetic tasks, and simulated
environments to study agent behavior, without seeking to collect personal
information. Repository-based tasks, such as SWE-bench~\citep{swebench},
may nevertheless expose contributor names, email addresses, or identifying
text in code comments and issue discussions. This source-dependent privacy
risk is shared with software-engineering benchmarks and can persist in
derived trajectories. Public availability alone does not establish that
the content is free of personal information or unrestricted for redistribution.
Any release of our data will therefore be subject to source-specific privacy
review and licensing restrictions, including redaction of sensitive content
and unnecessary personal identifiers while preserving required attribution.

The human review involved four paper authors with doctoral or AI/LLM
research backgrounds who participated voluntarily. We report their
judgments using anonymous rater identifiers and aggregate statistics.

\subsubsection*{AI use statement}
We used generative AI to aid writing and editing; to support research ideation
and execution, including experimental design, method implementation, data
processing, analysis and interpretation; and to generate synthetic tasks,
diagnoses and proposed corrections as described in Section~\ref{sec:generation}
and the construction appendices. We also used AI for literature discovery,
design critique and illustration. We test generated code and check measured
claims against sources and versioned records; automated checks and the completed
AI-assisted human audit have the scopes stated in Appendices~\ref{app:release-audit}
and~\ref{app:human-design}.
Figure~\ref{fig:overview} uses author-supplied artwork.
The authors take responsibility for the paper and its claims.

\subsubsection*{Reproducibility statement}
Appendices~\ref{app:certify}, \ref{app:protocols}, and~\ref{app:repro} specify splits,
interventions, statistics and run manifests. Following acceptance, we plan to
open-source the code on GitHub and release model checkpoints and data, subject
to source licenses, privacy review and redistribution rights. The planned
release includes splits, prompts, evaluation protocols and result generators.

\bibliographystyle{iclr2027_conference}
\bibliography{main}

\appendix
\startcontents[appendix]
\clearpage
\raggedbottom
\section*{Supplementary Material}
\label{app:contents}
This supplement opens with the scope and limitations of the collection and experiments
(Appendix~\ref{app:limitations}). The dataset analyses in Appendix~\ref{app:analysis-core}
examine coverage, error profiles, and corrective
evidence. Full-collection coverage uses the source-linked pair index in Figure~\ref{fig:corpus};
the error-profile and rollout studies identify their dated cohorts and counting units.
Training and replay experiments retain their evaluated cohorts.
Debugger-training results appear in the main text and Appendix~\ref{app:training-experiments}.
Actor results measure initial-state task success under the evaluated training recipes.

\subsection*{Contents}
\begingroup
\small
\printcontents[appendix]{}{1}{\setcounter{tocdepth}{1}}
\endgroup

Collection attempts, the frozen diagnosis split and replay trials have different
denominators (Table~\ref{tab:population-map}). Human-review results appear in Appendix~\ref{app:human-results}.

Read Appendix~\ref{app:limitations} for the relation between collection scale and training use,
Appendix~\ref{app:analysis-core} for the findings and their evidence,
Appendix~\ref{app:recovery-case} for a worked repair, and
Appendix~\ref{app:training-experiments} for training and baseline comparisons.
Counting rules and release scope appear in Appendix~\ref{sec:scope};
protocols and prompts appear in Appendices~\ref{app:protocols} and~\ref{app:prompts}.

\clearpage
\section{Limitations and Reproducibility}
\label{app:repro}
\label{sec:limitations}
\label{app:limitations}
\label{app:manifest}
\paragraph{Collection scale and training use.}
\label{app:collection-training-scope}
We use the full collection to measure coverage across tasks, environments,
harnesses and policies (Figure~\ref{fig:corpus}). Detailed error-profile and rollout
studies use dated cohorts with their own denominators
(Appendix~\ref{app:analysis-core}). Collection size counts error--diagnosis pairs;
it does not count independent executions or training-ready examples.

Diagnosis supervision requires a supported attribution; repair SFT requires an
executed passing continuation with valid history, and preference learning requires
comparable executed alternatives (Table~\ref{tab:learning-objectives}).
Task grouping and checks of student-visible information guide selection;
they do not establish that smaller subsets are optimal.
The observed gains concern frozen, objective-specific subsets and recipes;
they do not measure full-collection utility or show that heterogeneous sources
cannot be pooled.

The final collection index, earlier frozen diagnosis subsets and separate
actor-development pool are versioned populations, not a verified nested
filtering chain from the final collection (Table~\ref{tab:population-map}).
We report selection within each experiment's source population; their counts
are not additive. Inclusion in coverage statistics does not establish a learning
benefit for each retained record.

\paragraph{Evaluated version and subsequent revisions.}
The diagnosis experiments use an earlier frozen release with incomplete policy context:
construction omitted policy system prompts and tool lists and predates the
current semantic trace-support and expanded failure-screening checks. Replay
already existed; we do not claim that all collection rows passed the later checks.
We retain the original inputs and splits and identify replacement-data conditions.
This does not resolve the earlier label defects (Appendix~\ref{app:release-audit}).
The completed historical human audit appears in Appendix~\ref{app:human-design}.
These results measure learning from the recorded supervision; they do not
validate the revised pipeline or the full collection.

\begin{artifactbox}{Scope of the reported evidence}
\label{app:data-use-scope}
\small
\begin{tabular}{@{}P{.20\linewidth}P{.73\linewidth}@{}}
\toprule
Scope & Interpretation of the measured results \\
\midrule
Label validity & Quotation matching checks presence, not entailment or a unique
cause. Teacher references and reconstructed traces require student-visibility
audits. The assisted human audit finds higher agreement on error location than
on diagnosis acceptance; its coverage-selected sample cannot estimate collection-wide accuracy. \\
Construction benefit & Paired replay tests correction utility, not attribution
accuracy. Production yield and historical student comparisons differ in teacher
access or task selection and do not isolate pipeline label quality. \\
Coverage and error rates & Failure-only samples cannot estimate deployment failure
rates. Unequal task and model mixtures confound harness rankings and model-version
trends. Text-only coverage, sparse coding cells and procedural stand-ins limit
generality; environment counts are not counts of public benchmarks. \\
Diagnostic transfer & Internal holdouts share environments and often harnesses
and policies with training. Public comparisons depend on reference access, prompts,
coordinates and budgets. Localization on known failures does not test
general failure detection, and current comparisons do not establish public SOTA. \\
Learned repair & Paired replay measures an intervention outcome, not learned
autonomous recovery. Actor gains vary by environment, and the single-seed recipe
comparisons also differ in exposure and update count. The action-DPO pilot measures
offline preferences rather than task recovery. \\
\bottomrule
\end{tabular}
\end{artifactbox}
Filtering changes task composition as well as label quality; without controlling
common task support, a training comparison cannot isolate these effects.
Appendix~\ref{app:experiment-claims} explains the scope of these comparisons.

\paragraph{Reproducibility materials.}
\label{tab:artifacts}
We retain manifests, splits, responses, replay outcomes, prompts, checksums and
exclusion reasons. Appendix~\ref{sec:scope} specifies task grouping;
Appendix~\ref{app:construction-costs} separates measured costs from projections.

\FloatBarrier
\label{app:repro:end}

\clearpage
\section{Dataset analysis and corrective evidence}
\label{app:analysis-core}
\subsection{Coverage and multiplicity in the current collection}
\label{app:collection-analysis}
This analysis uses all $50{,}228$ source-linked pairs in Figure~\ref{fig:corpus},
with the same pair identities and source-task grouping. We recompute the counts
from the hash-bound metadata index; no historical labeling sample enters this summary.
A source blob is a distinct stored source-trace object; its identity alone does
not establish an independent execution.

\begin{table}[ht]
\centering\small
\caption{\textbf{Current collection coverage and multiplicity.} Pair, source-blob and
task counts describe different units. Blob identities do not establish independent executions.}
\label{tab:current-collection-analysis}
\begin{tabular}{@{}lr@{}}
\toprule
Measure & Count \\
\midrule
Error--diagnosis pairs & $50{,}228$ \\
Distinct source blobs & $38{,}278$ \\
Environment-scoped source tasks & $9{,}961$ \\
Environments / harness families / policy models & $33 / 19 / 23$ \\
Nonempty environment--harness cells & $280$ \\
Source blobs with multiple diagnoses & $9{,}846$ \\
Pairs per source blob, mean & $1.31$ \\
Pairs per source task, median / 90th percentile & $2 / 10$ \\
\bottomrule
\end{tabular}
\end{table}

\paragraph{Collection breadth and concentration.}
BFCL, the largest source, supplies $11.0\%$
of pairs; the five largest environments supply $44.0\%$.
The five largest harness families supply $78.8\%$,
so the collection is broad but unevenly represented.
Table~\ref{tab:env-types} retains every environment and its source-task support,
including benchmark adaptations and simplified tasks. These shares describe
collection composition; without successful-rollout denominators they cannot rank
environment difficulty, models or harnesses.

\paragraph{Additional diagnoses share their underlying trace.}
Multiple diagnoses add $11{,}950$ pairs beyond one
per source blob. They can provide alternative explanations or corrections, but
do not add independent failure observations. Analyses of error prevalence must
group these records by trace and account for repeated source tasks.

\paragraph{Scope of the remaining analyses.}
Error-type analyses describe the historical labeled cohorts identified in their
captions. The replay analysis below uses its own executed cohort; the frozen
diagnosis release and actor experiments likewise retain their original inputs
and denominators. Collection growth does not change those experimental results.

\subsection{Repair outcomes and diagnostic revision}
\label{app:failure-record-insights}
MAST distinguishes failure profiles from system rankings and relates failure modes to
task outcomes~\citep{mast}. We extend this analysis perspective to two properties
recorded by AED: what happens when an action changes, and how the diagnosis changes
after an unsuccessful correction. The cohorts below have different inclusion rules;
neither estimates prevalence in the final collection. AgentDebug motivates diagnosis-guided
replay~\citep{agentdebug}; this analysis examines the recorded outcomes, not a new repair algorithm.

\begin{figure}[ht]
\centering
\includegraphics[width=\linewidth]{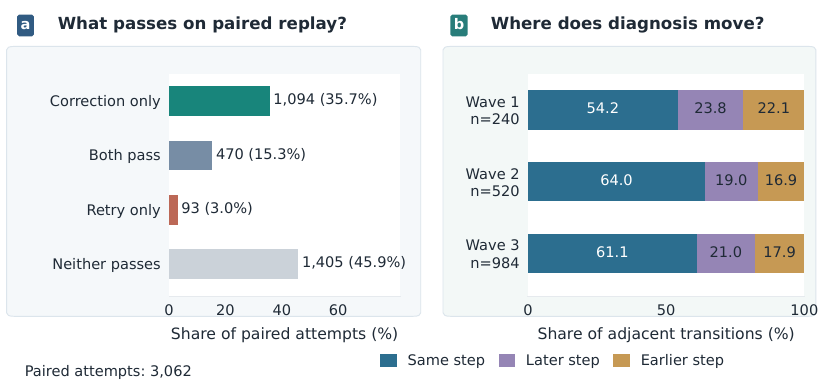}
\caption{\textbf{Two views of corrective evidence.}
(a) First-or-only proposal and original-action retry on $3{,}062$ paired attempts;
all four outcomes remain visible. (b) Adjacent attributed-step transitions across
three development waves, with each wave's transition count. Transitions within an
attempt are dependent. The panels use separate dated cohorts and aggregate reports;
they do not describe the final collection or assign validated root causes.}
\label{fig:failure-record-insights}
\end{figure}

\paragraph{Successful correction alone does not isolate corrective benefit.}
The first proposal succeeds on $1{,}564$ paired attempts, but the original-action
retry also succeeds on $470$ of them ($30.1\%$). The correction-only and retry-only
cells contain $1{,}094$ and $93$ attempts; $1{,}405$ fail in both arms.
Thus the net paired gain comes from the difference between the discordant cells,
not from counting all successful corrections as improvement.
The existing task-clustered interval for this contrast appears in Table~\ref{tab:pipeline-main}.
These observations motivate retaining successful retries and failed corrections,
which distinguish passing branches from evidence for an action preference.
A single paired outcome does not identify an individual causal effect, prove a
repair unnecessary, or satisfy shared-state preference admission by itself.

\paragraph{Revision usually revisits a location.}
Across $405$ multi-round attempts, $1{,}064$ of $1{,}744$ adjacent diagnoses retain
the same attributed step ($61.0\%$); $363$ move later and $317$ earlier.
The same-step category is the largest in each wave, rather than an artifact of
pooling waves. A fixed location leaves the explanation and proposed action free
to change. This argues for retaining the revision history, including unsuccessful
proposals, when studying how to correct a failure. It does not establish that
later diagnoses are more accurate or that more rounds cause better recovery.
Table~\ref{tab:repair-rounds} reports observed later-round yield and its stopping limits.

\begin{resulttakeaway}{Separate a passing branch from a useful correction}
Outcome-only records cannot distinguish success also reached by retrying the original
action. Final-diagnosis-only records hide the revisions used to reach that branch.
AED preserves both, enabling analyses of correction utility and hypothesis revision
under their recorded protocols. Training benefits require the separate controlled
comparisons in Section~\ref{sec:experiments}.
\end{resulttakeaway}

\paragraph{Success, symptoms and semantic labels measure different things.}
In MAST, some annotated failure modes also occur in successful traces.
Our successful-trajectory prompt control likewise changes error-reporting behavior
when the prompt permits abstention (Appendix~\ref{app:false-alarm}). Neither task
success nor an error claim provides a semantic correctness label by itself.
We therefore keep symptom counts, judge-assigned categories and executed outcomes
separate. The labeling-method study in Appendix~\ref{app:error-taxonomy}
reports judge disagreement; it does not estimate error frequencies in the full collection.

\FloatBarrier
\clearpage
\subsection{Crossed coverage of the current collection}
\label{app:current-crossed-coverage}
\begin{figure}[H]\centering
\includegraphics[width=.86\linewidth]{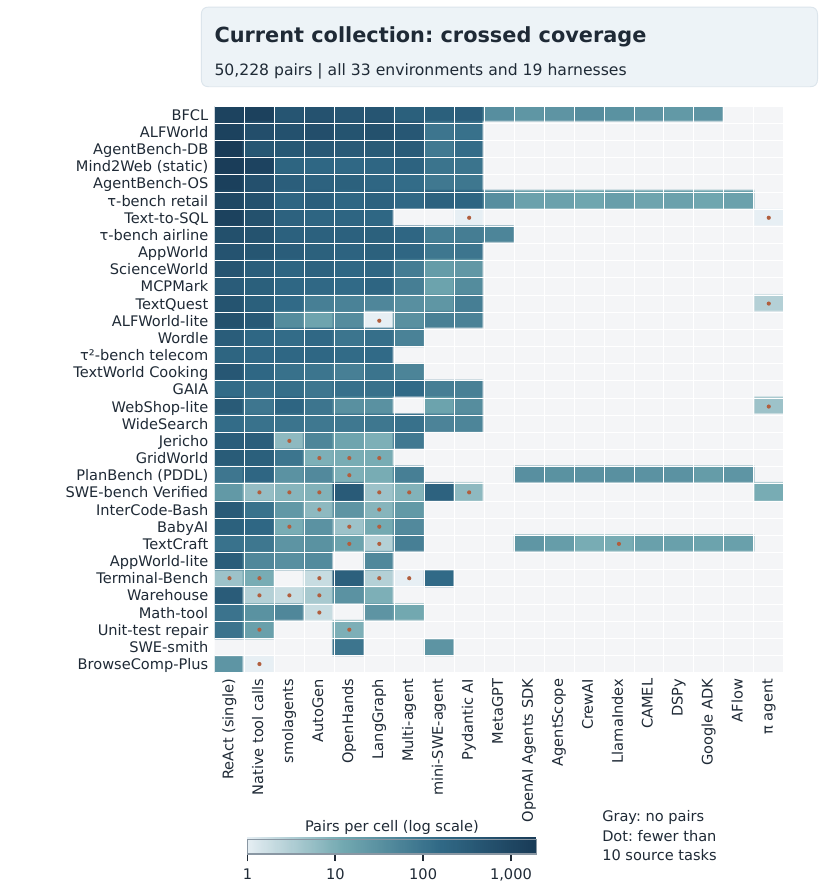}
\caption{\textbf{Where the current collection has support.} Rows and columns include all
$33$ environments and $19$ harness families, ordered by pair count.
The logarithmic color scale exposes both large and small cells. Gray cells contain no pairs;
they do not establish that the combination is unsupported by the harness.}
\label{fig:collection-crossed}
\end{figure}

We recompute Figure~\ref{fig:collection-crossed} from the same $50{,}228$-pair
metadata index as Figure~\ref{fig:corpus}. Color shows pair counts; dots identify occupied
cells with fewer than ten distinct source tasks. Of $280$
occupied environment--harness cells, $244$ meet this task-support threshold.

\begin{resulttakeaway}{Breadth does not imply a balanced comparison}
The collection covers many system configurations, with uneven support across combinations.
Comparisons of models or harnesses need shared tasks and successful-rollout denominators;
the number of collected failure pairs alone cannot rank their reliability.
\end{resulttakeaway}

\clearpage
\subsection{Historical error profiles by model and environment}
\label{app:restored-error-profiles}
\begin{figure}[H]\centering
\includegraphics[width=\linewidth]{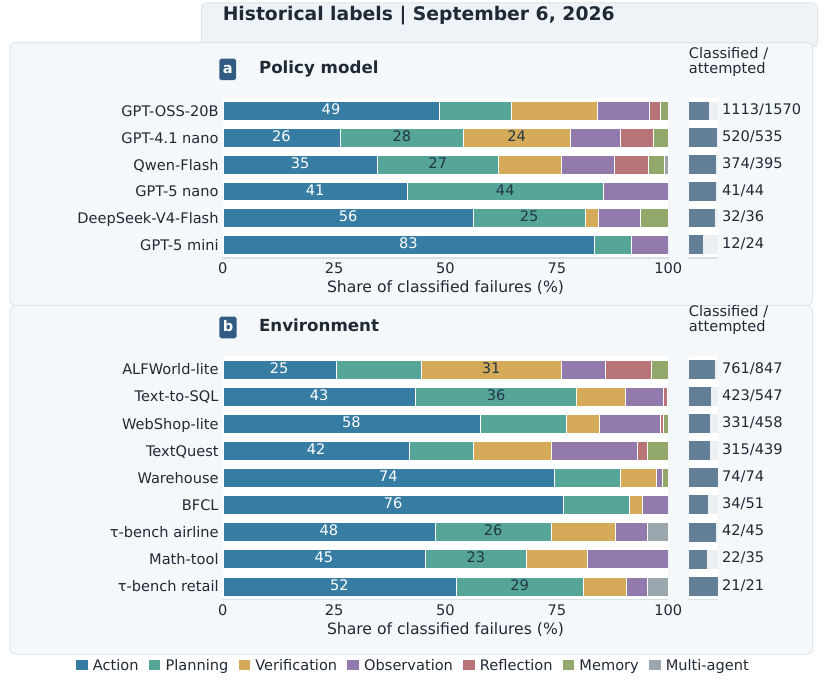}
\caption{\textbf{The same historical labels, viewed along two axes.}
Bars normalize by classified trajectories; adjacent counts retain abstentions in the denominator.
We display groups with at least $20$ attempted labels. All model groups meet
that floor; the environment panel omits $87$ attempts in smaller groups.
This selected cohort includes simplified and excluded environments. Different task mixtures
and selective abstention prevent model rankings or estimates of causal environment effects.}
\label{fig:historical-profiles}
\end{figure}

The September 6, 2026 snapshot contains $2{,}604$ distinct source trajectories:
machine labeling classified $2{,}092$ and returned abstentions for the rest.
We recount each trajectory once and retain the original codebook.
Figure~\ref{fig:historical-profiles} complements the harness profiles in
Figure~\ref{fig:taxonomy-harness} with model and environment breakdowns.

\begin{resulttakeaway}{Task-specific checks deserve separate error categories}
In the classified ALFWorld-lite traces, verification is the largest category
($239/761$); action and planning together account for
$336/423$ Text-to-SQL labels.
These contrasting profiles motivate retaining task context when designing diagnosis
rubrics and repair checks. They describe this historical sample, not the full collection.
\end{resulttakeaway}
Models in panel (a) generated the trajectories; they need not be the labeling judges.
Appendix~\ref{app:error-taxonomy} reports the separate judge-agreement study and its limitations.

\clearpage
\subsection{Observable symptoms with successful-rollout denominators}
\label{app:restored-observable-symptoms}
The September 5, 2026 development atlas records $3{,}596$ executions on
$200$ source tasks across five environments, nine policies and two harnesses.
It retains $2{,}774$ successes and $822$ failures;
$4$ of the $3{,}600$ planned executions are missing.
We recount the stored outcome and phenotype rows, without rerunning the environments
or relabeling the underlying traces.

\begin{figure}[H]\centering
\includegraphics[width=\linewidth]{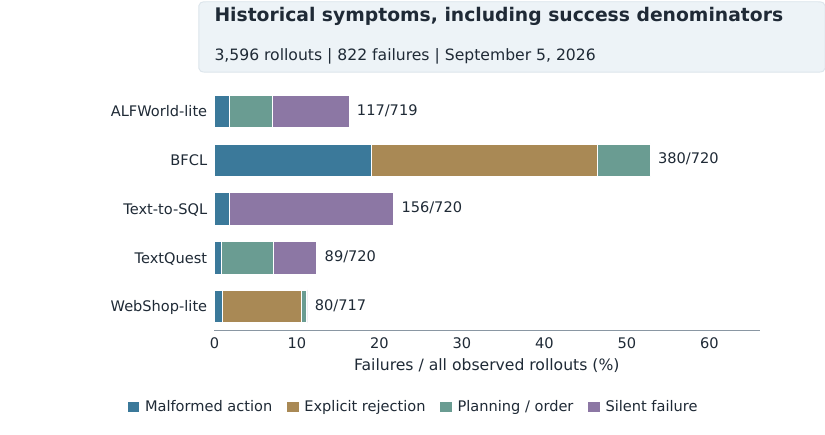}
\caption{\textbf{Failure symptoms as a share of all observed executions.}
Each failed execution contributes to one mechanical category; successful executions
remain in the denominator. End labels show failed/observed counts.
The historical harness implementations predate the typed-tool-schema and native-stop
fixes, so this figure describes that development cohort, not current harness quality.}
\label{fig:historical-symptoms}
\end{figure}
\begin{resulttakeaway}{Outcome checks expose failures that action validity misses}
Of the $822$ failed executions, $248$ include a completion action
that fails the task verifier. These traces require checking whether the agent met the
goal, beyond whether it issued an executable action. AED retains the outcome alongside
the actions and observations needed to investigate the discrepancy.
\end{resulttakeaway}
The mechanical categories distinguish malformed actions ($176$), explicit
rejections ($265$), planning/order failures ($133$) and silent failures
($248$). The rules include budget exhaustion under planning/order and certain
unsuccessful terminal actions under rejection. These are observable symptoms, not validated
root causes. We keep this rollout cohort separate from the judge-labeled profiles above:
the latter contain selected failures and cannot supply success denominators.
\FloatBarrier

\clearpage
\subsection{Complete distributions behind Figure~\ref{fig:corpus}}
\label{app:corpus-full-distribution}
The two panels below expand every category in the same $50,228$-pair
collection as Figure~\ref{fig:corpus}. No category is omitted or merged here.
Rows are ordered by pair count; shares describe collection composition, not failure rates.
\begin{table}[H]\centering\small
\caption{\textbf{Complete Figure~\ref{fig:corpus}(a): all 33 environments.}
A dagger marks a category included in an \emph{Other} group in the main figure.
Each share uses all $50,228$ pairs as its denominator.}
\label{tab:corpus-full-environments}
\renewcommand{\arraystretch}{1.12}
\setlength{\tabcolsep}{7pt}
\begin{tabular}{@{}rlrrr@{}}\toprule
Rank & Environment & Pairs & Share (\%) & Source tasks \\
\midrule
1 & \textcolor[HTML]{D47C84}{\rule{5pt}{5pt}}\enspace{}BFCL & 5,521 & 10.99 & 470 \\
2 & \textcolor[HTML]{4286A5}{\rule{5pt}{5pt}}\enspace{}ALFWorld & 4,838 & 9.63 & 1,431 \\
3 & \textcolor[HTML]{45A597}{\rule{5pt}{5pt}}\enspace{}AgentBench-DB & 4,407 & 8.77 & 1,120 \\
4 & \textcolor[HTML]{83AF75}{\rule{5pt}{5pt}}\enspace{}Mind2Web (static) & 4,050 & 8.06 & 527 \\
5 & \textcolor[HTML]{6B8EC1}{\rule{5pt}{5pt}}\enspace{}AgentBench-OS & 3,309 & 6.59 & 485 \\
6 & \textcolor[HTML]{598E83}{\rule{5pt}{5pt}}\enspace{}$\tau$-bench retail & 3,134 & 6.24 & 324 \\
7 & \textcolor[HTML]{BFA05B}{\rule{5pt}{5pt}}\enspace{}Text-to-SQL & 2,684 & 5.34 & 336 \\
8 & \textcolor[HTML]{A9AB69}{\rule{5pt}{5pt}}\enspace{}$\tau$-bench airline & 2,498 & 4.97 & 26 \\
9 & \textcolor[HTML]{6BA6AE}{\rule{5pt}{5pt}}\enspace{}AppWorld$^{\dagger}$ & 2,345 & 4.67 & 89 \\
10 & \textcolor[HTML]{9EBB53}{\rule{5pt}{5pt}}\enspace{}ScienceWorld$^{\dagger}$ & 1,655 & 3.29 & 334 \\
11 & \textcolor[HTML]{818FB8}{\rule{5pt}{5pt}}\enspace{}MCPMark$^{\dagger}$ & 1,402 & 2.79 & 29 \\
12 & \textcolor[HTML]{DFAC5A}{\rule{5pt}{5pt}}\enspace{}TextQuest$^{\dagger}$ & 1,217 & 2.42 & 644 \\
13 & \textcolor[HTML]{8774B3}{\rule{5pt}{5pt}}\enspace{}ALFWorld-lite$^{\dagger}$ & 1,198 & 2.39 & 618 \\
14 & \textcolor[HTML]{75A69B}{\rule{5pt}{5pt}}\enspace{}Wordle$^{\dagger}$ & 1,078 & 2.15 & 91 \\
15 & \textcolor[HTML]{6F9F75}{\rule{5pt}{5pt}}\enspace{}$\tau^2$-bench telecom$^{\dagger}$ & 1,030 & 2.05 & 36 \\
16 & \textcolor[HTML]{91A875}{\rule{5pt}{5pt}}\enspace{}TextWorld Cooking$^{\dagger}$ & 1,020 & 2.03 & 84 \\
17 & \textcolor[HTML]{B69BAC}{\rule{5pt}{5pt}}\enspace{}GAIA$^{\dagger}$ & 1,013 & 2.02 & 78 \\
18 & \textcolor[HTML]{D68C5E}{\rule{5pt}{5pt}}\enspace{}WebShop-lite$^{\dagger}$ & 857 & 1.71 & 464 \\
19 & \textcolor[HTML]{7C9BAF}{\rule{5pt}{5pt}}\enspace{}WideSearch$^{\dagger}$ & 843 & 1.68 & 143 \\
20 & \textcolor[HTML]{AA819D}{\rule{5pt}{5pt}}\enspace{}Jericho$^{\dagger}$ & 748 & 1.49 & 47 \\
21 & \textcolor[HTML]{B18CC0}{\rule{5pt}{5pt}}\enspace{}GridWorld$^{\dagger}$ & 718 & 1.43 & 428 \\
22 & \textcolor[HTML]{AC8D60}{\rule{5pt}{5pt}}\enspace{}PlanBench (PDDL)$^{\dagger}$ & 675 & 1.34 & 163 \\
23 & \textcolor[HTML]{576EA6}{\rule{5pt}{5pt}}\enspace{}SWE-bench Verified$^{\dagger}$ & 639 & 1.27 & 228 \\
24 & \textcolor[HTML]{B97464}{\rule{5pt}{5pt}}\enspace{}InterCode-Bash$^{\dagger}$ & 555 & 1.10 & 315 \\
25 & \textcolor[HTML]{BB927D}{\rule{5pt}{5pt}}\enspace{}BabyAI$^{\dagger}$ & 529 & 1.05 & 195 \\
26 & \textcolor[HTML]{7A9EB3}{\rule{5pt}{5pt}}\enspace{}TextCraft$^{\dagger}$ & 486 & 0.97 & 192 \\
27 & \textcolor[HTML]{779FA0}{\rule{5pt}{5pt}}\enspace{}AppWorld-lite$^{\dagger}$ & 479 & 0.95 & 470 \\
28 & \textcolor[HTML]{AE737E}{\rule{5pt}{5pt}}\enspace{}Terminal-Bench$^{\dagger}$ & 425 & 0.85 & 62 \\
29 & \textcolor[HTML]{A69B7A}{\rule{5pt}{5pt}}\enspace{}Warehouse$^{\dagger}$ & 363 & 0.72 & 181 \\
30 & \textcolor[HTML]{9A8DAB}{\rule{5pt}{5pt}}\enspace{}Math-tool$^{\dagger}$ & 226 & 0.45 & 220 \\
31 & \textcolor[HTML]{AA8567}{\rule{5pt}{5pt}}\enspace{}Unit-test repair$^{\dagger}$ & 132 & 0.26 & 37 \\
32 & \textcolor[HTML]{718C68}{\rule{5pt}{5pt}}\enspace{}SWE-smith$^{\dagger}$ & 126 & 0.25 & 67 \\
33 & \textcolor[HTML]{668FA1}{\rule{5pt}{5pt}}\enspace{}BrowseComp-Plus$^{\dagger}$ & 28 & 0.06 & 27 \\
\midrule
& \textbf{All 33 environments} & \textbf{50,228} & \textbf{100.00} & 9,961 \\
\bottomrule\end{tabular}
\par\smallskip\begin{minipage}{\linewidth}\footnotesize\raggedright
$\dagger$~Expanded \emph{Other 25 envs}: 19,787 pairs
(39.39\% of the collection), already included in the rows above.
Percentages are rounded independently.
Colors preserve environment identities from the Figure~\ref{fig:corpus} palette.
Environment types and frozen-release membership are listed separately in Table~\ref{tab:env-types}.
\end{minipage}\end{table}
\clearpage
\begin{table}[H]\centering\small
\caption{\textbf{Complete Figure~\ref{fig:corpus}(b): all 19 harnesses.}
A dagger marks a category included in an \emph{Other} group in the main figure.
Each share uses all $50,228$ pairs as its denominator.}
\label{tab:corpus-full-harnesses}
\renewcommand{\arraystretch}{1.12}
\setlength{\tabcolsep}{7pt}
\begin{tabular}{@{}rlrrr@{}}\toprule
Rank & Harness family & Pairs & Share (\%) & Source tasks \\
\midrule
1 & \textcolor[HTML]{647FA7}{\rule{5pt}{5pt}}\enspace{}ReAct (single) & 16,646 & 33.14 & 4,622 \\
2 & \textcolor[HTML]{647FA7}{\rule{5pt}{5pt}}\enspace{}Native tool calls & 9,614 & 19.14 & 3,231 \\
3 & \textcolor[HTML]{647FA7}{\rule{5pt}{5pt}}\enspace{}smolagents & 4,614 & 9.19 & 1,899 \\
4 & \textcolor[HTML]{647FA7}{\rule{5pt}{5pt}}\enspace{}AutoGen & 4,368 & 8.70 & 1,610 \\
5 & \textcolor[HTML]{647FA7}{\rule{5pt}{5pt}}\enspace{}OpenHands & 4,329 & 8.62 & 1,651 \\
6 & \textcolor[HTML]{647FA7}{\rule{5pt}{5pt}}\enspace{}LangGraph & 3,873 & 7.71 & 1,406 \\
7 & \textcolor[HTML]{647FA7}{\rule{5pt}{5pt}}\enspace{}Multi-agent (planner--executor) & 2,797 & 5.57 & 1,219 \\
8 & \textcolor[HTML]{647FA7}{\rule{5pt}{5pt}}\enspace{}mini-SWE-agent & 1,695 & 3.37 & 778 \\
9 & \textcolor[HTML]{647FA7}{\rule{5pt}{5pt}}\enspace{}Pydantic AI$^{\dagger}$ & 1,441 & 2.87 & 758 \\
10 & \textcolor[HTML]{647FA7}{\rule{5pt}{5pt}}\enspace{}MetaGPT$^{\dagger}$ & 124 & 0.25 & 54 \\
11 & \textcolor[HTML]{647FA7}{\rule{5pt}{5pt}}\enspace{}OpenAI Agents SDK$^{\dagger}$ & 103 & 0.21 & 69 \\
12 & \textcolor[HTML]{647FA7}{\rule{5pt}{5pt}}\enspace{}AgentScope$^{\dagger}$ & 96 & 0.19 & 61 \\
13 & \textcolor[HTML]{647FA7}{\rule{5pt}{5pt}}\enspace{}CrewAI$^{\dagger}$ & 94 & 0.19 & 58 \\
14 & \textcolor[HTML]{647FA7}{\rule{5pt}{5pt}}\enspace{}LlamaIndex$^{\dagger}$ & 94 & 0.19 & 62 \\
15 & \textcolor[HTML]{647FA7}{\rule{5pt}{5pt}}\enspace{}CAMEL$^{\dagger}$ & 91 & 0.18 & 58 \\
16 & \textcolor[HTML]{647FA7}{\rule{5pt}{5pt}}\enspace{}DSPy$^{\dagger}$ & 88 & 0.18 & 59 \\
17 & \textcolor[HTML]{647FA7}{\rule{5pt}{5pt}}\enspace{}Google ADK$^{\dagger}$ & 79 & 0.16 & 58 \\
18 & \textcolor[HTML]{647FA7}{\rule{5pt}{5pt}}\enspace{}AFlow$^{\dagger}$ & 62 & 0.12 & 41 \\
19 & \textcolor[HTML]{647FA7}{\rule{5pt}{5pt}}\enspace{}$\pi$ agent$^{\dagger}$ & 20 & 0.04 & 20 \\
\midrule
& \textbf{All 19 harnesses} & \textbf{50,228} & \textbf{100.00} & 9,961 \\
\bottomrule\end{tabular}
\par\smallskip\begin{minipage}{\linewidth}\footnotesize\raggedright
$\dagger$~Expanded \emph{Other 11 harnesses}: 2,292 pairs
(4.56\% of the collection), already included in the rows above.
Percentages are rounded independently.
Source tasks are distinct within each harness. The same task can occur under several
harnesses; the total is the collection's 9,961 unique environment--task keys,
not the sum of this column.
\end{minipage}\end{table}

\clearpage
\section{Evidence grades, counting and populations}
\label{app:evidence-counting}
\subsection{Interpreting execution evidence}
\label{sec:algorithm}

Evidence grades let a user select diagnosis-only records, observed recoveries or repeated
matched contrasts. A diagnosis whose citations cannot be
resolved is \textbf{ungrounded}. A grounded diagnosis with neither replay nor a named checkpoint
is \textbf{E0}. Explicit replay-status fields also distinguish a named but unexecuted checkpoint
from a completed trial; Appendix~\ref{app:adaptive-k} documents the legacy grading convention.
Executed replacements that never succeed or succeed on only some trials are recorded as
\textbf{no recovery} or \textbf{partial recovery}. \textbf{E1} records an observed recovery on the
replacement branch when the original action was not tested adequately or also succeeded.
\textbf{E2} requires at least two trials per
arm and completion of both arms, with every treatment continuation passing and every matched
control failing.

A successful trial establishes observed recovery, not a unique root cause or a reliable effect.
Even repeated contrasts depend on the restored state, policy, harness, and verifier. Grades can
be recomputed from the recorded outcomes without another environment run.
Appendix~\ref{app:certify} gives the grading procedure and assumptions.

\subsection{Corpus scope and counting}
\label{sec:scope}

The release contains failures that arose during agent execution. Imported natural failures can
contribute diagnosis records even when their original environment cannot be restored. We list
procedural stand-ins separately from the benchmarks they approximate (Appendix~\ref{app:data}).
Task-level split assignments group alternate diagnoses and replay variants together; the
component-disjoint evaluation additionally separates the leakage groups defined at the freeze.
We treat oversized category families separately (Appendix~\ref{app:exposure}).

We count an \emph{error--diagnosis pair} when a natural failure has a diagnosis that names a
candidate error step, explains it, and provides at least one quotation grounded in the failed
trace. This count does not require successful replay. Records with an executed, passing correction
form a smaller subset, which we report with their control outcomes and replay settings, so
collection counts, diagnosis counts, and replay counts stay separate.
\paragraph{Current collection index.}
Figure~\ref{fig:corpus}, Table~\ref{tab:env-types} and
Appendix~\ref{app:collection-analysis} use the same source-linked pairs.
Breadth counts require at least ten distinct source tasks per environment,
harness family or policy model. A different policy, seed or sampling setting
can produce another failed execution of the same task; an additional diagnosis
of one execution adds a pair, not a new trajectory. The frozen diagnosis release
imposes separate environment, exposure and split checks.
Table~\ref{tab:population-map} separates configuration inventory, annotation attempts
and frozen training rows. Citation admission tests trace support, not semantic correctness.
Retrieval failures affect $440$ of the $452$ GAIA and WideSearch rows: holdout sensitivity
uses the $835$ rows left after removing the $108$ affected cases
(Appendix~\ref{app:infra-scan}).

\begin{table}[t]
\centering\small
\setlength{\tabcolsep}{4pt}
\caption{The populations answer different questions; attempt, configuration and source-task
counts are not interchangeable, and the annotation audit does not enlarge any training cohort.}
\label{tab:population-map}
\begin{tabular}{@{}p{2.65cm}p{4.7cm}p{5.4cm}@{}}
\toprule
Population & Counting unit and size & Use in this paper \\
\midrule
Current collection & $50{,}228$ pairs over $38{,}278$ source blobs and $9{,}961$ source tasks & Source-linked collection; separate from the frozen training release \\
Collection inventory & $15{,}338$ configuration records with steps and a trace reference & Collection coverage; not an admitted training split \\
Panel annotation audit & $11{,}116$ attempts; $6{,}381$ pass citation grounding & Annotation validity under automated checks \\
Frozen diagnosis release & $4{,}319$ rows over $2{,}309$ source tasks in $15$ environments (train, development, holdout); $418$ more rows over $121$ previously exposed tasks ship as an exploratory split & The evaluated dataset; a separate, earlier snapshot \\
Pilot diagnosis split & $359$ train, $201$ development, $169$ holdout rows & Source for the three-seed pilot ($168$ shared source tasks per arm) \\
Final paired cohort & $948$ tasks in $133$ families of the $3157$-row training split ($1678$ tasks); $943$ held-out rows over $496$ tasks & Final-scale comparison (Table~\ref{tab:training-completed}) \\
Paired replay & $3062$ attempts over $1284$ source tasks, both arms from one checkpoint & Correction validity (Table~\ref{tab:repair}) \\
\bottomrule
\end{tabular}
\end{table}

\subsection{Notes on the resource comparison}
\label{app:rw-table-notes}

Who\&When annotates 184 natural multi-agent failures, its three annotators spending
a combined 84.3 hours on decisive steps when no fault was injected~\citep{whoandwhen}.
In AgenTracer's released v1.0.0 data, $1306$ of the $3208$ training rows carry an
injection label; its agentic split's responsible-agent field has one value and
therefore no within-split variation~\citep{agentracer}.

\begin{artifactbox}{Reading Table~\ref{tab:labels}}
\small
nat./inj./syn.: natural, injected, teacher-synthesised failures.
Ctrl. requires the original action re-executed from the same checkpoint.
\cmark{} supported; \xmark{} absent; \pmark{} partial; ``--'' not stated.
$^\ddagger$Full multimodal resource.
Counts use reported units. AED's collection row uses the same current index as
Figure~\ref{fig:corpus}; Table~\ref{tab:env-types} identifies the environment types.
\dataset{} represents overall capabilities across constituent datasets, with
subset-specific training and replay. Subcohort counts below are not additive.
\end{artifactbox}

{\centering\scriptsize
\setlength{\tabcolsep}{2.4pt}
\resizebox{\linewidth}{!}{\begin{tabular}{@{}l l r r r r c c c c c l@{}}
\toprule
Work & Errors & Envs & Harn. & Models & Count & Step & Evid. & Exec.\ fix & Ctrl. & Train & Agree. \\
\midrule
\dataset{} diagnosis release (ours) & nat. & 15 & 9 & 10 & 4{,}319 & \cmark & \cmark & \xmark & \xmark & D & -- \\
\dataset{} replay study (ours) & nat. & 11 & -- & -- & 3{,}062 & \cmark & \cmark & \cmark & \cmark & -- & -- \\
\bottomrule
\end{tabular}}\par}

Envs/Harn./Models: environments, harnesses (including multi-agent frameworks)
and generating policies; \dataset{} requires ten source tasks per entry
(Section~\ref{sec:method}). Table~\ref{tab:labels} uses AED's audited count of
error--diagnosis pairs; repeated
diagnoses of one execution do not add independent failures.

The latest audit links $50{,}228$ structurally supported pairs across $33$ typed environments, $19$ harness families and $23$ named policy models.
These breadth counts use one pair index and require ten source tasks per category;
model aliases share one identity. The checks verify source linkage, the recorded
error step and quoted trace support, not semantic correctness or training admission.
For the completion batch, we re-diagnosed source-verified stored failures with a
strict-citation single judge, without an added ground-truth reference or replay. We retained
one supported new diagnosis per previously unpaired source blob and preserved
rejected attempts in the audit. These rows augment the collection, not the frozen
training or evaluation sets. A deterministic, environment-stratified spot-check
found semantic defects beyond citation support; we quarantined flagged records.
This defect-finding check does not estimate accuracy or validate the remaining collection.
These are collection environments, including
synthetic tasks and simplified ports, not a count of independent public benchmarks.

Other reported units include TRAIL errors, MAST traces, diagnosis rows and paired replay attempts.
Step includes spans.
Agree.: reported human agreement (raw rates or chance-corrected coefficients), comparing
programmatic labels with humans for AEGIS and failure-mode labels for MAST.
TrajErrBench reports Fleiss' $\kappa=.91/.67$ for critical-error \emph{steps}
on its $\tau^2$-Bench / SWE-Bench Pro subsets, respectively
(\citealp{trajdebug}, Table~5).
AED reports raw preferred-step agreement of $\humanPreferredStepAgree\%$
($59/\humanPreferredStepPairs$) under shared AI assistance. Both reviewers select
a location on these $\humanPreferredStepPairs$ of $80$ historical records,
including preferred choices from multiple candidates. The single-step subset
has $\humanStepAgree\%$ agreement ($46/\humanStepPairs$;
Appendix~\ref{app:human-results}). Raw rates and chance-corrected coefficients
are different statistics; annotation targets, sampling and assistance also differ.
The agreement column does not rank resources.

Partial Exec.\ fix covers initial-state reruns or system interventions; AgenTracer's
oracle replay is construction-only. \dataset{} marks cover its replayable collection
subset; $0$ of its $4{,}319$ diagnosis rows join a replay record.
Who\&When Pro uses its full text/image/video resource. Its authors name GPT-4.1
and Gemini 3 Flash as primary backbones, not an exhaustive model count.
Its benchmark and our text-only environment counts have different scopes.

\paragraph{Comparison with AEGIS.}
\citet{aegis} use successful executions and controlled fault injection to obtain
agent/error attribution labels, then verify that the altered execution fails.
This supplies a known intervention and a matched successful baseline; natural
failures require separate checks of diagnosis quality.
Their agent/error-set F1 measures a different target from our exact-step accuracy.
We therefore do not import their reported scores into our localization tables or
claim a matched training advantage over AEGIS.
Such a resource comparison would require a common attribution target, compatible
labels, the same base model and training budget, and an independent test set.
AEGIS's failed trajectories and AED's error--diagnosis pairs also use different
counting units. In Table~\ref{tab:labels}, Exec.\ fix and Ctrl. concern correction
and original-action replay, not the execution checks used to validate fault injection.

\section{Construction, Replay, and Evidence}
\label{app:certify}
\label{app:assumptions}
\label{app:admission}
\paragraph{Implementation and prior work.}
AgentDebug~\citep{agentdebug} motivates diagnosis-guided recovery.
Our diagnosis bridge calls AgentDebugX's Python attribution API~\citep{agentdebugx}
for action blame, evidence and proposed corrections.
AED provides environment adapters, evidence checks and verifier-backed paired replay.
Records distinguish diagnosis engines and retain available library provenance.
CUADebug~\citep{cuadebug} studies diagnosis and repair for screenshot-based
computer-use agents; our collection remains text-only.

\paragraph{Verifier validation.}
Reference-solution tests identified two verifier problems in the imported environments. In
AgentBench-DB, gold SQL reproduced the stored table hash for only $221$ of
$455$ state-changing tasks, and in BFCL the ground-truth call list failed the state
checker on $112$ of $386$ instances. Episodes on the tasks these checks flagged are quarantined; the
checks did not cover every task, and the release still holds rows on tasks whose gold answer
fails the grader (Appendix~\ref{app:infra-scan}). These checks test the environment ports before their outputs are used as failure
labels; they do not assess the debugger's explanations.

\paragraph{Historical replay-cohort settings.}
The replay cohort used an unguided all-at-once debugger. Reference-assisted variants are
recorded separately; a same-task success exists for $39.5\%$ of the failed rollouts in the
reference-availability census. Paired replay uses temperature 0.2 for the first sample and 0.8
for subsequent samples, compared with 0.7 for the source rollout. The two replay arms share this
schedule, but replay and source-rollout outcomes need not follow the same sampling distribution.

\subsection{Construction algorithm}
\label{app:method-details}
\begin{artifactbox}{AET stage contracts}
\renewcommand{\arraystretch}{1.12}
\begin{tabular}{@{}P{.16\linewidth}P{.78\linewidth}@{}}
\toprule
Stage & Stored information and admission boundary \\
\midrule
1. Collect & Adapters supply tasks, instructions, tools and success verifiers.
Save policy, harness and environment identities, acting units, actions,
observations and final outcomes, including successes. Failures arise naturally;
we do not inject them. Preserve service outages, harness failures and unusable
graders for audits, but exclude them from agent-error targets. Collection
requires no replay support. \\
\end{tabular}\par\smallskip
\begin{tabular}{@{}P{.16\linewidth}P{.78\linewidth}@{}}
2. Diagnose & Following AgentDebug~\citep{agentdebug}, propose an error step,
responsible agent or component, explanation with trace quotations, and replacement
action. Record privileged reference or verifier information separately from
learner inputs. Where supported, replay feedback guides revisions within a fixed
budget. Keep every attempt and distinguish the first proposal from one selected
after retries. Multiple debuggers can reuse a stored failure. \\
\end{tabular}\par\smallskip
\begin{tabular}{@{}P{.16\linewidth}P{.78\linewidth}@{}}
3. Ground & Resolve cited steps, acting units and quotations in the exported
student input. Check tool validity and information availability. Semantic judges
assess the attribution and repair; both judges must accept in the dual-judge
subset (Appendix~\ref{app:quality-rubric}). Save accepted and rejected decisions.
These are machine judgments; human validity and training eligibility require
separate assessment. \\
\end{tabular}\par\smallskip
\begin{tabular}{@{}P{.16\linewidth}P{.78\linewidth}@{}}
4. Replay & Start two new executions from a restored or reconstructed checkpoint:
one substitutes the correction and one repeats the original action. Match policy,
harness, budget, verifier and temperature schedule; save seeds, sample counts,
checkpoint fidelity and both outcomes, including failed repairs and passing
controls. The failed source rollout does not serve as this control. Recovery
supports the correction in that setting, not necessarily its explanation or a
unique root cause. Coached retries use a separate protocol. \\
\end{tabular}\par\smallskip
\begin{tabular}{@{}P{.16\linewidth}P{.78\linewidth}@{}}
5. Build views & Diagnosis targets use supported attributions without requiring
replay. Recovery targets use an executed, passing continuation and preserve its
branch identity. Only eligible continuation actions receive loss; input history,
tool observations and rejected continuation actions receive no loss. Preferences
also require a valid action comparison under the same visible input. Source-task
grouping, declared splits and content checks determine training eligibility
separately from label quality. Retain records excluded from a view and their
selection history. Derive evidence grades from stored outcomes. \\
\bottomrule
\end{tabular}
\end{artifactbox}

\paragraph{Failure outcomes and attributable errors.}
\label{app:failure-definition}
Adapters evaluate task completion using answer or result matching, executable
tests, or goal-state checks. We distinguish a recorded negative outcome from
a valid task-failure verdict and from a supported attribution.

\begin{artifactbox}{Failure boundaries}
\renewcommand{\arraystretch}{1.12}
\begin{tabular}{@{}P{.26\linewidth}P{.68\linewidth}@{}}
\toprule
Observed case & Interpretation and admission rule \\
\midrule
Wrong answer, failed tests or unmet goal & A valid negative task verdict supplies
a diagnosis candidate; it does not identify the erroneous decision. \\
\addlinespace[3pt]
Tool error followed by task success & Exclude from the failed-run cohort.
An intermediate exception need not cause task failure. \\
\addlinespace[3pt]
Budget exhaustion, loop or early stop & A candidate under a defensible execution
budget. Inspect where progress broke down; do not blame the cutoff step by default. \\
\addlinespace[3pt]
Invalid tool action & A candidate if the agent had a valid tool contract.
Schema bugs, service outages and environment crashes require separate accounting. \\
\addlinespace[3pt]
No usable verifier verdict & Quarantine for agent-error supervision.
An unavailable reference or grader exception cannot establish task failure. \\
\addlinespace[3pt]
Recorded-page action mismatch & Establishes deviation from the recorded path,
not inability to recover in a live browser; exclude from strict admission. \\
\bottomrule
\end{tabular}
\end{artifactbox}

For attribution, reviewers seek the earliest \emph{supported} consequential
mistake with a feasible alternative, not a later symptom or a unique global
root cause. They can use the resulting observation as retrospective evidence;
the proposed action must respect information available before execution
(Appendix~\ref{app:quality-rubric}). Unsupported attributions fail semantic admission.

The collection gate checks recorded failure flags, source linkage, step presence
and trace quotations. Automated screening recognizes known non-agent signatures;
it does not cover every adapter-specific ungradable outcome. Thus structural
collection counts do not certify semantic validity or training eligibility.
Recovery targets additionally require an executed, passing continuation.

\subsection{Evidence ladder and statistical claims}
\label{app:adaptive-k}
\begin{center}\small
\begin{tabular}{@{}P{.21\linewidth}P{.73\linewidth}@{}}
\toprule
Evidence state & Required observation \\
\midrule
Ungrounded & Citation support is absent, regardless of replay outcome. \\
E0 & Grounded diagnosis; replay was not attempted and no checkpoint was named. \\
No / partial recovery & Executed treatment never succeeds / has mixed outcomes. A named but unexecuted checkpoint also receives the legacy no-recovery label; typed replay status distinguishes it from a failed trial. \\
E2 & All treatment trials pass, all controls fail, at least two trials per arm, and both arms complete their planned counts. \\
\bottomrule
\end{tabular}
\end{center}
E1 records recovery without the E2 matched-contrast requirements. Ladder grades,
the schema's historical certificate flag and multiplicity-adjusted certificates
are separate fields; none establishes unique causation or global minimality.

For executed arms, report $\hat p_T=s_T/n_T$, $\hat p_C=s_C/n_C$ and
$\widehat\Delta=\hat p_T-\hat p_C$ with both denominators. Unexecuted controls
remain null. Tag structural outcomes separately from Monte Carlo trials.
Confirmatory runs freeze sample counts and continuation conditions; adaptive
supply runs retain their stopping rule and receive a separate analysis.

\subsection{Lineage and checkpoint fidelity}
\begin{center}\small
\begin{tabular}{@{}P{.21\linewidth}P{.73\linewidth}@{}}
\toprule
Check & Requirement and interpretation \\
\midrule
Branch identity & Recompute the v1 attestation of source trace, pre-action prompt, restored state, trial, applied action and actor transcript. \\
Native messages & Match tool-call identifiers and result ordering. Keep reconstructed external logs in a separate fidelity stratum. \\
Action authorship & Attribute an action to the judge only for substitution. A policy's coached continuation is not the judge's proposed action. \\
Exclusions & Reject a training package when invalid-lineage exclusions exceed the implemented $5\%$ ceiling. This controls attrition, not semantic accuracy. \\
Family references & An imported ALFWorld family match is not an executed success on the target instance. The probe-to-judge prompt still needs family-scope propagation; keep those results separate from exact-task reference effects. \\
\bottomrule
\end{tabular}
\end{center}
\label{app:certify:end}

\section{Construction Audit and Annotation Quality}
\label{app:audit}
\subsection{Infrastructure and task-contract defects}
\label{app:infra-scan}
Frozen inputs retain the defects below because training preceded the audits.
Later exporter fixes do not retroactively repair the evaluated release.
The retrieval flags cover GAIA and WideSearch only; contract omissions require
re-derivation from the teacher prompt.

\begin{center}\small
\begin{tabular}{@{}P{.19\linewidth}P{.75\linewidth}@{}}
\toprule
Retrieval audit & Recorded extent \\
\midrule
Affected rows & In $440$ of the $4{,}737$ rows ($307$ train, $25$ development, $108$ holdout, none in the exploratory split), every search/fetch call failed and none returned a result: $242$ GAIA rows and $198$ WideSearch rows. Holdout coverage is $52$ of $56$ and $56$ of $56$, respectively. \\
Targets & All $440$ targets assert an error; $396$ mention the backend in the explanation or evidence, and $62$ place the decisive step at the first failed call. \\
Score sensitivity & Report the full $943$ cases and the $835$ that remain after excluding the affected diagnoses. The influence of the $307$ training rows on checkpoints is unmeasured. \\
Training exposure & Paired arms carry $75$ of their $948$ tasks with this signature, the same $75$ in all four arm files, and the v14 scale arm $109$ of its $1{,}656$. \\
Later releases & The exporter now rejects this signature. Frozen v14 retains flagged rows, as does the v15 cut ($578$ of its $5{,}602$ rows, $145$ in its holdout); both ship the flag file. \\
\bottomrule
\end{tabular}
\end{center}

\begin{center}\small
\begin{tabular}{@{}P{.19\linewidth}P{.75\linewidth}@{}}
\toprule
Other defect & Evidence and effect on interpretation \\
\midrule
AgentBench-DB grader & Gold SQL fails to reproduce the table hash on $234$ of its $455$ state-changing tasks. The release carries $54$ rows on $47$ of them ($43$ training, $11$ held out), beside $34$ rows on tasks whose gold does reproduce and $338$ whose task identity postdates that census and is unchecked. These failures cannot validate agent-error labels. \\
Teacher contract & In AgentBench-DB, AgentBench-OS and AppWorld, $617$ of the $1{,}016$ rows were labelled before the submission contract entered the prompt, $100$ of them in the holdout. The split follows harness commit; shipped metadata lacks this flag. \\
Visible error cue & In Mind2Web, the first-tool-error rule scores $100\%$ on its $78$ held-out cases, where the base scores $96.15\%$. Removing them leaves base $42.77$ micro exact step over the remaining $865$ cases; the strongest prompt-only rule reaches $29.71$. Common cases preserve comparability, but this cue raises aggregate accuracy. \\
Incomplete goal field & In BFCL, $219$ of the release's $414$ BFCL rows ($34$ of $63$ held out) blame a later conversational turn than the stored first-turn goal. Later turns remain visible in the trace; the diagnosis concerns the whole conversation. \\
\bottomrule
\end{tabular}
\end{center}

\subsection{Construction inventory and counting ledger}
\label{app:construction-ledger}
The collection inventory counts configurations; the annotation audit counts debugging
attempts. Several diagnoses can refer to one failed run. Neither count measures
independent source tasks. Table~\ref{tab:population-map} identifies the populations
used here; Table~\ref{tab:env-types} identifies excluded environment types.
The 16 September 2026 inventory counts configurations, not independent trials;
repeated executions can share an identifier.
\begin{center}\small
\begin{tabular}{@{}P{.53\linewidth}rP{.25\linewidth}@{}}
\toprule
Inventory stage & Count & Unit \\
\midrule
At least one failure record & $31{,}903$ & Configurations \\
Quarantined success/failure identity conflict & $1{,}029$ & Configurations \\
After quarantine & $30{,}874$ & Configurations \\
Core failure inventory & $16{,}920$ & In fifteen core environments \\
Positive step count and trajectory reference & $15{,}338$ & Eligible configurations \\
Failed basic eligibility & $1{,}582$ & Configurations \\
Distinct source tasks among eligible records & $5{,}376$ & Tasks \\
With a diagnosis record & $10{,}057$ & Configurations \\
With step and explanation & $9{,}464$ & Structural screen only \\
\bottomrule
\end{tabular}
\end{center}
Structural completeness does not establish grounding, semantic validity or replay
success. The separate annotation audit covers $11{,}116$
diagnosis attempts; $6{,}381$ ($57.4\%$) pass citation grounding.
Those attempts cannot be appended to the configuration funnel.

\subsection{Current annotation quality audit}
\label{app:panel-quality}
Panel consensus is associated with stable attribution, but citation support can
still fail. The solo comparator also serves as the panel's arbiter, and the two
pipelines differ in implementation. Their agreement is not an independent
correctness standard or an isolated voting effect.

\begin{center}\small
\begin{tabular}{@{}P{.31\linewidth}P{.63\linewidth}@{}}
\toprule
Matched ALFWorld audit & Measured observation \\
\midrule
Attribution & $323$ paired ALFWorld traces; $297$ matched cells with a step and explanation from both constructions. Exact-step agreement is $46.8\%$ and within-one-step agreement $53.9\%$. \\
Vote concentration & Within one step: unanimous $76.6\%$, two-of-three $55.2\%$, three-way split $40.0\%$. \\
Citation admission & Panel $243/323$ ($75.2\%$); solo $285/323$ ($88.2\%$). Of the $80$ panel failures, $77$ lack a supporting quotation; the other three cannot ground cited text at the claimed event. \\
Quotation mechanism & No arbitrated output has an empty quotation; $78/213$ voter-selected outputs contain at least one. One empty field need not fail admission if another valid citation exists. \\
\bottomrule
\end{tabular}
\end{center}

\begin{center}\small
\begin{tabular}{@{}P{.31\linewidth}P{.63\linewidth}@{}}
\toprule
Separate audit & Observation and limit \\
\midrule
Environment association & Over $5{,}725$ diagnosed records, consensus is $89.9\%$ in single-decision groups versus $72.5\%$ in long-horizon groups. Models, lengths and tasks differ. \\
Full citation audit & $6{,}381$ of $11{,}116$ attempts pass. Empty quotations occur in $47\%$ of mid-tier records versus $10\%$ of stronger-arbiter records; the case sets differ. \\
Re-annotation probes & Across five groups, the solo model anchors a quotation on $159$ of $183$ records with an empty panel quotation ($86.9\%$). Quote recovery establishes neither admission nor semantic correctness. \\
Test--retest & $3470$ pairs: $2845$ consensus-admitted first verdicts repeat the exact step at $88.2\%$ and fall within one step at $94.1\%$; the $625$ not admitted reach $58.2\%$ and $73.4\%$. The exact-step gap is $29.9$ points. \\
Human review & The completed $80$-record AI-assisted audit, its sampling scope and agreement measures appear in Appendix~\ref{app:human-results}. It is separate from the machine audits above. \\
\bottomrule
\end{tabular}
\end{center}
Admission selects an existing member's verdict; it does not merge explanations.
Its association with test--retest stability does not show that gating causes
correctness. Citation grounding, reproducibility and human validity require
different evidence.

\begin{figure}[ht]
\centering
\includegraphics[width=\linewidth]{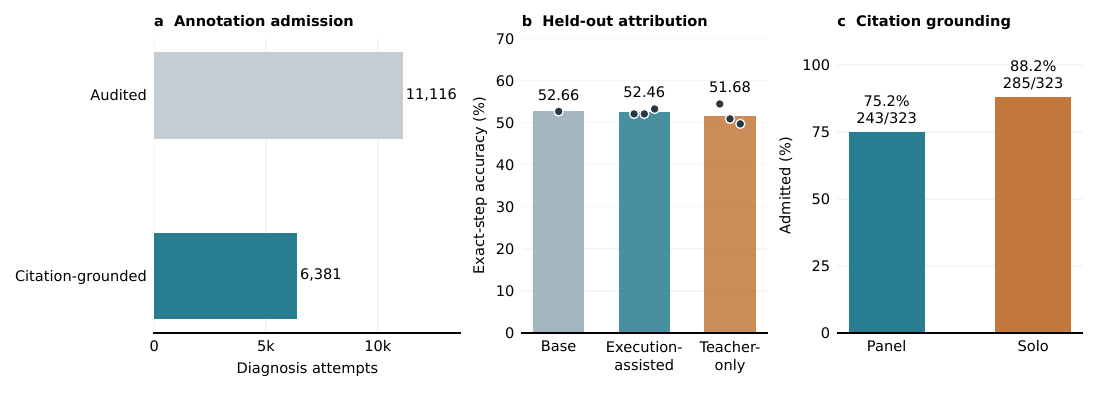}
\caption{Three separate measurements, not one pipeline: annotation admission, paired attribution
SFT on the pilot holdout, and panel versus solo citation grounding. Each panel has its own
denominator and the populations are not successive stages of one funnel.}
\label{fig:current-evidence}
\end{figure}
\subsection{Model-panel adjudication of the blinded semantic sample}
\label{app:semantic-panel}
The model panel accepts step attribution more often than full evidence support
(Table~\ref{tab:semantic-panel}). It finds the attributed step defensible or a
defensible alternative on $102$ of $120$ records; expert adjudication remains unrun.

\begin{center}\small
\begin{tabular}{@{}P{.20\linewidth}P{.74\linewidth}@{}}
\toprule
Audit condition & Protocol and limitation \\
\midrule
Sample and judges & $120$ blinded records, $30$ per stratum of horizon (single-decision / long-horizon) by construction (consensus / arbitrated-or-single). Three vendor families: GPT-6 Astra, Gemini 3.8 Flash and Claude Opus 5. \\
Visible input & The debugger's rendered failed trajectory, terminal outcome and diagnosis (step, unit, explanation, evidence and replacement). Replay outcomes, grades, label sources and annotators were hidden. Three fixed questions with one-sentence reasons; temperature $0$ where supported. \\
Completion & All records received three verdicts; spend by the hub's price table was \$$17.40$. \\
Shared model families & GPT-5.6 Sol wrote $57$ of the sampled diagnoses, Gemini 3.6 Flash $43$, Claude Haiku 4.5 $19$, GPT-5 mini $1$. Authored case sets differ, confounding the same-family comparison. \\
\bottomrule
\end{tabular}
\end{center}

Citation format helps explain the low evidence agreement: $33$ diagnoses carry
an evidence item that is a bare event identifier. The construction contract
accepts a resolvable pointer, but it provides no quotation for judges to assess.
The majority verdict is \emph{supports} on $2$ of those against $26$ of the other $87$.
This association does not isolate a causal effect of citation format.

All-three endorsement of both step and full evidence support is $9$ of $120$;
the consensus-admitted and arbitrated-or-single strata do not separate at this
size ($2$ against $7$ of $60$). Five of the $12$ records whose step the majority
rejects are GAIA retrieval-backend failures (Appendix~\ref{app:infra-scan}).
These related-model judgments do not replace independent human validation.

\begin{table}[h]
\centering\small
\caption{Blinded model-panel adjudication of the $120$-record semantic sample. Majority of three
seats per question; Split means no majority; Unanimous counts records on which all three seats
gave the same verdict.}
\label{tab:semantic-panel}
\resizebox{\linewidth}{!}{
\begin{tabular}{@{}lrrrrrr@{}}
\toprule
Question (majority of three seats) & \multicolumn{3}{c}{Majority verdict} & Split & Unanimous & Fleiss $\kappa$ \\
\midrule
Attributed step is a defensible decisive step & yes $76$ & alternative $26$ & no $12$ & $6$ & $70$ & $0.44$ \\
Quoted evidence supports the claim & supports $28$ & partially $85$ & does not $4$ & $3$ & $37$ & $0.13$ \\
Proposed replacement is plausibly better & yes $82$ & unclear $16$ & no $17$ & $5$ & $81$ & $0.52$ \\
\midrule
\multicolumn{7}{@{}l}{All three seats: step defensible and evidence at least partial $68$ of $120$; step yes and evidence supports $9$ of $120$ (Wilson $95\%$ $4.0$ to $13.6\%$)} \\
\bottomrule
\end{tabular}
}
\end{table}

\subsection{Release audit and revision history}
\label{app:release-audit}
The frozen release predates the checks of
student-visible support and expanded failure screening; its construction inputs omitted policy
system prompts and tool lists (Appendix~\ref{app:policy-inputs}).
A case-by-case audit of the frozen release rejected $37/60$ sampled rows. Equal sampling across
environments makes this a defect-finding study, not an estimate of corpus-wide prevalence.
The current checks address such defects; they do not retroactively validate the frozen inputs
used in Section~\ref{sec:experiments}. The failure cases and filter ledger follow.
The historical human-review results and protocol appear in Appendix~\ref{app:human-design}.

The review exposed incorrect tool schemas, missing instructions, and
infrastructure or grading failures attributed to the policy. The
frozen release also retains $54$ rows on tasks
whose reference answer fails the grader (Appendix~\ref{app:infra-scan}). Its responsible-agent
labels carry a defect of the same kind: $125$ of the $943$ held-out rows are planner--executor
transcripts whose gold agent is a unit that does not appear in them; the exporter now takes the
agent from the decisive step's actor, and the revised rubric rejects a row whose named
unit did not act in the trace.

The revised rubric checks exported input and target together for attribution,
evidence, repair, leakage and input faithfulness. Two model-family judges review
semantic criteria after deterministic checks. In a manual disagreement sample,
the stricter judge was right on $8$ of $10$ rows; $4$ of those $8$ predated
explicit markers for parallel tool calls. Export corrections are recorded, and
missing required evidence blocks strict admission.

Judges receive the student view. Long inputs retain the beginning, attributed
step and end; judges must reject earliest-step claims when preceding steps are
omitted. Acceptance therefore need not cover every character of a long input.

Table~\ref{tab:quality-ledger} applies the fifth rubric version to revised exports of the frozen
release's three evaluated splits. The model evaluations of Section~\ref{sec:experiments} used the
earlier frozen inputs, so these counts describe neither a retroactively filtered evaluation nor
human acceptance, and they do not redefine the $98$-case held-out subset frozen under the earlier
(second) rubric version, before any score on it. Later collections may retain a diagnosis and
its adverse verdict as metadata; retaining that record does not make it an accepted training
target. Likewise, semantic acceptance and replay coverage are separate properties, not successive
levels on a universal correctness scale.

\begin{table}[t]
\centering\small
\setlength{\tabcolsep}{5pt}
\caption{\textbf{Quality-control ledger.} Rows (source tasks) of the frozen release's evaluated
splits that remain after each stage of the fifth rubric version. These are retrospective checks,
not the training and evaluation inputs used for the reported model results.
The separate historical human audit is reported in Table~\ref{tab:human-audit};
its sampled strata do not form another stage of this split-specific ledger.}
\label{tab:quality-ledger}
\begin{tabular}{@{}lrrr@{}}
\toprule
Stage & Train & Development & Holdout \\
\midrule
Input records & $3{,}157$ ($1{,}678$) & $219$ ($135$) & $943$ ($496$) \\
After deterministic checks & $283$ ($208$) & $20$ ($12$) & $87$ ($70$) \\
After semantic filtering & $53$ ($46$) & $9$ ($6$) & $18$ ($14$) \\
\bottomrule
\end{tabular}
\end{table}

\subsection{Quality rubric and certification}
\label{app:quality-rubric}
The automated rubric has twelve items. A row must pass all except item 10,
which records ambiguity without rejecting the row. Deterministic checks decide
items 4 and 11; Claude Opus 5 and Gemini 3.1 Pro (preview) judge the remaining
items after those checks. Both judges must accept each rejecting item.
The human form is shorter (Appendix~\ref{app:human-design}).

\begin{artifactbox}[aedteal]{Admission rubric: what must hold}
\renewcommand{\arraystretch}{1.12}
\begin{tabular}{@{}p{.06\linewidth}p{.24\linewidth}p{.63\linewidth}@{}}
\toprule
Item & Check & Accept when; reject otherwise \\
\midrule
1 & Agent failure & An available agent decision could have avoided failure; exclude infrastructure, harness and grader faults. Budget exhaustion under a reasonable budget can be an agent failure. \\
2 & Attributed step & The trace supports a mistake at the named step, with no earlier supported decisive mistake. A final stop or a later symptom is insufficient. \\
3 & Responsible unit & The named unit appears in the trace and made the decision. \\
4 & Quotations & Every quote resolves verbatim in the student input at or before the blamed step; a truncation marker is not evidence. \\
5 & Entailment & The visible evidence supports the explanation's factual claims. \\
6 & Information access & The diagnosis uses visible evidence; the repair uses information available before the action, not its result or a future answer. \\
7 & Executability & The tool exists, arguments satisfy its schema, and the action can be issued at that point without placeholders. \\
8 & Corrective change & The proposal changes the blamed action and addresses the stated mistake. \\
9 & Specificity & The explanation names a concrete action, contradicted fact and consequence. \\
10 & Ambiguity & Record defensible alternatives. This item does not reject the row. \\
11 & Leakage and secrets & No unseen gold answer, verifier answer or credential survives in the student input or target. \\
12 & Input fidelity & Preserve the received task, instructions, tools and decision order; mark omissions and parallel calls. \\
\bottomrule
\end{tabular}
\end{artifactbox}

\begin{artifactbox}{Mechanical checks and export handling}
\renewcommand{\arraystretch}{1.1}
\begin{tabular}{@{}p{.24\linewidth}p{.70\linewidth}@{}}
\toprule
Check family & Implementation boundary \\
\midrule
Structure and visibility & Require an admitted attempt, an asserted error, a visible step and matching answer fields. Missing required records fail closed. Preserve malformed turns and parallel-call notes. \\
Environment validity & Reject retrieval outages, task quarantine, gold-grader contradictions, harness-submitted actions and known prompt/tool-contract defects. Exclude static Mind2Web from this admission protocol. \\
Evidence and leakage & Require a quote from before the blamed step; reject future-only evidence, copied gold actions, verifier text and arguments based on hidden content. \\
Repair and attribution & Validate tool arguments and changed action; reject later actions copied back as repairs. Require a unit that acted and sufficient voter support. \\
Export & Restore policy instructions and tool lists, align answer fields, re-derive the unit and redact credentials on both sides. Log each transformation. \\
\bottomrule
\end{tabular}
\end{artifactbox}

Rubric v5 also rejects later parallel calls labeled without turn-order notes and
quotes that were unavailable when the action was chosen. Among $69$ disagreements
on re-admitted rows, a hand check of $10$ found the stricter judge correct on $8$;
$4$ of those involved unavailable parallel-call results. The other judge was correct
on $2$. This targeted check is not a population accuracy estimate.
The $98$-case evaluation subset remains frozen under v2; v5 does not redefine it.
Versioned questions, full fail definitions and ordered check signatures remain in
the software release. We have not measured certification rates for the
corrected-contract pool or the full collection.

\paragraph{Registered controls.}
\label{app:registered-audits}
Appendix~\ref{app:false-alarm} reports the completed error-reporting control;
Appendix~\ref{app:human-design} gives the human-audit design and measured outcomes.

\clearpage
\section{Case Studies}
\label{app:case-studies}
\subsection{A failed trajectory, diagnosed and repaired}
\label{app:recovery-case}
\noindent\textbf{Case A | AgentBench-OS: a command succeeds, but counts the wrong lines.}
Actor: GPT-4.1 nano (ReAct); construction-time debugger: GPT-5 mini.

\begin{recoveryphase}{aedblue}{Task and failed trajectory}
Count lines containing \texttt{Linux} across the home directory's \texttt{.txt}
files, processing each file once. The agent finds the files and identifies those
with a match, then counts \emph{all} their lines.
\smallskip
\begin{tabular}{@{}P{.09\linewidth}P{.67\linewidth}P{.16\linewidth}@{}}
\toprule
Step & Action (early steps summarized) & Result \\
\midrule
0 & Find the text files. & Four files. \\
1 & Identify files containing \texttt{Linux}. & Three files. \\
2 & Concatenate those files and count their lines. & \texttt{5} \\
3 & Submit the count. & Verifier: fail. \\
\bottomrule
\end{tabular}
\end{recoveryphase}
\begin{recoveryphase}{aedcoral}{Error detected | Step 2 loses the line-level filter}
The debugger attributes the failure to the counting command. Selecting files that
contain a match does not make every line in those files a match. The shell reports
no execution error; the mistake is in what the agent counts.
\smallskip\par
\textbf{Original command, verbatim:}\par
\texttt{cat /root/file1.txt /root/file2.txt /root/file3.txt | wc -l}
\end{recoveryphase}
\begin{recoveryphase}{aedpurple}{Repair executed | Filter the lines before counting}
Replay resumes before the blamed action with the same actor and harness.
The replacement preserves the file list and filters the contents:
\smallskip\par
\texttt{grep -h "Linux" /root/file1.txt /root/file2.txt /root/file3.txt |\newline
\hspace*{1em}wc -l}
\smallskip\par
The tool returns \texttt{4}; the actor submits \texttt{4}.
\end{recoveryphase}
\begin{recoveryphase}{aedteal}{Verified outcome | Corrected branch passes; original-action control fails}
\begin{tabular}{@{}P{.51\linewidth}P{.19\linewidth}P{.22\linewidth}@{}}
\toprule
Branch from the shared checkpoint & Final answer & Task verifier \\
\midrule
Re-execute original action & \texttt{5} & \textcolor{aedcoral}{\textbf{FAIL}} \\
Execute proposed correction & \texttt{4} & \textcolor{aedteal}{\textbf{PASS}} \\
\bottomrule
\end{tabular}
\end{recoveryphase}
{\footnotesize Editorial condensation of one recorded replay pair. Both branches
bind to the same source prefix; restoration uses prefix re-execution. Ground-truth
access was permitted during diagnosis. This observed recovery does not establish
reliability, a unique cause or learned self-repair. Full records and checksums
accompany the case in the supplementary source materials.\par}

\clearpage
\subsection{Trace, target and competing interpretation}
\label{app:complete-cases}
Case B is a short frozen-v13 training example. Cases C and D are historical
v12r audit examples chosen to expose a tool-success mismatch and an unsupported
diagnosis. This is a purposive illustration, not an accuracy sample.
Editorial summaries are labeled; complete records and exact targets remain in the
source materials. None establishes learned recovery.
\begin{artifactbox}[aedamber]{Case B | BFCL | A quoted trace can still support a bad target}
\casefield{Task} Find every \texttt{test\_document.txt} in the current directory.\par\smallskip
\begin{tabular}{@{}P{.08\linewidth}P{.35\linewidth}P{.50\linewidth}@{}}
\toprule
Step & Action & Observation \\
\midrule
0 & \texttt{find(path=".", name=...)} & \texttt{matches: ["./test\_document.txt"]} \\
1 & \texttt{finish\_turn()} & Next user request: copy the file into archives and rename it. \\
\bottomrule
\end{tabular}
\casefield{Recorded target} Blame step 1 for not presenting the match; propose
\texttt{finish\_turn()} again. Step 0 is listed as an alternative.
\casefield{Reading} The proposed action repeats the blamed action, while the environment
has advanced to a new request. Located quotes alone do not resolve this contradiction.
Semantic review is required; this is not a verified repair.
\casefield{Source} Frozen v13 training example B. Actions and observations are abbreviated;
the target has not been silently corrected.
\end{artifactbox}

\begin{artifactbox}[aedamber]{Case C | BFCL | Mini-SWE-agent | Tool success, incomplete content}
\casefield{Visible request} Send \texttt{The latest stock price of XTC is \char36 150.75.} to
\texttt{USR003}.\par\smallskip
\begin{tabular}{@{}P{.08\linewidth}P{.45\linewidth}P{.40\linewidth}@{}}
\toprule
Step & Action & Observation \\
\midrule
1 & End the previous turn & The next user supplies the message request. \\
3 & \texttt{send\_message(..., message="The")} & \texttt{sent\_status: true} \\
6 & \texttt{view\_messages\_sent()} & \texttt{USR003: ["The"]} \\
\bottomrule
\end{tabular}
\casefield{Recorded target} Attribute the failure to step 3: the sent message is incomplete.
\casefield{Reading} Transport success does not satisfy the content request.
The excerpt shows the mismatch, but the original wire must be checked before assigning
truncation to the policy rather than the harness.
\casefield{Source} Purposively selected historical audit case;
intervening steps omitted. No independent human adjudication.
\end{artifactbox}

\begin{artifactbox}[aedamber]{Case D | AgentBench-DB | AutoGen | Missing verification is not proof of failure}
\casefield{Task} Change the attendance for November 30 from \texttt{1,458} to \texttt{1,500}.\par\smallskip
\begin{tabular}{@{}P{.08\linewidth}P{.52\linewidth}P{.33\linewidth}@{}}
\toprule
Step & Action (abbreviated) & Observation \\
\midrule
0 & \texttt{SELECT ... WHERE Date = 'November 30'} & Row includes \texttt{1,458}. \\
1 & \texttt{UPDATE ... SET Attendance = '1,500'} & \texttt{[]} \\
2 & Commit final answer \texttt{1,500} & Final answer committed. \\
\bottomrule
\end{tabular}
\casefield{Recorded target} Blame step 2 for committing without a verification query.
\casefield{Reading} The absence of a second \texttt{SELECT} does not show that the
\texttt{UPDATE} failed. Inspect the post-update state and grader before accepting this
diagnosis; the empty SQL result alone does not settle it.
\casefield{Source} Purposively selected historical audit case;
no independent adjudication. This example illustrates a label limitation.
\end{artifactbox}

\FloatBarrier

\section{Taxonomy definitions and validation study}
\label{app:error-taxonomy}
\label{app:taxonomy}
\paragraph{Taxonomy scope.}
Construction collects an attribution, explanation, evidence and proposed action;
it does not ask for an error-type label. We induce the taxonomy afterwards and
store assignments in a versioned sidecar. Relabeling adds a version without
overwriting previous assignments; each version retains its definitions and
executable labeling state. Error-type frequencies refer to the historical labeled
cohort specified below, not the full collection. This study tests labeling reliability. Its machine
assignments are neither training targets nor admission criteria.

\subsection{Independent labeling-method study}
\label{app:atlas-exploratory}
\label{app:failure-composition}
\paragraph{Historical harness-profile cohort.}
Figure~\ref{fig:taxonomy-harness} uses the September 6 labeling snapshot,
not the current $50{,}228$-pair collection. It contains $2{,}604$ distinct source
trajectories, of which the machine judge classified $2{,}092$ and abstained on
the remainder. We count each source trajectory once, regardless of how many
diagnoses it has. The figure shows harnesses with at least $20$ attempted
labels: $2{,}588$ trajectories across nine harnesses; $16$ trajectories fall
outside these groups. Each composition bar divides family counts by classified
trajectories, while the adjacent counts retain the attempted-label denominator.
Task and environment mixtures differ, and abstentions can be selective.
The profiles describe this selected cohort and do not estimate harness-caused
failure rates. The labeling studies below explain why these assignments remain
exploratory.
Appendix~\ref{app:restored-error-profiles} reports the model and environment
breakdowns from the same labels, with classification coverage beside each group.

\paragraph{Measured disagreement between two machine taxonomy judges.}
A separate run applies GPT-5 mini and DeepSeek-V4-Flash to $400$ sequentially selected
items from the annotation queue under the same seed taxonomy and a $12{,}000$-character tail window. At least one judge abstains on $43$ items, leaving $357$ jointly classified items. The two machine judges agree on $154/357$ modes ($43.1\%$): Cohen's $\kappa=0.313$ at mode level
and $\kappa=0.329$ after collapsing to families, agreement between two models rather than between
human annotators. Of $203$ mode disagreements,
$186$ cross family boundaries. The largest directed confusion is
premature termination to ignored planning constraint
($29$ cases), with $9$ in the reverse direction.

Thus family aggregation does little to resolve the disagreement. These figures describe
a limited, nonrandom cohort, conditional on both judges assigning a mode; they do not
measure agreement with human judgments. We retain the labels as exploratory machine
assignments, not validated training targets or estimates of corpus-wide semantic error
prevalence. Mechanical trace phenotypes and executed repair evidence remain separate
measurements. The confusion-matrix counts and both $\kappa$ values were recomputed from the aggregate
agreement record; the raw judge responses were not independently recounted.

\paragraph{A tested decision-order rubric.}
The seed taxonomy uses the same judges as the earlier agreement study. The candidate adds
family-first decision order and tie-break rules for commonly confused modes, keeping the
mode definitions and reply format fixed. Its renderer is v2, with a 12,000-character window
and 4,096 output-token cap. The trial selects 400 new items; 13 items lack a complete four-call
comparison after recorded reasoning-budget exhaustion, leaving 387 common items.
\begin{table}[ht]
\centering\small
\caption{Rubric trial on 387 common items. Agreement conditions on both judges assigning
a label within each arm, yielding different valid subsets; malformed responses are judge B
parse failures on the common cohort, not semantic abstentions.}
\label{tab:rubric-trial}
\begin{tabular}{@{}lrrrr@{}}
\toprule
Prompt & Both labelled & Mode $\kappa$ & Family $\kappa$ & Malformed (judge B) \\
\midrule
Seed & 275 & 0.258 & 0.186 & 88 \\
Rubric & 266 & 0.327 & 0.285 & 104 \\
\bottomrule
\end{tabular}

\end{table}
The mode-level change in $\kappa$ is $+0.069$ (paired item bootstrap 95\% interval
$[-0.022,+0.157]$); the family change is $+0.099$ ($[-0.018,+0.212]$).
Both intervals include zero. The experiment costs \$5.172 in recorded calls and does not
establish improved labeling reliability. We independently recompute both bootstrap intervals
from the stored verdict rows. Differential parse losses and arm-specific valid subsets limit
interpretation; additional paid labeling should follow parser validation.
The reproducibility materials retain the exact tested annotation instructions.
Its compact decision order is: invalid output $\rightarrow$ action; visible contradictory
information $\rightarrow$ observation (unobserved asserted state $\rightarrow$ memory);
unsatisfied termination $\rightarrow$ verification; explicit false progress assessment
$\rightarrow$ reflection; otherwise an infeasible next choice $\rightarrow$ planning.
This experimental annotation prompt is separate from the diagnosis and student prompts in
Appendix~\ref{app:prompts}.
\FloatBarrier

\section{Result tables for the body figures}
\label{app:result-tables}
Figure~\ref{fig:headline-results}a summarizes paired replay;
Table~\ref{tab:repair} gives its full results in the appendix.
The tables below provide the resource comparison and replay-stratum details.

\begin{table}[h]
\centering\scriptsize
\setlength{\tabcolsep}{3pt}
\caption{Resource against resource under one base, recipe and scorer: micro exact step (\%), and
unit$+$step only on AgenTracer's test split (our held-out cases share one gold agent). Unparseable
answers and prompts over the serving window ($1$ held-out and $16$ AgenTracer test rows) are misses;
trained rows are mean $\pm$ sd over three seeds. $\Delta$: seed mean of the paired contrast, with the
widest per-seed $95\%$ bootstrap interval, not an interval for the mean.}
\label{tab:cross-resource}
\resizebox{\linewidth}{!}{
\begin{tabular}{@{}lrrrrr@{}}
\toprule
 & & & \multicolumn{1}{c}{Our holdout ($n{=}943$)} & \multicolumn{2}{c}{AgenTracer test ($n{=}790$)} \\
\cmidrule(lr){4-4}\cmidrule(lr){5-6}
Training resource & Tasks & Seeds & Step & Step & Unit$+$step \\
\midrule
Qwen3-8B base & $0$ & --- & $48.04$ & $30.76$ & $21.14$ \\
\dataset\ (ours) & $1{,}656$ & $3$ & $48.14 \pm 0.42$ & $31.65 \pm 0.33$ & $25.78 \pm 0.19$ \\
AgenTracer released data v1.0.0 & $1{,}656$ & $3$ & $45.71 \pm 0.11$ & $40.72 \pm 0.48$ & $36.16 \pm 0.48$ \\
\midrule
$\Delta$ (\dataset\ $-$ AgenTracer data), paired & & & $+2.44$ [0.21, 4.81] & $-9.07$ [-12.41, -5.70] & $-10.38$ [-14.05, -6.96] \\
\bottomrule
\end{tabular}
}
\end{table}

\begin{table}[h]
\centering\small
\caption{The frozen v14 diagnosis release by environment. Rows and source tasks are counted over
the three evaluated splits; harness families and policy models are distinct values of the stored
provenance fields; paired replay attempts are the replay package's count for the environment
(Table~\ref{tab:replay-by-env}) and ``---'' means that environment has none.}
\label{tab:release-scale}
\begin{tabular}{@{}lrrrrr@{}}
\toprule
Environment & Rows & Source tasks & Harness families & Policy models & Paired replay attempts \\
\midrule
ALFWorld & $958$ & $763$ & $9$ & $5$ & $205$ \\
AppWorld & $438$ & $54$ & $9$ & $7$ & $291$ \\
AgentBench-DB & $426$ & $342$ & $9$ & $5$ & $530$ \\
BFCL & $414$ & $189$ & $9$ & $5$ & $579$ \\
Mind2Web (static) & $395$ & $243$ & $9$ & $10$ & --- \\
$\tau$-bench retail & $348$ & $113$ & $9$ & $6$ & $497$ \\
GAIA & $253$ & $68$ & $9$ & $10$ & --- \\
ScienceWorld & $207$ & $162$ & $9$ & $9$ & $57$ \\
WideSearch & $199$ & $97$ & $9$ & $5$ & --- \\
MCPMark & $186$ & $22$ & $9$ & $8$ & --- \\
AgentBench-OS & $152$ & $112$ & $9$ & $5$ & $458$ \\
SWE-bench Verified & $143$ & $117$ & $2$ & $7$ & $64$ \\
$\tau$-bench airline & $113$ & $6$ & $9$ & $10$ & $358$ \\
Terminal-Bench & $82$ & $17$ & $2$ & $4$ & $22$ \\
SWE-smith & $5$ & $4$ & $2$ & $1$ & $1$ \\
\midrule
\textbf{All 15 environments} & $4{,}319$ & $2{,}309$ & $9$ & $10$ & $3{,}062$ \\
\bottomrule
\end{tabular}

\end{table}

\subsection{Paired replay by environment}
\label{app:replay-by-env}
The following breakdowns use search-selected proposals, not the first proposals
in Table~\ref{tab:repair}. Under
the harness that produced the failure, with its state fully restored ($871$ pairs, $456$ tasks),
the difference is $47.1$ points [$42.3$, $51.7$]; a substituted ReAct (single) harness continued $1846$ pairs
($+43.5$ points) and $345$ had only partly restored multi-agent state ($+50.7$ points;
Table~\ref{tab:replay-by-continuation} below).
Table~\ref{tab:replay-by-env} breaks the selected-proposal result down by
environment. Every environment with at least one discordant pair favours the proposed replacement;
the one environment omitted from the rows contributed a single attempt with no discordant pair.
The attempt population matches Table~\ref{tab:repair}, but proposal selection and
the paired control outcomes differ; these are not strata of its first-proposal effect.
We retain one occurrence per diagnosis ID, choosing the first path in
lexicographic order. There are $684$ IDs in more than one collection cell because
later waves repeated configurations; $143$ disagree on the replay outcome.
Choosing the last occurrence instead changes the paired difference to $46.0$
points. This sensitivity describes the two tested duplicate-resolution rules.

\begin{table}[h]
\centering\small
\caption{Search-selected continuation success by environment. Orig.\ and Prop.\ are the pass rates of the
original and proposed action; Disc.\ $P{:}O$ counts pairs where only the proposal or only the
original passed; $p$ is the attempt-level exact McNemar test. These diagnostic $p$ values
assume independent attempts and do not account for repeated source tasks.
For selected-proposal provenance comparisons, use the task-clustered intervals
in Table~\ref{tab:replay-by-continuation}.}
\label{tab:replay-by-env}
\begin{tabular}{@{}lrrrrrl@{}}
\toprule
Environment & Pairs & Orig.\ pass & Prop.\ pass & $\Delta$ & Disc.\ $P{:}O$ & McNemar $p$ \\
\midrule
BFCL & $579$ & $24.0$ & $64.2$ & $+40.2$ & $248{:}15$ & $10^{-55}$ \\
AgentBench-DB & $530$ & $10.2$ & $55.1$ & $+44.9$ & $242{:}4$ & $10^{-66}$ \\
$\tau$-bench retail & $497$ & $22.7$ & $75.3$ & $+52.5$ & $274{:}13$ & $10^{-65}$ \\
AgentBench-OS & $458$ & $10.7$ & $89.5$ & $+78.8$ & $365{:}4$ & $10^{-102}$ \\
$\tau$-bench airline & $358$ & $3.1$ & $45.5$ & $+42.5$ & $159{:}7$ & $10^{-38}$ \\
AppWorld & $291$ & $5.2$ & $13.7$ & $+8.6$ & $38{:}13$ & $10^{-4}$ \\
ALFWorld & $205$ & $11.7$ & $41.5$ & $+29.8$ & $67{:}6$ & $10^{-14}$ \\
SWE-bench Verified & $64$ & $12.5$ & $35.9$ & $+23.4$ & $17{:}2$ & $10^{-4}$ \\
ScienceWorld & $57$ & $7.0$ & $75.4$ & $+68.4$ & $41{:}2$ & $10^{-10}$ \\
Terminal-Bench & $22$ & $18.2$ & $31.8$ & $+13.6$ & $3{:}0$ & $0.250$ \\
\midrule
\textbf{All environments} & $3062$ & $13.7$ & $59.1$ & $+45.3$ & $1454{:}66$ & $<10^{-300}$ \\
\bottomrule
\end{tabular}

\end{table}

Table~\ref{tab:replay-by-continuation} splits the same $3062$ pairs by how the continuation was
run, using the provenance the replay runner recorded on every attempt. Both arms of a pair use the same continuation
policy, harness, budget and temperature schedule; selection and state-fidelity limits remain, and
the strata differ in continuation harness and completeness of state restoration. $871$ pairs
ran under the harness that produced the failure from a fully restored state and speak about that
system directly. $1846$ pairs ran under a substituted ReAct~\citep{react} (single) harness because the source
harness (smolagents, Pydantic AI, AutoGen, LangGraph, or Multi-agent (planner--executor)) is not
supported by the replay runner; they identify the effect of the replacement under the substituted
continuation, not under the harness that failed. $345$ pairs are marked as having incompletely restored multi-agent state, so their pre-action state was reconstructed in
part. All three strata show positive selected-proposal contrasts. Their estimates
are distinct from the first-proposal contrast highlighted in the main text.

\begin{table}[h]
\centering\small
\caption{Search-selected continuation success by continuation provenance, same columns as
Table~\ref{tab:replay-by-env}; the interval is the task-clustered percentile bootstrap.}
\label{tab:replay-by-continuation}
\resizebox{\linewidth}{!}{\begin{tabular}{@{}lrrrrrl@{}}
\toprule
Continuation & Pairs & Orig.\ pass & Prop.\ pass & $\Delta$ [95\% CI] & Disc.\ $P{:}O$ & McNemar $p$ \\
\midrule
Substituted ReAct (single) continuation & $1846$ & $19.2$ & $62.7$ & $+43.5$ [39.5, 48.2] & $859{:}56$ & $10^{-185}$ \\
Same harness, state restored & $871$ & $6.1$ & $53.2$ & $+47.1$ [42.3, 51.7] & $419{:}9$ & $10^{-111}$ \\
Multi-agent state partly restored & $345$ & $4.1$ & $54.8$ & $+50.7$ [44.2, 57.8] & $176{:}1$ & $10^{-51}$ \\
\bottomrule
\end{tabular}
}
\end{table}

Two counts bound how the paired continuation of Section~\ref{sec:repair} could be read as an
artefact of the budget rather than of the action: only $11$ of the $3062$ attempts merely exhausted
their step budget, and none of those sits in the cell where only the original action passed; the
eleventh environment in the corpus contributed a single attempt and no discordant pair, which is
why Table~\ref{tab:replay-by-env} reports ten.

The paired continuation in Section~\ref{sec:repair} tests a proposed replacement against its original
action; it does not measure the benefit of training a debugger.

\subsection{The located-failure cohort and its failure accounting}
\label{app:located-cohort}
The located-recovery cohort of Table~\ref{tab:located} was frozen before execution: $150$ attempts drawn from
$1235$ eligible, one per source task, across $11$ environments, with the selection rule and the
analysis rules stored in the cohort file and carried into every results file written against it.
Arm order is randomised within an attempt. The rules require that every planned trajectory stay in
the denominator, that infrastructure outcomes get their own column rather than being folded into
either arm, and that an unknown-as-failure sensitivity be reported beside the complete-case result,
because a crash can be caused by the intervention it is scored under.

Counting every attempt and scoring a crash as a failure moves the four arms to $18.6$, $26.0$,
$31.3$ and $35.3$ per cent, against the complete-case $18.7$, $27.1$, $32.2$ and $36.1$; the
ordering and the reading of the section do not change. The exact McNemar test on attempts
binarised as "recovered at least once" agrees with the intervals where they are far from zero
($p=0.024$ for the diagnosis against re-application, $p=0.001$ for the deployed text) and not
where they are close ($p=0.42$ for the content-free nudge, $p=0.12$ for the content effect); it
discards the per-attempt rates and the task clustering the intervals use, and is reported as a
supplementary check rather than as the inference.

The four arms differ only in the coach text at the pre-action state. The content-free arm sends
"Reconsider your next action."; the clean arm appends the recorded explanation and directive; the
deployed arm replaces the opening with "Your previous attempt failed." and calls the explanation a
diagnosis of a mistake the actor is about to make. Every branch stamps the framing it received, and
the run is rejected if any branch stamps a framing other than its arm's; the reported run has no
such mismatch. The earlier continuation cap exhausted the reasoning budget in
$52$ of $450$ coached continuations, compared with $13$ after correction.
That superseded run estimated the content effect at $+0.8$ points
($-6.3$ to $7.8$); we use the corrected run's $+5.6$ ($0.0$ to $11.3$).

\subsection{Resource against resource under one recipe}
\label{app:cross-resource}
\begin{figure}[h]
\centering
\includegraphics[width=\linewidth]{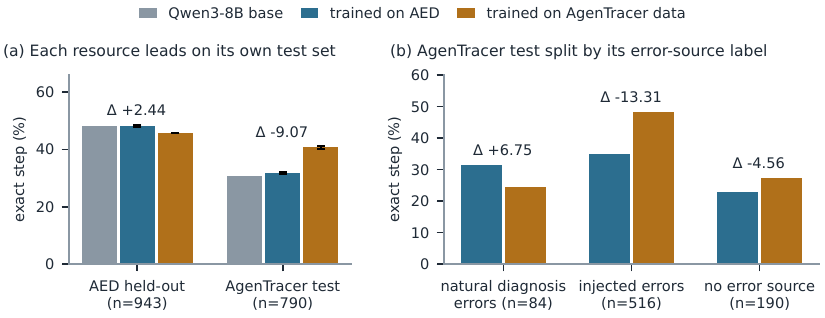}
\caption{\textbf{Resource against resource under one base, recipe and scorer.} Exact step on each
resource's test set for the untrained base and for students trained on \dataset{} or AgenTracer
data, followed by AgenTracer's test split under its own error-source labels.}
\label{fig:cross-resource}
\end{figure}
Table~\ref{tab:cross-resource} holds the cells of the comparison described in
Section~\ref{sec:cross-resource}: Qwen3-8B trained under one recipe and one two-key contract on
\dataset\ and on AgenTracer's released v1.0.0 training data, scored on our frozen held-out set and on
their released test split. The AgenTracer-data arm is not AgenTracer-8B, whose weights are not
released; the comparison is between two resources, not two systems.

Both resources improve joint attribution on AgenTracer's test split, where the
responsible agent varies. Neither establishes an exact-step gain on the other's
test set. On our holdout, the joint score also rewards matching a constant unit
label; \dataset{} does not improve exact step over base, with $3$ of $3$ seed
intervals containing zero.

\begin{center}\small
\begin{tabular}{@{}P{.34\linewidth}P{.25\linewidth}r@{}}
\toprule
Evaluation and metric & Resource versus base & Difference, pp \\
\midrule
AgenTracer, unit$+$step & \dataset\ & $4.64 \, [1.52, 7.59]$ \\
 & AgenTracer data & $15.02 \, [11.01, 19.49]$ \\
Our holdout, unit$+$step & \dataset\ & $5.16 \, [1.79, 8.22]$ \\
 & AgenTracer data & $-3.08$ \\
Our holdout, exact step & \dataset\ & $+0.11 \, [-2.72, 2.99]$ \\
\bottomrule
\end{tabular}
\end{center}

\begin{center}\small
\begin{tabular}{@{}P{.34\linewidth}P{.22\linewidth}P{.36\linewidth}@{}}
\toprule
AgenTracer error source & Leading resource & Exact-step margin, pp \\
\midrule
Injected errors ($516$) & AgenTracer data & $13.31 \, [8.53, 18.02]$ \\
Natural diagnosis errors ($84$) & \dataset{} & $+6.75$; interval reaches zero \\
Unspecified source ($190$) & AgenTracer data & $4.56$ (coding and mathematics) \\
\bottomrule
\end{tabular}
\end{center}
The largest margin occurs on injected errors, which are absent from \dataset{}.
The natural-error stratum establishes no resource advantage; these subsets do
not isolate error source from task and annotation differences. Contrasts are
seed means; brackets give the widest per-seed $95\%$ percentile-bootstrap
interval (over source tasks or question identifiers), not an interval for the
mean. All intervals are nominal.

\section{Training results and comparison protocols}
\label{app:training-experiments}
This section reports completed training comparisons and their evaluation protocols,
including gains and regressions.

\subsection{External-label transfer, in both directions}
\label{app:external-transfer}
Table~\ref{tab:training-results} reports Who\&When~\citep{whoandwhen} and
TrajErrBench~\citep{trajdebug}, whose labels originate outside this project.
Their transfer results differ: over TrajErrBench's
$486$ cases the untrained base scores
$18.72\%$ while the three seeds score $15.43$, $15.43$ and $20.99\%$, so two of the three
regress. We report both rather than the favourable one alone. The two benchmarks differ in
what a step is -- Who\&When labels the decisive step of a multi-agent transcript, TrajErrBench
the first faulty step of a single-agent trajectory -- and we did not measure which of those
differences drives the sign. Different annotators do not establish task-disjoint
transfer; the task-overlap sensitivity below further limits that interpretation.
Both cohorts score every case, with an unparseable answer counted as a miss.
For both benchmarks, the producer maps the three trained evaluation runs to the
checkpoints reported in Table~\ref{tab:training-results} and records a common
training-data digest. This mapping has not been independently verified against
the weights loaded for evaluation.

\paragraph{External task overlap.}
An audit after the original evaluations found shared task identities or exact
goals between training and Who\&When. The audit did not flag trajectory overlap
at its stated similarity threshold. Absence of such a hit does not rule out
other overlap. We rescore the existing predictions, keeping
each student paired with the base on the same cases; no model or prompt is retuned.

\begin{table}[ht]
\centering\small
\caption{Post-hoc Who\&When task-overlap sensitivity. Mean exact-step change
against base across the larger arm's three seeds, under the unified prompt.
Intervals resample shared evaluation items, conditional on these seeds;
they do not measure uncertainty over new training seeds.}
\label{tab:external-task-overlap}
\begin{tabular}{@{}lrrr@{}}
\toprule
Evaluation subset & Cases & Gain (pp) & $95\%$ interval \\
\midrule
All cases & $184$ & $+7.97$ & $[1.63, 14.49]$ \\
No flagged task overlap & $149$ & $+4.92$ & $[-2.24, 12.08]$ \\
Flagged task overlap & $35$ & $+20.95$ & $[5.71, 36.19]$ \\
\bottomrule
\end{tabular}
\end{table}

The larger improvement occurs on shared tasks. All three point estimates remain
positive on the unflagged cases, but their seed-mean interval includes zero;
the original headline does not establish transfer beyond task overlap.
Task identity and normalized goals define the flagged Who\&When cases. The wider
audit also checks goal-text containment and trajectory shingles, with matching
rules and per-case decisions retained in the companion artifacts. Removing flagged
TrajErrBench cases does not reverse its negative seed-mean contrast. These are
retrospective sensitivity analyses, not results from decontaminated retraining,
which the two paragraphs below report.
\paragraph{Replacement-data retraining.}
After replacing flagged training rows, we repeat the larger-arm recipe with all planned seeds.
Under the pre-specified item-bootstrap rule and original base decode, the Who\&When
mean exact-step contrast is $+7.79$ pp ($95\%$ interval $[1.45, 14.31]$).
We retain this protocol-specific result alongside the original overlap sensitivity.
The new comparison does not rescue the losses under other evaluation interfaces
(Table~\ref{tab:public-retrain-complete}).
\begin{table}[ht]
\centering\small
\setlength{\tabcolsep}{3.5pt}
\caption{All fixed public conditions after replacement-data retraining. Exact-step
percentages for each seed and their mean change against the retained raw-verified base.
AEB uses an equal-weight mean over its three environments; other rows use case accuracy.
Intervals resample source-task groups, conditional on these seeds and fixed environments.
They are sensitivity intervals, not multiplicity-adjusted tests or replacements for the
pre-specified item-bootstrap result above. Interfaces can differ in inputs, answer
contracts and step coordinates; rows do not isolate a prompt-wording effect.
HC/AG: hand-crafted/algorithm-generated; gold: supplied task reference answer.}
\label{tab:public-retrain-complete}
\begin{tabular}{@{}llrrrl@{}}
\toprule
Interface & Cohort & Seed 17 & Seed 202 & Seed 828 & Mean $\Delta$ [95\% interval] \\
\midrule
Shared & HC & $5.17$ & $8.62$ & $3.45$ & $-9.77$ [$ -17.24, -2.87 $] \\
 & HC + gold & $10.34$ & $6.90$ & $6.90$ & $-12.64$ [$ -24.14, -1.72 $] \\
 & AG & $9.52$ & $11.90$ & $3.17$ & $-31.48$ [$ -40.21, -22.75 $] \\
 & AG + gold & $6.35$ & $8.73$ & $5.56$ & $-32.80$ [$ -41.53, -24.34 $] \\
 & AEB & $15.33$ & $22.00$ & $15.00$ & $+1.78$ [$ -3.67, 7.33 $] \\
 & TEB & $4.12$ & $11.73$ & $6.58$ & $-20.51$ [$ -24.55, -16.39 $] \\
\addlinespace[2pt]
Unified & AEB & $21.67$ & $23.67$ & $22.67$ & $+3.67$ [$ -0.34, 7.89 $] \\
 & TEB & $17.49$ & $17.90$ & $17.08$ & $-0.21$ [$ -3.29, 3.02 $] \\
\addlinespace[2pt]
Unified W\&W & W\&W & $33.15$ & $34.24$ & $32.61$ & $+6.70$ [$ 0.71, 12.89 $] \\
\addlinespace[2pt]
Benchmark-native & HC & $5.17$ & $5.17$ & $3.45$ & $+2.87$ [$ -3.45, 9.21 $] \\
 & HC + gold & $5.17$ & $5.17$ & $5.17$ & $+1.72$ [$ -3.45, 6.90 $] \\
 & AG & $11.11$ & $7.94$ & $11.11$ & $-11.38$ [$ -19.31, -3.44 $] \\
 & AG + gold & $7.94$ & $8.73$ & $9.52$ & $-5.56$ [$ -11.90, 0.79 $] \\
\bottomrule
\end{tabular}
\end{table}
We keep the original and later base decodes: one shared base response set does not
measure decode variability. With the later base and source-task resampling, the
Who\&When mean contrast remains positive. The unified AEB base is the original decode.
Raw-answer, input-hash and declared-checkpoint checks pass for the full panel;
served paths do not attest loaded weight bytes. Replacing rows changes training
examples and optimization paths, so these comparisons do not isolate a causal contribution
of overlap or bound its share of the original gain. We retain both original and retrained results.

\paragraph{Sensitivity to replacing flagged training examples.}
We also compare each replacement-data model with the original model from the
same training seed. All three planned runs completed. The training total remains
$1{,}656$ rows: $1{,}574$ retained and $82$ replaced, including $30$ training rows
associated with the Who\&When overlap audit. Evaluation uses the full $184$-case
Who\&When cohort, not just the $35$ flagged evaluation cases.
The replacement-minus-original changes are $-1.63$ pp for seed $17$
($95\%$ interval $[-3.53,1.59]$), $+2.17$ for seed $202$ ($[-2.26,5.67]$),
and $-1.09$ for seed $828$ ($[-5.26,3.50]$). No per-seed interval excludes zero;
the nominal paired $p$ values are $0.45$, $0.42$ and $0.80$.
All six models satisfy the answer contract on all cases.
Each interval conditions on one pair of trained checkpoints. All seeds use the
same original and replacement datasets, so they assess this fixed replacement
scheme across training runs. The comparison measures sensitivity to replacing
flagged examples; it does not identify their causal contribution to the original gain.

\paragraph{Full-policy input sensitivity.}\label{app:tau2-policy-sensitivity}
We restored the complete source policy text in the same airline and retail cases,
keeping other messages, labels, checkpoints and decoding settings fixed.
This post-hoc input check does not replace the primary evaluation.

\begin{table}[ht]
\centering\small
\caption{Exact-step hits on all $400$ $\tau^2$ cases. Both policy conditions use
the recorded budget-forcing protocol. First-call hits and continuation counts
describe the full-policy run; invalid answers remain misses.}
\label{tab:tau2-policy-sensitivity}
\begin{tabular}{@{}lrrrr@{}}
\toprule
Model & Clipped policy & Full policy & First call & Continued \\
\midrule
Base & $87$ & $78$ & $59$ & $65$ \\
SFT, seed 17 & $63$ & $71$ & $71$ & $0$ \\
SFT, seed 202 & $64$ & $62$ & $62$ & $0$ \\
SFT, seed 828 & $92$ & $83$ & $83$ & $0$ \\
\bottomrule
\end{tabular}
\end{table}

Restoring policy text does not yield a consistent student improvement:
one seed improves and two decline. The base also loses accuracy, so a smaller
student--base gap under this input does not show better student localization.
New-run raw responses reproduce both scoring passes, with runtime paths and
checkpoint-file hashes recorded. Historical base raw generations remain unavailable.
The comparison does not exclude other context effects or identify the cause of transfer loss.

\subsection{Completed full-fine-tuning recipes}\label{app:fullft-recipes}
We fine-tune all parameters of Qwen3-8B in BF16 with AdamW~\citep{adamw}, a learning rate
of $10^{-5}$, cosine decay, $3\%$ warmup and gradient clipping at $1.0$.
Table~\ref{tab:fullft-recipes} reports the completed run manifests; no rows were dropped.
For diagnosis, we average target-token loss within each task and then across tasks.
For actors, we average over target tokens in each optimizer window; an example is one
supervised assistant turn. We mask the input context in both recipes.

\begin{table}[ht]
\centering\small
\setlength{\tabcolsep}{3.5pt}
\caption{Recorded training sizes and optimization budgets. Examples are per epoch;
updates cover the full run. Batch denotes examples per optimizer update.}
\label{tab:fullft-recipes}
\begin{tabular}{@{}lrrrrrr@{}}
\toprule
Recipe & Tasks & Rows & Examples & Batch & Epochs & Updates \\
\midrule
\multicolumn{7}{@{}l}{\textit{Diagnosis: seeds 17, 202, 828; sequence limit 24,576}} \\
Full diagnosis & $948$ & $948$ & $948$ & $12$ & $2$ & $158$ \\
Full diagnosis & $1{,}656$ & $1{,}656$ & $1{,}656$ & $12$ & $2$ & $276$ \\
\midrule
\multicolumn{7}{@{}l}{\textit{Actor: seed 17; sequence limit 12,288}} \\
Success-only & $1{,}093$ & $1{,}093$ & $10{,}595$ & $16$ & $2$ & $1{,}326$ \\
Preventive & $1{,}183$ & $1{,}790$ & $11{,}912$ & $16$ & $2$ & $1{,}490$ \\
Post-error action & $1{,}183$ & $1{,}790$ & $11{,}912$ & $16$ & $2$ & $1{,}490$ \\
Post-error reflection & $1{,}183$ & $1{,}790$ & $11{,}912$ & $16$ & $2$ & $1{,}490$ \\
Success-only, longer & $1{,}093$ & $1{,}093$ & $10{,}595$ & $16$ & $3$ & $1{,}989$ \\
\bottomrule
\end{tabular}
\end{table}

Each GPU processes one example per microbatch. Diagnosis uses six GPUs with two
accumulation steps, except the 948-task seeds 202 and 828, which use four GPUs
with three steps. The actor runs use eight GPUs with two steps and zero-loss
padding for the last optimizer window of each epoch.
We select diagnosis checkpoints on development family-macro exact-step accuracy
with thinking disabled, requiring at least $80\%$ parseability. Within $1$ point
of the best, we prefer fewer epochs, then the lower learning rate. All six
948- and 1,656-task runs select epoch 2; Appendix~\ref{app:fullft-scaling}
reports the smaller scaling arms. Actors use the final checkpoint. The actor mixtures differ in
task coverage and optimization budget; the longer success-only run is a duration
sensitivity, not a matched-budget control.

\subsection{Diagnosis targets and evaluation}
\label{app:g2-designs}
Table~\ref{tab:training-results} compares full-diagnosis fine-tuning with prompted
models on the frozen holdout. Both the unfiltered and certified-subset
full-diagnosis rows have completed evaluation.

Full fine-tuning uses explanation, evidence and proposed action as the target,
with attribution fields first. Frozen-protocol and budget-forced evaluations
remain separate. Agreement is with recorded teacher labels; the historical human
audit does not relabel this test set.
Claims of superiority require a paired interval excluding zero; matching requires
the predeclared two-point non-inferiority margin.

\paragraph{Prompted baselines.} The prompted rows saw the identical $943$
prompts and decoding (temperature $0$ where the provider accepts it, $8{,}192$-token budget,
thinking as each channel allows), were never trained on \dataset, and each answered all $943$ rows;
an unparseable answer is a miss, parse rates run from $96.50$ to $100.00$, and their family-macro is
recomputed from per-item predictions over the holdout's $56$ families, as for the trained rows.

\subsection{Agent post-training design}
\label{app:g3-design}
Table~\ref{tab:g3} compares held-out task success from the initial state, without
a debugger-supplied location. Training uses the corresponding benchmarks'
train-side tasks. The historical pilot's success-only, preventive and reflective
mixtures differ in task coverage, updates and token budget. The current
post-error action-only control isolates the addition of reflection more closely:
its erroneous action, feedback and corrected actions match the reflective arm.
Training completion alone does not establish an evaluated policy benefit.

\subsection{Comparison scope and interpretation}
\label{app:experiment-claims}
The reported comparisons use separate cohorts for diagnosis production,
correction replay, diagnostic learning and actor training. Their input
information and evaluation units determine the conclusions each supports.

\paragraph{Counting and uncertainty.}
All planned evaluation cases remain in the denominator, including missing and
unparseable answers, except for the explicitly identified complete-case recovery
and real-environment summaries. We report their failure accounting alongside
the scores. Source-task clustering addresses repeated tasks; training-seed
variation measures a different source of uncertainty. First proposals and
search-selected corrections retain separate paired controls.

\paragraph{Overlap sensitivity of the internal comparison.}
\label{app:overlap-sensitivity}
We found no exact overlap in task identities, task content, complete traces or the
audited trajectory prefixes. The internal holdout shares environments with training,
and most examples also share the harness and policy; it measures
within-distribution generalization. We also checked similar input text and repeated
user goals across task variants. In both checks, the gain from the larger training
set occurs outside the flagged subsets, whose estimated gains are zero or negative
and whose intervals include zero. These checks do not rule out other forms of
train--test similarity. The audit records retain the matching rules, subgroup scores
and uncertainty.

\paragraph{Actor comparisons.}
The recipes differ in task coverage, per-environment exposure and optimizer
updates. Their contrasts describe the combined recipes and do not isolate the
effect of repair supervision. Post-error action-only and reflective arms share
erroneous actions, observed feedback and corrected continuations; reflection also
changes the context for later actions and the supervised-token budget. A
state-changing error requires a continuation executed from that post-error state;
a passing pre-error branch cannot be spliced after it.

\paragraph{Construction and human-review scope.}
Diagnosis production measures yield under the stated output requirements and
teacher access. Agreement with a construction teacher and admission yield do not
establish human-assessed label correctness. The prepared human audit assesses
individual records under shared AI assistance, not blinded paired outputs from
different construction methods. Its coverage-selected sample does not estimate
collection-wide accuracy. The historical student comparisons also vary teacher
access or target construction, so their contrasts do not isolate pipeline quality.

\paragraph{Public comparisons.}
Within each reported protocol, comparisons share frozen cases, input information
and scoring rules. First-call and budget-forced evaluations remain separate.
Who\&When's responsible agent, AgentErrorBench's module and TrajErrBench's
error mode are distinct labels. Published native-protocol scores provide context
outside the shared-prompt tables. The AgenTracer-data comparison uses a common
student recipe and compares training resources, not the original AgenTracer
algorithm. These comparisons support protocol-specific conclusions rather than
a public leaderboard ranking.

\subsection{Diagnosis production on a common failure pool}
\label{app:construction-production}
The common failure pool covers $1{,}068$ source tasks, $16$ environments, $10$ harness families and $15$ policy models.
It excludes evaluation-exposed tasks and caps contributions per task. Every
configuration receives the same failed executions. Table~\ref{tab:construction-production}
includes abstentions, malformed diagnoses and engine errors in the input
denominator. The main table shows representative single-pass, deep-analysis,
multi-model and citation-first configurations; the full comparison is below.

\begin{table}[H]
\centering\small\resultsetup
\caption{Historical production configurations on the common failure pool.
Pairs satisfy the recorded structured-diagnosis and trace-citation rule.
This is an output-compatibility measurement, not human-validated accuracy.}
\label{tab:construction-production}
\begin{tabular*}{\linewidth}{@{\extracolsep{\fill}}P{.5\linewidth}*{3}{>{\raggedleft\arraybackslash}p{.14\linewidth}}@{}}
\toprule
Diagnosis configuration & Pairs & Yield & USD / input \\
\midrule
AgentDebugX: all-at-once & $735$ & $49.0\%$ & $0.0052$ \\
AgentDebugX: binary search & $3$ & $0.2\%$ & $0.0100$ \\
AgentDebugX: counterfactual & $0$ & $0.0\%$ & $0.0136$ \\
AgentDebugX: heuristic & $802$ & $53.5\%$ & $0.0000$ \\
AgentDebugX: ensemble & $437$ & $29.1\%$ & $0.0152$ \\
AgentDebugX: success reference & $756$ & $50.4\%$ & $0.0054$ \\
AED citation-first judge & $1{,}403$ & $93.5\%$ & $0.0067$ \\
Iterative investigation & $192$ & $12.8\%$ & $0.0088$ \\
Multi-model consensus & $726$ & $48.4\%$ & $0.0247$ \\
Rule-based triage & $0$ & $0.0\%$ & $0.0000$ \\
Success-trace comparison & $0$ & $0.0\%$ & $0.0000$ \\
AgentDebugX: deep analysis & $987$ & $65.8\%$ & $0.0143$ \\
Prospective diagnosis & $713$ & $47.5\%$ & $0.0110$ \\
\bottomrule\end{tabular*}

\end{table}

\paragraph{What differs.}
Single-model configurations use GPT-5-mini. Consensus uses GPT-5-mini,
Claude Haiku 4.5 and Gemini 3.6 Flash voters, with Gemini as arbiter.
Each engine keeps its production renderer and decoding defaults. Success-reference
methods additionally use available same-task successful traces. Task reference
answers and verifier signals are withheld; historical inputs also omit the
policy's system prompt. AgentDebugX adapters invoke its diagnosis implementations
\citep{agentdebugx}; deep analysis omits the separate taxonomy-generalization call.
Thus neither the teacher ensemble nor rendering is controlled across all rows.

\paragraph{What the cost and yield establish.}
Costs use whole-job diagnosis meters divided by every input, including failed
calls. They exclude trajectory generation, correction replay and hosting.
Pair yield requires a structured location and a matched citation under the
historical export rule; a matching quote does not establish a correct attribution
or a supported correction. Methods that do not emit that citation format can
have low yield without being poor localizers. The comparison supports choosing
an export-compatible production route, not a claim of superior diagnosis quality.

\subsection{Construction cost and scale estimates}
\label{app:construction-costs}

\paragraph{Accounting unit.}
For a cohort with a fixed acceptance rule, cost per retained record is total
construction spend divided by retained records, including spend on rejected
candidates. Rollout, diagnosis, review and optional replay have separate meters;
hosting and human labor require additional accounting. The studies below cover
different cohorts and must not be added into a full-collection bill.
USD values use recorded token usage and the configured channel prices, not
independently reconciled invoices.
The collection waves use GPT-5.6-Sol with the all-at-once diagnosis engine;
their policy and task mixtures differ. Semantic-review and selected-row cost
estimates for a separate pool appear in Table~\ref{tab:concise-admission-full}.

\begin{table}[H]
\centering\small
\caption{API cost by pipeline stage in two historical collection waves, re-read
on September 24. A cell is one environment--harness--policy run configuration.
Completed cells supply Table~\ref{tab:cost-env-harness}; stopped and unsettled
cells remain in this accounting ledger. Only rollout and diagnosis appear in these meters.}
\label{tab:cost-stage-ledger}
\begin{tabular}{@{}lrr@{}}
\toprule
Meter scope & Cells & Logged USD \\
\midrule
Completed cells: rollout & $1{,}198$ & $415.53$ \\
Completed cells: diagnosis & $1{,}198$ & $343.44$ \\
Completed cells: combined & $1{,}198$ & $758.96$ \\
\midrule
Stopped cells (settled meters) & $10$ & $0.12$ \\
Unsettled meters (logged to date) & $7$ & $8.84$ \\
\bottomrule
\end{tabular}

\end{table}

\begin{table}[H]
\centering
\caption{\textbf{Construction API cost by environment and harness.} Both panels
partition the same completed cells; they are not additive. Runs includes successful
and failed task attempts. Diag. (D) counts runs with a recorded valid-step diagnosis,
before independent semantic acceptance. USD includes rollout and diagnosis;
USD/D is their combined cost per diagnosis. Model and task mixtures differ across
rows, so this is descriptive accounting, not a controlled harness ranking.}
\label{tab:cost-env-harness}
\begin{minipage}[t]{.49\linewidth}\vspace{0pt}\centering\scriptsize
\setlength{\tabcolsep}{2.5pt}
\begin{tabular}{@{}lrrrr@{}}
\toprule
Environment & Runs & Diag. & USD & USD/D \\
\midrule
AgentBench DB & $2{,}140$ & $418$ & $20.62$ & $0.049$ \\
AgentBench OS & $2{,}140$ & $413$ & $18.18$ & $0.044$ \\
ALFWorld & $1{,}150$ & $754$ & $100.43$ & $0.133$ \\
BabyAI & $630$ & $201$ & $15.71$ & $0.078$ \\
BFCL & $1{,}600$ & $666$ & $51.63$ & $0.078$ \\
GridWorld & $1{,}000$ & $115$ & $10.51$ & $0.091$ \\
InterCode Bash & $1{,}060$ & $103$ & $6.98$ & $0.068$ \\
Jericho & $590$ & $388$ & $29.32$ & $0.076$ \\
MCPMark & $717$ & $398$ & $186.74$ & $0.469$ \\
PDDL & $910$ & $453$ & $37.68$ & $0.083$ \\
ScienceWorld & $927$ & $489$ & $68.54$ & $0.140$ \\
$\tau$ airline & $1{,}013$ & $721$ & $83.77$ & $0.116$ \\
$\tau$ retail & $1{,}630$ & $401$ & $39.82$ & $0.099$ \\
Text-to-SQL & $1{,}000$ & $265$ & $11.38$ & $0.043$ \\
TextCraft & $1{,}060$ & $274$ & $12.09$ & $0.044$ \\
TextQuest & $1{,}000$ & $160$ & $15.18$ & $0.095$ \\
TextWorld Cooking & $432$ & $225$ & $21.37$ & $0.095$ \\
Warehouse & $1{,}000$ & $41$ & $7.97$ & $0.194$ \\
WebShop-lite & $1{,}000$ & $142$ & $8.87$ & $0.062$ \\
Wordle & $630$ & $271$ & $12.17$ & $0.045$ \\
\midrule
Total & $21{,}629$ & $6{,}898$ & $758.96$ & $0.110$ \\
\bottomrule\end{tabular}\end{minipage}
\hfill
\begin{minipage}[t]{.49\linewidth}\vspace{0pt}\centering\scriptsize
\setlength{\tabcolsep}{2.5pt}
\begin{tabular}{@{}lrrrr@{}}
\toprule
Harness & Runs & Diag. & USD & USD/D \\
\midrule
AFlow & $90$ & $72$ & $4.79$ & $0.067$ \\
AgentScope & $240$ & $99$ & $4.12$ & $0.042$ \\
AutoGen & $2{,}320$ & $600$ & $49.69$ & $0.083$ \\
AutoGen group & $460$ & $274$ & $16.78$ & $0.061$ \\
CAMEL & $240$ & $91$ & $4.31$ & $0.047$ \\
CrewAI & $220$ & $99$ & $3.79$ & $0.038$ \\
DSPy & $240$ & $94$ & $7.22$ & $0.077$ \\
Google ADK & $240$ & $82$ & $3.87$ & $0.047$ \\
LangGraph & $2{,}321$ & $591$ & $48.82$ & $0.083$ \\
LlamaIndex & $240$ & $93$ & $6.61$ & $0.071$ \\
MetaGPT & $180$ & $142$ & $4.78$ & $0.034$ \\
Planner--executor & $834$ & $474$ & $45.72$ & $0.096$ \\
Native tools & $4{,}572$ & $1{,}363$ & $166.52$ & $0.122$ \\
OpenAI Agents & $240$ & $113$ & $3.84$ & $0.034$ \\
OpenHands & $2{,}318$ & $555$ & $79.78$ & $0.144$ \\
ReAct & $4{,}555$ & $1{,}451$ & $241.89$ & $0.167$ \\
smolagents & $2{,}319$ & $705$ & $66.43$ & $0.094$ \\
\midrule
Total & $21{,}629$ & $6{,}898$ & $758.96$ & $0.110$ \\
\bottomrule\end{tabular}\end{minipage}

\end{table}

\paragraph{Interpreting the variation.}
The cost of obtaining a failure depends on policy success, trace length and tool
interaction; the cost of retaining a pair also depends on the acceptance rule.
Completing diagnosis does not guarantee a trace-supported or semantically accepted
target. These tables therefore do not price the final quality-admitted release.
Stopped cells have settled meters but no completed collection protocol; unsettled
meters may omit outstanding calls. We show their logged spend separately instead
of treating either group as zero-cost output.

\begin{table}[H]
\centering\scriptsize\resultsetup
\caption{\textbf{Diagnosis-only production cost on the common failure pool.}
The same representative configurations as Table~\ref{tab:pipeline-main}; all
configurations and yields appear in Table~\ref{tab:construction-production}.
Pair means a structured location with a matching trace citation under the
historical export rule. The last column is a linear budget scenario for
50,000 such outputs, not measured spending or a quality-certified release estimate.}
\label{tab:cost-methods}
\begin{tabular*}{\linewidth}{@{\extracolsep{\fill}}lrrrr@{}}
\toprule
Configuration & Calls/input & USD/input & USD/pair & USD/50k pairs \\
\midrule
AgentDebugX: all-at-once & $1.01$ & $0.0052$ & $0.0106$ & $528$ \\
AgentDebugX: deep analysis & $4.79$ & $0.0143$ & $0.0218$ & $1{,}089$ \\
Multi-model consensus & $3.18$ & $0.0247$ & $0.0510$ & $2{,}549$ \\
AED citation-first judge & $1.00$ & $0.0067$ & $0.0072$ & $359$ \\
\bottomrule
\end{tabular*}

\end{table}

For method $m$, with total diagnosis spend $C_m$ over all input failures and
$N_m$ compatible outputs, we report $c_m=C_m/N_m$ and project $M c_m$ for $M$
outputs. This assumes the same task mix, teacher, prices, trace lengths and
acceptance yield. Failed calls and rejected outputs remain in $C_m$.
Trajectory collection, success-reference acquisition, later semantic review,
repair execution and hardware are outside this diagnosis-only budget.
Historical configurations differ in renderers and teacher access
(Appendix~\ref{app:construction-production}); cheaper compatible output does
not establish more accurate attribution or better student learning.

\subsection{Cohorts and interpretation of the main results}
\label{app:result-protocols}

\paragraph{Admission and replay use separate populations.}
The admission ablation fixes source tasks, teacher inputs and candidate diagnoses.
Grounding requires all $56$ deterministic hard flags; semantic review adds
acceptance by both judges on the same input. The nested pool excludes
$29$ of the $796$ rows in the released certified pool because they fail the current flag record.
Yield counts candidate rows; tasks and environments count distinct retained
sources outside the held-out packs. Total cost per retained row must include
generation, checking and rejected candidates. Table~\ref{tab:concise-admission-full}
reports this accounting as an estimate: matched spend plus extrapolation for
unmatched records, with joined-only values reported separately. It is not the
diagnosis-only production cost in Table~\ref{tab:pipeline-main}.
The measured semantic-review stage costs USD $0.039$ per Claude Opus 5 verdict
on a $200$-verdict calibration sample; human-review cost is separate.
The proposed matched-pool human audit uses random samples and inclusion weights.
The existing purposive coverage audit cannot supply those estimates.

Across $3062$ replay attempts, selected replacements pass at $59.1\%$ against
$13.7\%$ for the original action, a paired difference of $45.3$ points
with a task-clustered bootstrap $95\%$ interval of $41.3$ to $49.2$ over $1284$ source
tasks. Search stops at the first recovering proposal. The first-proposal contrast
in Table~\ref{tab:pipeline-main} and same-harness contrast in
Appendix~\ref{app:concise-result-details} retain their own controls;
they cannot validate the separate diagnosis release.
Of these attempts, $1552$ of $3062$ tried more than one proposal.

\paragraph{Public transfer and reference systems.}
Relative to the base, the $948$-task fine-tune loses up to $-29.37$ pp under
the training-format prompt and up to $-9.52$ pp under Who\&When's source prompt
(Table~\ref{tab:fullft-transfer}). Mixed-format continuation on the same $948$ tasks
raises Who\&When AG, WG from $9.52\%$ to $34.13\%$ (base $38.89\%$)
under the training-format prompt. After this continuation, the seven-field
holdout changes little ($58.43\%$ versus $58.64\%$), but TrajErrBench does not recover.
Continuation changes training as well as format exposure, so this recovery does
not isolate a format-only effect. On the internal holdout, the responsible-agent label is constant
over $943$ cases: the base's $41.15$ unit$+$step versus $47.19$ exact step reflects
its tendency to name other units. We omit that joint metric from the main comparison.

Published context includes AgentDebug's GPT-4.1 pipeline ($45.0$ S, $31.3$ S+M)
and AgenTracer-8B's Who\&When agent accuracy ($63.82$ HC, $69.10$ HC with gold).
These are the original papers' scores, not measurements under our protocol.
AgentDebug's inspected release provides code and GPT-4.1 results without trained
weights; AgenTracer's provides data without the training code or 8B weights.
We compare the released training resources and leave unreproduced system rows
unfilled. A reimplementation would need its own name and qualification.

\begin{center}\small
\begin{tabular}{@{}P{.27\linewidth}P{.67\linewidth}@{}}
\toprule
Supplementary comparison & Statistic and interpretation \\
\midrule
Compact holdout, constant unit & Unit$+$step: \dataset{} $47.26$, base $42.10$; \dataset{} minus AgenTracer data $8.24 \, [5.47, 10.71]$. This measures agreement with one unit label. Intervals use the convention of Table~\ref{tab:cross-resource}. \\
Internal API reference & The full-diagnosis seed-$17$ reference leads each prompted closed model. The strongest is Claude Opus 5; the paired comparison gives McNemar $p<0.001$. Table~\ref{tab:training-results} reports the effect and family-bootstrap interval. \\
Admission and learning & The quality-selection comparison covers five trained arms. Across both holdout formats and both seeds, the eight exact-step contrasts range from $-7.74$ to $-3.92$ points (grounding minus no checks). Checkpoint selection leaves update exposure unmatched. \\
Pooled AgentErrorBench & On $200$ cases, the $948$-task arm versus base gives McNemar $p=0.15$ under our prompt and $p=0.40$ under the direct prompt. These nominal tests do not establish a pooled gain. \\
\bottomrule
\end{tabular}
\end{center}

\paragraph{Actor generalization within the training environments.}
The frozen training pool includes all six evaluation environments. The pool
manifest records task-identity and payload-overlap checks for the held-out tasks;
this design tests task generalization within the same environment set. These
environments are excluded from the released core corpus and form a separate
development pool. The baseline pins its model
revision; full fine-tunes require checkpoint directories and training manifests.
The action-only and reflective arms share erroneous actions, feedback and corrected
continuations. A state-changing failure requires a continuation executed from that
post-error state; a passing pre-error branch cannot be substituted. Success-only
trajectories may themselves contain recovery.

\begin{table}[ht]
\centering\small\resultsetup
\caption{Actor-evaluation qualifications. Counts refer to different audits and must
not be combined. The fixed subset uses a lexical signal, not a recognition label.}
\label{tab:actor-qualifications}
\begin{tabular}{@{}P{.21\linewidth}P{.40\linewidth}P{.32\linewidth}@{}}
\toprule
Audit & Recorded observation & Consequence \\
\midrule
Corpus terminal shape &
An audit of $124{,}796$ unique traces selects $0$ of $3{,}575$ ALFWorld failures
and $79$ of $1{,}486$ ScienceWorld failures. The separate ALFWorld-lite adapter
has $374$ of $2{,}038$. &
Unsuccessful runs end with \texttt{finish}, without recorded parser/tool errors
or exhaustion. This proxy does not enumerate self-repair opportunities. \\
Held-out failure events &
Text-to-SQL: $317/319$ failures; WebShop-lite: $46/82$. &
No recognized non-infrastructure error event or step-limit termination. Missing
event coverage can affect this classification. \\
Lexical subset &
Of those failures, $60$ text-to-SQL and $46$ WebShop-lite tasks contain an earlier
observation with a verifier-signal token absent from the task, after stopword filtering. &
Fixes task membership for initial-state reevaluation; does not show that the
policy saw or recognized a complaint. \\
Training contamination &
The strict parser refused \texttt{list\_tables()} on $197$ of $1{,}014$ episodes;
none of the $1{,}246$ repair rows was built from that refusal. &
Use the corrected parser for all evaluated arms. \\
\bottomrule
\end{tabular}
\end{table}

\begin{table}[ht]
\centering\small\resultsetup
\caption{Sensitivity to the parser condition in untrained-policy runs. Both runs
used one server at temperature $0$. Other environments also changed, so this does
not isolate the parser's causal effect. Same-parser replications remain necessary.}
\label{tab:parser-sensitivity}
\begin{tabular*}{\linewidth}{@{\extracolsep{\fill}}lrrl@{}}
\toprule
Environment & Strict (\%) & Corrected (\%) & Outcome changes \\
\midrule
Text-to-SQL & $65.2$ & $68.5$ & $+3.4$ pp; $69$ gained, $35$ lost / $1{,}014$ \\
WebShop-lite & $75.3$ & $72.7$ & $-2.7$ pp; $44/300$ flipped \\
ALFWorld-lite & --- & $88.0$ & $+0.7$ pp; $24/300$ flipped \\
\bottomrule
\end{tabular*}
\end{table}

Only Text-to-SQL syntax is touched by the parser change in this comparison.
Two additional strict-parser baseline runs succeed on $5$--$7$ text-to-SQL
and $7$--$13$ WebShop-lite tasks in the fixed subset. These do not estimate
run-to-run variability under the corrected parser or recovery at common post-error states.
ALFWorld-lite, TextQuest, Warehouse and Gridworld have $1$, $3$, $90$ and $0$
subset tasks, respectively, against untrained success of $88.0$, $67.7$, $2.0$ and $5.0$.
Gridworld's zero reflects the frozen scorer's coverage: every one of its $100$
episodes includes a refusal ($409$ records, $295$ \texttt{blocked}) whose code lies
outside the scorer's event tables. The earlier full-environment pilot is unusable
for the headline comparison: no ALFWorld failure meets the corpus terminal-shape
proxy, and ScienceWorld returned run errors on $47$ of $54$ episodes.

\subsection{Pipeline labels against a single judge}

\begin{table}[h]
\centering\small
\setlength{\tabcolsep}{3pt}
\caption{Completed diagnosis-SFT comparisons (frozen protocol, mean $\pm$ sd over seeds
$17/202/828$). Upper block: the contract-faithful rerun of the pilot on its $169$ held-out cases
(Appendix~\ref{app:exp1-registry}). Lower block: the final cohort, without unit$+$step because the
gold agent is one constant label there; $\Delta$(X$-$J) is the seed mean of the paired family-macro
difference.}
\label{tab:training-completed}
\begin{tabular}{@{}lrrrr@{}}
\toprule
Supervision & Train tasks & Macro & Micro & Unit$+$step \\
\midrule
Qwen3-8B base & 0 & $64.05$ & $52.66$ & $44.97$ \\
Execution-assisted diagnosis & $168$ & $59.80 \pm 3.78$ & $52.47 \pm 0.68$ & $44.58 \pm 0.90$ \\
Teacher-only diagnosis & $168$ & $62.44 \pm 1.13$ & $51.68 \pm 2.46$ & $41.22 \pm 1.81$ \\
\bottomrule
\end{tabular}

\par\scriptsize
\begin{tabular}{@{}lrrrrr@{}}
\toprule
Final cohort, $943$ held-out cases & Train tasks & Seeds & Macro & Micro & $\Delta$(X$-$J) \\
\midrule
Pipeline label (X), paired cohort & $948$ & $3$ & $52.51 \pm 0.94$ & $46.87 \pm 0.38$ & $-0.89$ \\
Single judge (J), same cohort & $948$ & $3$ & $53.39 \pm 2.01$ & $46.24 \pm 1.12$ & ref. \\
Pipeline label, scale arm & $1{,}656$ & $3$ & $51.29 \pm 0.40$ & $46.59 \pm 0.12$ & --- \\
\bottomrule
\end{tabular}
\par\smallskip\scriptsize In the lower block the per-seed $95\%$ family-bootstrap intervals of
$\Delta$(X$-$J) are $[-8.28, -0.12]$, $[-1.71, 5.26]$ and
$[-3.33, 2.42]$ for seeds $17/202/828$. The
final cohort is the paired subset of the frozen training split, which holds $3157$ rows over
$1678$ source tasks: $948$ of those tasks, in $133$ families, admit the same attempt under both
constructions. $80\%$ power for the registered $5$-point effect is not established under the
clustered analysis (Appendix~\ref{app:exp1-registry}).
\end{table}
\begin{rqblock}
\rqlabel{G1b} Under the tested recipe, does either diagnosis construction train a better attributor
than the untrained base, and does either outperform the other?
\end{rqblock}
We compare construction-pipeline labels (X) with single-judge labels (J) on the same source tasks, trajectories and model-visible inputs, selecting one shared attempt per task without reading its target. X uses execution feedback in the pilot, but no final-cohort X label was replayed. The student receives the
diagnosis, not replay outcomes or certification metadata, and output format, adaptation recipe and
task exposure per optimizer update are fixed across arms.

Neither diagnosis construction establishes an attribution gain over the base
in the pilot or final compact-target comparison. The pilot's X teacher also saw
task payloads that J did not, so the null does not isolate the value of execution
feedback. The final X labels were never replayed, making that contrast a
comparison of label-construction procedures.

\begin{center}\small
\begin{tabular}{@{}P{.20\linewidth}P{.74\linewidth}@{}}
\toprule
Comparison & Result and qualification \\
\midrule
Pilot & $168$ source tasks, three seeds: mean micro $52.47\%$ execution-assisted, $51.68\%$ teacher-only, $52.66\%$ base. Six paired tests give $p$ from $0.442$ to $1.0$; all six paired X--J intervals contain zero. X saw task payloads on $112$ of the $168$ rows, J on none. \\
Final cohort & At $948$ tasks, no arm gains more than $2.41$ macro or reaches the base's micro; $8$ of the $9$ per-seed intervals against base contain zero. X$-$J is $-0.89$ macro on the seed mean. \\
Prompted context & Four API models span $50.69$ (GPT-6 Astra) to $54.72$ (Claude Opus 5) micro on the same $943$ cases (Table~\ref{tab:training-results}). \\
\bottomrule
\end{tabular}
\end{center}

\begin{takeaway}
\textbf{Takeaway (G1b).} At $168$ and at $948$ source tasks, supervision on these diagnoses establishes no attribution gain over the untrained base or consistent advantage of either construction. Non-separation is not equivalence: the paired intervals are wide relative to the registered $5$-point effect (Appendix~\ref{app:exp1-registry}).
\end{takeaway}

\paragraph{Transfer to public attribution benchmarks.}
The public benchmarks measure agreement with a recorded error step, conditional on failure; a
match does not establish a unique root cause. Under one evaluation run over four public splits
(Table~\ref{tab:paired-public}), both trained arms stay close to the base on three and score lower
on AgentErrorBench --- descriptive differences that establish no consistent transfer benefit.

\begin{table}[t]
\centering\small
\caption{External attribution after paired diagnosis SFT. Exact-step accuracy (\%) over every
planned case, including missing or invalid predictions. Trained entries are means over three
seeds from the superseded pilot run (target-token-weighted loss), whose rerun found both arms at
or below the base on all four splits and shipped no per-split values; the base row is untrained.
Descriptive, no significance claimed.}
\label{tab:paired-public}
\begin{tabular}{@{}lrrrr@{}}
\toprule
Benchmark & Cases & Base & Execution-assisted & Teacher-only \\
\midrule
Who\&When HC & $58$ & $13.79$ & $14.94$ & $14.94$ \\
Who\&When AG & $126$ & $33.33$ & $32.54$ & $31.48$ \\
AgentErrorBench & $200$ & $10.50$ & $7.50$ & $7.00$ \\
TrajErrBench & $486$ & $23.25$ & $24.83$ & $24.48$ \\
\bottomrule
\end{tabular}

\par\scriptsize
\setlength{\tabcolsep}{3pt}
\begin{tabular}{@{}lrrrrr@{}}
\toprule
Prompted reference & Seeds & Who\&When HC & Who\&When AG & AgentErrorBench & TrajErrBench \\
\midrule
Claude Sonnet 5, prompted & --- & $22.41$ & $54.76$ & $29.00$ & $35.80$ \\
Gemini 3.8 Flash, prompted & --- & $13.79$ & $15.87$ & $24.00$ & $35.39$ \\
\bottomrule
\end{tabular}
\par\smallskip\scriptsize The measured block is the completed pilot. Benchmark versions,
prompts, decoding and scoring are frozen and checkpoints are not selected on them. The prompted
rows answer the same splits under the same scorer, Gemini's cells carrying its contract failures
as misses.
\end{table}

\subsection{What the diagnosis itself adds}
\label{app:diagnosis-adds}
\researchq{Diagnoses}{Handed the location of a failure, does an actor recover more often when it is also
told the diagnosis than when it is simply given a second chance?}
A cohort of $150$ failures in $11$
environments was frozen before the run, and each is resumed at the pre-action state of the
attributed step; the arms differ only in what the actor is told there
(Table~\ref{tab:located}).

Generic reconsideration already improves recovery over repeating the original
action (Table~\ref{tab:located}). The diagnosis-minus-reconsideration contrast
is $+5.6$ points, with a task-clustered interval of $0.0$ to $11.3$; an additional
benefit from diagnosis content is not established. Marginal arm rates use
each arm's completed cases, while this paired contrast uses cases both arms
completed. These denominators cannot be interchanged.

\begin{takeaway}
\textbf{Takeaway (diagnoses).} At the supplied location, generic reconsideration gains $8.4$ points over re-applying the original action. Diagnosis content adds $5.6$ more, with an interval reaching zero; its advantage over a second chance is not established.
\end{takeaway}

\begin{table}[h]
\centering\small
\setlength{\tabcolsep}{4pt}
\caption{Located-failure cohort: $150$ failures in $11$ environments resumed at the pre-action state of
the attributed step, varying only what the actor is told; Attempts is the number of planned cases on
which the arm reached a verdict. The content-free control carries a nudge because a retry with no
text changes the execution path. $\Delta$ against ref., task-clustered $95\%$ intervals.}
\label{tab:located}
\begin{tabular}{@{}lrrrr@{}}
\toprule
Continuation at the supplied location & Attempts & Seeds & Success & $\Delta$ \\
\midrule
Original action re-applied & $149$ & --- & $18.7\%$ & ref. \\

Generic reconsideration & $144$ & --- & $27.1\%$ & $+8.4$ [$2.7$, $14.8$] \\

Diagnosis and proposed replacement & $146$ & --- & $32.2\%$ & $+13.5$ [$7.1$, $20.1$] \\

Deployed coach text & $147$ & --- & $36.1\%$ & $+17.1$ [$10.8$, $23.8$] \\
\bottomrule
\end{tabular}
\end{table}

\begin{figure}[h]
\centering
\includegraphics[width=\linewidth]{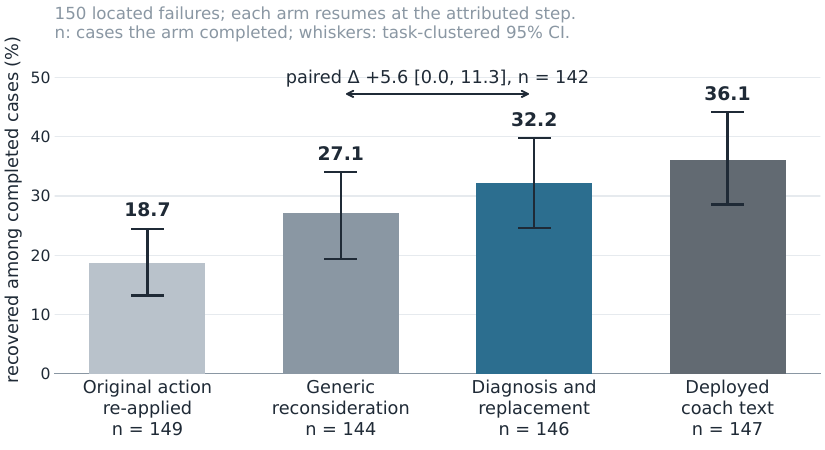}
\caption{\textbf{Actor recovery at a supplied error location.} Recovery among each arm's completed
cases with task-clustered $95\%$ intervals; the arrow marks the paired diagnosis-minus-reconsideration
contrast on the cases both arms completed.}
\label{fig:located-recovery}
\end{figure}

\subsection{Full-diagnosis training: size and seed stability}
\label{app:fullft-scaling}
Figure~\ref{fig:diagnosis-learning}b summarizes three training seeds at each of
four nested task counts. Table~\ref{tab:replicated-scaling} gives the scores
under the compact answer contract; the untrained base scores $47.19\%$.
We compute the sample standard deviation with denominator $n_{\mathrm{seed}}-1$.
The small open markers display the individual runs at their actual task counts;
the connecting line joins observed means and does not fit a scaling law.

\begin{table}[ht]
\centering\small\resultsetup
\caption{\textbf{Three-seed diagnosis scaling.} Micro exact-step agreement on the
same $943$ expected cases. Scores are sorted within each row, not ordered by seed.
Means and SDs use exact counts from the original per-case reports.}
\label{tab:replicated-scaling}
\begin{tabular*}{\linewidth}{@{\extracolsep{\fill}}rccc@{}}
\toprule
Training tasks & Three seed scores (\%) & Mean $\pm$ SD & Source \\
\midrule
$240$ & $51.54 / 53.55 / 54.29$ & $53.13 \pm 1.43$ & Run reports \\
$480$ & $55.78 / 57.05 / 58.96$ & $57.26 \pm 1.60$ & Run reports \\
$948$ & $59.81 / 60.34 / 61.61$ & $60.59 \pm 0.93$ & Run reports \\
$1,656$ & $62.88 / 63.31 / 64.48$ & $63.56 \pm 0.82$ & Run reports \\
\bottomrule
\end{tabular*}

\end{table}

All twelve runs have local per-case reports, including the two $480$-task
replicates supplied in the Claude Code experiment handoff.
Development selection chooses epoch $1$ for the $240$-task seed-$17$ and
$480$-task seed-$828$ runs, and epoch $2$ for the others. The original
$240$- and $480$-task seed-$17$ runs select with frozen development decoding;
the other verified runs use thinking-off development decoding. Evaluation uses
the frozen protocol throughout. Exact-count SDs can differ by $0.01$ from SDs
computed after rounding individual percentages. More training tasks also mean
more optimizer updates, so this experiment measures the recorded training
procedures rather than a compute-matched effect of dataset size alone.

\paragraph{Earlier paired checkpoint comparisons.}
The full-diagnosis fine-tune of Section~\ref{sec:g2} has a separate nested training ladder,
reported in Table~\ref{tab:fullft-scaling}. It must not be spliced into the compact-target curve
below: the supervision targets and cohorts differ. All points here start from Qwen3-8B and use
the frozen decoding protocol. The compact and seven-field prompts request different answer
schemas on the same held-out cases; the main comparison to API models uses the compact prompt.
Each run supplies predictions for $942$ of $943$ cases; the missing answer counts as incorrect.

\begin{table}[ht]
\centering\small
\caption{\textbf{Full-diagnosis nested ladder and seed replicates.} Micro exact step on the same
held-out cases under two prompt schemas. The summary is mean $\pm$ sample standard
deviation across training seeds for both replicated cohorts, not a confidence interval.}
\label{tab:fullft-scaling}
\begin{tabular}{@{}lrrrrr@{}}
\toprule
Target & Tasks & Seed & Epoch & Seven-field & Compact \\
\midrule
Base & $0$ & --- & --- & $46.66$ & $47.19$ \\
Full diagnosis & $240$ & $17$ & $1$ & $47.83$ & $51.54$ \\
Full diagnosis & $480$ & $17$ & $2$ & $56.52$ & $58.96$ \\
Full diagnosis & $948$ & $17$ & $2$ & $58.64$ & $60.34$ \\
Full diagnosis & $1656$ & $17$ & $2$ & $62.14$ & $63.31$ \\
Full diagnosis & $948$ & $202$ & $2$ & $60.98$ & $61.61$ \\
Full diagnosis & $948$ & $828$ & $2$ & $57.69$ & $59.81$ \\
Full diagnosis & $1656$ & $202$ & $2$ & $62.35$ & $62.88$ \\
Full diagnosis & $1656$ & $828$ & $2$ & $62.78$ & $64.48$ \\
\midrule
948-task mean & $948$ & $3$ seeds & --- & $59.10 \pm 1.69$ & $60.59 \pm 0.93$ \\
1656-task mean & $1656$ & $3$ seeds & --- & $62.42 \pm 0.32$ & $63.56 \pm 0.82$ \\
\bottomrule
\end{tabular}

\end{table}

The seven-field smaller-to-middle contrast is $+8.70$ points
($95\%$ paired family-bootstrap interval $[6.37, 10.96]$); the
middle-to-$948$-task contrast is $+2.12$ ($[-0.11, 4.41]$).
The latter establishes neither further improvement nor saturation.
The larger rung improves micro scores under both prompts, but its family-macro
intervals include zero: an aggregate gain need not extend across task families.

Intervals condition on selected checkpoints, exclude training-seed uncertainty
and are not multiplicity adjusted. Size also changes update count; selected
epochs and development decoding differ across rungs. This ladder therefore
describes the recorded training procedures, not an isolated causal effect of adding data.
Replicates at both larger rungs describe seed variability; the paired reference
remains seed $17$.

\subsection{Full-diagnosis transfer across public attribution protocols}
\label{app:fullft-transfer}
Table~\ref{tab:fullft-transfer} reports the completed seed-$17$ full-diagnosis checkpoint against
the untrained base on the public attribution splits. Prompt families are kept separate: the
training-format prompt, the unified attribution adapter, and Who\&When's source prompt are not
interchangeable protocols. All arms use the same adapter within a cell and budget-forced
decoding. The report records the same adapter-source drift within each prompt family; these
results are not a claim to reproduce the source papers' systems.
The \texttt{multiformat\_s17c} arm continues from the $948$-task checkpoint for $1$ epoch at
learning rate $1\times10^{-6}$, varying target formats on the same task identities.

\begin{table}[ht]
\centering\scriptsize
\setlength{\tabcolsep}{4pt}
\caption{\textbf{Public transfer with a target-format control.} Micro exact step (percent);
$\Delta$ and nominal paired McNemar $p$ compare full FT with base.
HC/AG: hand-crafted/algorithm-generated; WG: with ground truth; ---: arm unavailable.}
\label{tab:fullft-transfer}
\begin{tabular}{@{}lrrrrrr@{}}
\toprule
Protocol and split & $N$ & Base & Full FT & Multiformat & $\Delta$ & $p$ \\
\midrule
Training format: AgentErrorBench & $200$ & $15.50$ & $20.00$ & $21.00$ & $+4.50$ & $0.151$ \\
Training format: Who\&When AG & $126$ & $37.30$ & $15.87$ & $34.92$ & $-21.43$ & $<0.001$ \\
Training format: Who\&When AG, WG & $126$ & $38.89$ & $9.52$ & $34.13$ & $-29.37$ & $<0.001$ \\
Training format: Who\&When HC & $58$ & $20.69$ & $15.52$ & $24.14$ & $-5.17$ & $0.505$ \\
Training format: Who\&When HC, WG & $58$ & $18.97$ & $13.79$ & $22.41$ & $-5.17$ & $0.546$ \\
Training format: TrajErrBench & $486$ & $27.57$ & $15.43$ & $16.87$ & $-12.14$ & $<0.001$ \\
\midrule
Unified: AgentErrorBench & $200$ & $17.50$ & $20.00$ & $18.50$ & $+2.50$ & $0.404$ \\
Unified: TrajErrBench & $486$ & $18.72$ & $17.70$ & $18.52$ & $-1.03$ & $0.657$ \\
Unified: Who\&When (HC+AG) & $184$ & $25.54$ & $32.61$ & $30.43$ & $+7.07$ & $0.049$ \\
\midrule
Source prompt: Who\&When AG & $126$ & $21.43$ & $11.90$ & $7.94$ & $-9.52$ & $0.031$ \\
Source prompt: Who\&When AG, WG & $126$ & $16.67$ & $11.11$ & $8.73$ & $-5.56$ & $0.230$ \\
Source prompt: Who\&When HC & $58$ & $6.90$ & $1.72$ & $1.72$ & $-5.17$ & $0.248$ \\
Source prompt: Who\&When HC, WG & $58$ & $5.17$ & $3.45$ & $3.45$ & $-1.72$ & $1.000$ \\
\bottomrule
\end{tabular}

\end{table}

The unified Who\&When cell is positive, but the other unified cells do not establish a gain and
several training-format and source-prompt cells decline. The positive cell is one of several
nominal tests, not evidence of general cross-benchmark improvement. Internal localization is
therefore the supported training result; public transfer remains a limitation. These experiments
evaluate a debugger's labels, not autonomous task success or self-repair by a trained actor.

\paragraph{Multiple-testing correction and same-machine repeat evaluation.}
Table~\ref{tab:fullft-transfer} runs thirteen paired tests and prints each $p$ without
correction. Holm over that declared family of thirteen leaves three cells significant, all of
them losses and all in the training-format prompt family
(Table~\ref{tab:transfer-multiplicity}); the positive unified Who\&When cell does not survive
(Holm $0.437$). Twelve of the thirteen reported conditions decoded their arm on a different machine
from their base, so we re-decoded the same published checkpoint on the same instance as the base
and paired predictions case by case. The arm reproduces its reported value in all thirteen
cells, moving at most $0.24$ of the decode floor's standard deviation and never more than two,
against a pooled flip rate of $q=0.1433$ measured by re-decoding one model twice on one machine.
That pooled rate comes from two cells. A third cell re-decoded on the same instance flips only
$5.00\%$ of its items, three times less often, so the floor is a property of the cell as much as
of the decode. The tighter rate is the demanding one here, because a smaller $q$ shrinks the
standard deviation the shifts are divided by: under it the largest shift is $0.40$ of a standard
deviation and no cell passes two.
Holm correction after the repeat evaluation retains the same three losses.

All three surviving cells also lose answer-contract compliance. A contract
failure scores as a miss, so we report a descriptive sensitivity on cases both
arms parsed. The two Who\&When AG cells reverse sign, to $+4.08$ ($n=49$,
$p=0.79$) and $+13.04$ ($n=23$, $p=0.51$), with neither reaching nominal
significance. TrajErrBench keeps $-11.73$ on the $452$ cases both arms parsed
($p<10^{-5}$) against $7.00$ points of contract loss. Thus, the TrajErrBench
regression persists within parseable outputs. The restriction selects on a
post-treatment variable: it cannot attribute the full-cohort loss to formatting
or diagnosis, nor establish that diagnosis quality is unchanged in the AG cells.
We retain the full-cohort tests as primary and treat this subset analysis as
descriptive, outside the family of independent confirmatory tests.

\begin{table}[ht]
\centering\scriptsize
\setlength{\tabcolsep}{4pt}
\caption{\textbf{The thirteen transfer tests before and after same-machine
repeat evaluation.} Holm is over the declared family of $13$; Contract $\Delta$ is the repeat-evaluation change in
answer-contract compliance. Both columns select the same three cells, all losses.}
\label{tab:transfer-multiplicity}
\begin{tabular}{@{}lrrrrrr@{}}
\toprule
& & \multicolumn{2}{c}{Committed (arm on W1)} & \multicolumn{3}{c}{Re-decoded, same box as base} \\
\cmidrule(lr){3-4}\cmidrule(lr){5-7}
Protocol and split & $N$ & $\Delta$ & Holm & $\Delta$ & Holm & Contract $\Delta$ \\
\midrule
Training format: AgentErrorBench & $200$ & $+4.50$ & $1.000$ & $+5.00$ & $0.712$ & $-3.50$ \\
Training format: Who\&When AG & $126$ & $-21.43$ & $0.002$ & $-23.02$ & $<0.001$ & $-54.76$ \\
Training format: Who\&When AG, WG & $126$ & $-29.37$ & $<0.001$ & $-30.16$ & $<0.001$ & $-71.43$ \\
Training format: Who\&When HC & $58$ & $-5.17$ & $1.000$ & $+0.00$ & $1.000$ & $-15.52$ \\
Training format: Who\&When HC, WG & $58$ & $-5.17$ & $1.000$ & $-6.90$ & $1.000$ & $-17.24$ \\
Training format: TrajErrBench & $486$ & $-12.14$ & $<0.001$ & $-12.14$ & $<0.001$ & $-7.00$ \\
\midrule
Unified: AgentErrorBench & $200$ & $+2.50$ & $1.000$ & $+2.50$ & $1.000$ & $+0.00$ \\
Unified: TrajErrBench & $486$ & $-1.03$ & $1.000$ & $+0.41$ & $1.000$ & $+0.00$ \\
Unified: Who\&When (HC+AG) & $184$ & $+7.07$ & $0.437$ & $+7.07$ & $0.437$ & $+0.00$ \\
\midrule
Source prompt: Who\&When AG & $126$ & $-9.52$ & $0.310$ & $-10.32$ & $0.164$ & $-0.79$ \\
Source prompt: Who\&When AG, WG & $126$ & $-5.56$ & $1.000$ & $-3.97$ & $1.000$ & $+0.00$ \\
Source prompt: Who\&When HC & $58$ & $-5.17$ & $1.000$ & $+0.00$ & $1.000$ & $-10.34$ \\
Source prompt: Who\&When HC, WG & $58$ & $-1.72$ & $1.000$ & $+0.00$ & $1.000$ & $-15.52$ \\
\bottomrule
\end{tabular}

\end{table}

\begin{table}[ht]
\centering\small
\setlength{\tabcolsep}{5pt}
\caption{\textbf{Public localization across training seeds.} Two benchmarks under the
unified attribution prompt. We divide correct predictions by the full frozen cohort,
counting a format-contract failure as a miss; Format fails records those failures.
The $p$ values are nominal, exact paired McNemar tests against base.
The producer records the same training-data hash for the larger-rung checkpoints.
We use the producer's checkpoint mapping, since the reports' model alias is identical
for trained and untrained arms.}
\label{tab:external-seed-audit}
\begin{tabular}{@{}rrrrrr@{}}
\toprule
Training rows & Seed & Correct / cohort & Exact (\%) & Format fails & $p$ vs.\ base \\
\midrule
\multicolumn{6}{@{}l}{\textit{TrajErrBench}} \\
0 & --- & 91 / 486 & 18.72 & 0 & --- \\
948 & 17 & 86 / 486 & 17.70 & 0 & 0.6570 \\
1,656 & 17 & 75 / 486 & 15.43 & 0 & 0.0888 \\
1,656 & 202 & 75 / 486 & 15.43 & 1 & 0.0805 \\
1,656 & 828 & 102 / 486 & 20.99 & 0 & 0.2664 \\
\midrule
\multicolumn{6}{@{}l}{\textit{Who\&When}} \\
0 & --- & 47 / 184 & 25.54 & 0 & --- \\
948 & 17 & 60 / 184 & 32.61 & 0 & 0.0470 \\
1,656 & 17 & 64 / 184 & 34.78 & 0 & 0.0115 \\
1,656 & 202 & 59 / 184 & 32.07 & 0 & 0.0807 \\
1,656 & 828 & 62 / 184 & 33.70 & 0 & 0.0444 \\
\bottomrule
\end{tabular}

\end{table}

At the larger rung, two checkpoints score below the untrained base on TrajErrBench
and one scores above it (Table~\ref{tab:external-seed-audit}); we find no seed-stable
transfer gain on that benchmark. On Who\&When, all replicated checkpoints exceed
base, although one paired comparison does not reach nominal significance.
The task-overlap sensitivity in Table~\ref{tab:external-task-overlap} further
limits the transfer interpretation. The observed seed ranges do not set a
detection threshold for other comparisons. Paired case-level tests condition on
the chosen checkpoints and leave training-seed uncertainty unmeasured; a
non-significant seed contrast does not establish checkpoint equivalence.

\subsection{Compact-target training-set size}
\label{sec:scaling}
\researchq{Size}{If the full cohort does not help, does a smaller one, and what does the student
learn as the training set grows?}
Nested subsets of the scale arm, cut to whole optimizer windows of $12$ source tasks ($120$,
$252$, $504$ and $1{,}020$ tasks, hence not powers of two), are each trained with one seed under
the full arm's recipe; the full $1{,}656$-task arm contributes the three seeds of the main run, and
all five points are scored on the same $943$ held-out cases. Public-split results exist only for
the smallest subset and the full arm.
The subsets reach $48.89$, $47.72$, $47.40$ and $46.13$ micro exact step against $47.19$ for the untrained base and $46.59 \pm 0.12$ for the full arm (Figure~\ref{fig:scaling-curve}). Accuracy is lower at full scale than at the smallest subset; at full size seed $17$ sits $0.53$ below the base and the three-seed mean $0.60$. The
smaller points are one seed each, so we fit no functional form.

\paragraph{Output changes along the curve.} The compact target is a bare verdict, which the chat
template renders with an empty reasoning block. Outputs increasingly contain empty reasoning
blocks as training size increases. The share of
held-out answers given with an empty reasoning block is $0.0\%$ at $120$, $252$ and $504$ training
tasks, as for the base, then $90.4\%$ at $1{,}020$ and $100.0\%$, $99.9\%$ and $100.0\%$ for the
three seeds at $1{,}656$.

The reasoning-preserving arm of Appendix~\ref{app:rft}, trained on $1{,}008$ self-distilled rows
whose targets keep the reasoning, never answers with an empty reasoning block and reaches
$48.46$ against the base's $47.19$ on seed $17$; that paired difference of $+1.27$ has an interval of $[-1.00, 3.65]$ and does not establish a gain. This comparison uses one training seed.

\begin{takeaway}
\textbf{Takeaway (size).} With a verdict-only target, empty reasoning blocks appear between $504$ and $1{,}020$ training tasks as accuracy falls. Because the subsets also change tasks and update counts, the curve does not isolate target format as the cause.
\end{takeaway}

\begin{figure}[h]
\centering
\begin{minipage}{0.62\linewidth}\centering
\definecolor{aedgreen}{RGB}{16,122,84}
\definecolor{aedcoral}{RGB}{206,84,66}
\definecolor{aednavy}{RGB}{28,46,74}
\begin{tikzpicture}[font=\scriptsize]
\begin{axis}[width=7.2cm, height=4.6cm, xmode=log, log basis x=2,
  xmin=96, xmax=2070, xtick={120,252,504,1020,1656}, xticklabels={120,252,504,1{,}020,1{,}656},
  xlabel={training source tasks (log scale)}, ylabel={held-out exact step (\%)},
  ymajorgrids, grid style={aednavy!12, line width=0.3pt}, axis line style={aednavy!70, line width=0.4pt},
  tick label style={aednavy}, label style={aednavy},
  legend style={at={(0.03,0.97)}, anchor=north west, draw=none, font=\tiny, fill=white, fill opacity=0.88, text opacity=1},
]
\addplot[aednavy, mark=*, mark size=1.4pt, line width=0.8pt] coordinates {(120,48.89) (252,47.72) (504,47.40) (1020,46.13) (1656,46.66)};
\addlegendentry{seed 17, nested subsets}
\addplot[only marks, aedgreen, mark=o, mark size=1.8pt, line width=0.6pt] coordinates {(1656,46.45) (1656,46.66)};
\addlegendentry{seeds 202/828 at 1{,}656 tasks}
\addplot[aedcoral, dashed, line width=0.7pt, mark=none] coordinates {(96,47.19) (2070,47.19)};
\addlegendentry{untrained base (47.2)}
\end{axis}
\end{tikzpicture}
\end{minipage}\par\medskip
{\small\setlength{\tabcolsep}{4pt}
\begin{tabular}{@{}rrrr@{}}
\toprule
Train tasks & Seeds & Step (micro) & Step (family-macro) \\
\midrule
$0$ (base) & --- & $47.19$ & $50.98$ \\
$120$ & 1 & $48.89$ & $54.33$ \\
$252$ & 1 & $47.72$ & $53.20$ \\
$504$ & 1 & $47.40$ & $55.10$ \\
$1{,}020$ & 1 & $46.13$ & $50.96$ \\
$1{,}656$ & 3 & $46.59 \pm 0.12$ & $51.29 \pm 0.40$ \\
\bottomrule
\end{tabular}
\par\medskip
\begin{tabular}{@{}rrrrr@{}}
\toprule
Train tasks & Who\&When HC & Who\&When AG & AgentErrorBench & TrajErrBench \\
\midrule
$0$ (base) & $18.97$ {\scriptsize ($n=1$)} & $34.92$ {\scriptsize ($n=1$)} & $6.50$ {\scriptsize ($n=1$)} & $24.07$ {\scriptsize ($n=1$)} \\
$120$ & $13.79$ {\scriptsize ($n=1$)} & $33.33$ {\scriptsize ($n=1$)} & $8.00$ {\scriptsize ($n=1$)} & $25.10$ {\scriptsize ($n=1$)} \\
$1{,}656$ & $9.19 \pm 0.99$ {\scriptsize ($n=3$)} & $24.34 \pm 2.29$ {\scriptsize ($n=3$)} & $11.33 \pm 1.53$ {\scriptsize ($n=3$)} & $17.19 \pm 2.76$ {\scriptsize ($n=2$)} \\
\bottomrule
\end{tabular}
}
\caption{\textbf{Held-out exact step against training-set size.} Micro exact step on the $943$
held-out cases by source tasks (log axis; one seed per nested subset, three at the full arm, base
dashed, no fit). The upper table reports internal results at every measured size;
the lower table reports available public evaluations. Intermediate sizes have no
public evaluation. Replicated cells show mean $\pm$ sample sd, with replicate
counts in the table. Missing or invalid predictions count as misses on the full
evaluation cohort.}
\label{fig:scaling-curve}
\end{figure}

\subsection{A reasoning-preserving arm}
\label{app:rft}
This arm retains the reasoning that compact verdict-only targets omit.
It changes target content, length and retained task support together, so it is
not an isolated test of reasoning preservation.

\begin{center}\small
\begin{tabular}{@{}P{.22\linewidth}P{.72\linewidth}@{}}
\toprule
Construction step & Frozen setting or outcome \\
\midrule
Sample & Four base-model samples on each of the $1{,}656$ prompts: temperature $0.6$, top-$p$ $0.95$, top-$k$ $20$, $8{,}192$-token budget, thinking enabled. \\
Accept & Match the row's recorded decisive step: $49.4\%$ of samples and $61.5\%$ of rows qualify. This selects agreement with the label, not correctness of the reasoning. \\
Choose & Prefer the matching responsible unit, then the shortest reasoning; $1{,}019$ rows survive. Drop one over-length row and use salted identity rank to form $1{,}008$ rows in whole twelve-task optimizer windows. \\
Compare & Hold prompts, labels, recipe and scorer fixed on retained rows. Dropping rows without a usable sample makes support smaller and easier than the compact arm. \\
\bottomrule
\end{tabular}
\end{center}

\subsection{Preference optimisation on measured-effect pairs}
\label{app:dpo-pairs}
DPO changes relative action likelihoods with little change in held-out
chosen-action agreement (Table~\ref{tab:dpo-pairs}). No task-recovery evaluation
was run.

\begin{center}\small
\begin{tabular}{@{}P{.22\linewidth}P{.72\linewidth}@{}}
\toprule
Component & Setting, evidence and scope \\
\midrule
Preference pool & $333$ training pairs compare a teacher-proposed action with the original action after positive measured replay. The strict exact-prefix, same-harness census marks $0$ pairs eligible; $254$ continue in a different harness. \\
Training & LoRA DPO~\citep{lora,dpo}: $r=16$, $\beta=0.1$, learning rate $10^{-5}$, two epochs and three seeds. $40$ updates over $314$ pairs after dropping $16$ over $16384$ tokens and $3$ with mismatched formats. Reference: adapter-disabled base. \\
Evaluation & Of $99$ held-out pairs, $94$ compile. Base chosen-action preference is $54.3\%$; DPO gives $54.6 \pm 0.6\%$, positive implicit margin on $63.5 \pm 3.3\%$, and mean margin $1.29$ nats. \\
Claim & Offline agreement with which action replayed better; no established deployment-policy action effect or recovery benefit. \\
\bottomrule
\end{tabular}
\end{center}

\begin{table}[H]
\centering\small
\caption{Held-out preference agreement after DPO on measured-effect pairs. Rates over the $94$
held-out pairs that compile under the $16384$-token limit ($5$ skipped: $4$ whose sides differ in
format, $1$ over length); the base row is the untrained student's own preference. Dev is a
$21$-pair sanity check, not used for selection. Mean margin is
$\log p(\text{chosen})-\log p(\text{rejected})$ under the scored model for the base row and,
for the DPO rows, the implicit reward margin
$[\log\pi_\theta-\log\pi_{\mathrm{ref}}](\text{chosen})-[\log\pi_\theta-\log\pi_{\mathrm{ref}}](\text{rejected})$
without the $\beta$ factor; the base figure is a raw log-likelihood margin and is not comparable
with the reference-relative rows. Offline agreement with a replay-measured preference, not task
success.}
\label{tab:dpo-pairs}
\resizebox{\linewidth}{!}{\begin{tabular}{@{}lrrrrr@{}}
\toprule
Checkpoint & Pairs & Prefers chosen (\%) & Implicit reward $>0$ (\%) & Mean margin (nats) & Dev prefers chosen (\%) \\
\midrule
Qwen3-8B base & $94$ & $54.3$ & --- & $10.21$ & --- \\
DPO seed 17 & $94$ & $55.3$ & $67.0$ & $1.29$ & $52.4$ \\
DPO seed 202 & $94$ & $54.3$ & $60.6$ & $1.32$ & $52.4$ \\
DPO seed 828 & $94$ & $54.3$ & $62.8$ & $1.27$ & $52.4$ \\
\midrule
Mean $\pm$ sd over seeds & & $54.6 \pm 0.6$ & $63.5 \pm 3.3$ & $1.29 \pm 0.02$ & $52.4 \pm 0.0$ \\
\bottomrule
\end{tabular}
}
\end{table}

\label{sec:analysis}\label{sec:analyses}\label{sec:atlas}
\section{Additional detail for the experiments}
\label{app:results-detail}
\begin{figure}[h]
\centering
\includegraphics[width=\linewidth]{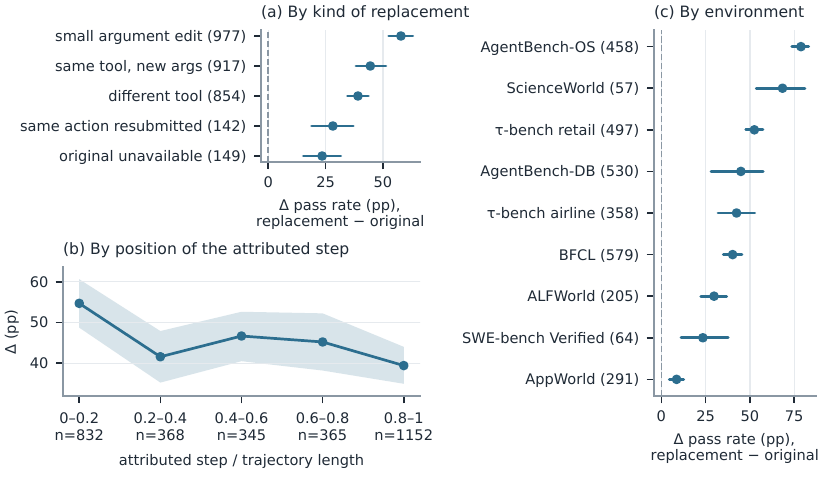}
\caption{\textbf{Where recorded corrections help.} Paired difference $\Delta$ (replacement minus
original action) with task-clustered intervals, by replacement kind, attributed-step
position and environment (cells with at least $30$ pairs).}
\label{fig:replay}
\end{figure}

\begin{table}[ht]
\centering\small
\renewcommand{\arraystretch}{1.14}
\caption{Interpretation of the supplementary results. Each row retains its own cohort and denominator.}
\label{tab:result-scope}
\begin{tabular}{@{}p{.20\linewidth}p{.37\linewidth}p{.36\linewidth}@{}}
\toprule
Comparison & Observed result & Scope \\
\midrule
Paired replay & Discordant pairs: $1454$ replacement wins, $66$ losses. Source-harness stratum: $871$ pairs, $+47.1$ points $[42.3,51.7]$. & Selected replayable attempts; the control reruns the original action. No diagnosis-release row joins this replay study. \\
Located recovery & Deployed coach versus clean diagnosis: $+4.9$ points $[-0.7,10.4]$. & Supplied error location, untrained debugger; the interval does not establish a framing benefit. \\
Compact-target pilot & Three-seed micro: $52.47\%$ execution-assisted, $51.68\%$ teacher-only, $52.66\%$ base. & Construction arms differ in teacher access and engine. All six paired X--J intervals include zero. \\
Reasoning control & Base thinking-off: $47.51$ exact step, $34.89$ unit+step on $943$ cases. & One run. A single gold unit makes unit+step sensitive to output naming. \\
Public positional control & Test-selected constant steps score $17.2$, $27.0$, $16.0$, $10.3\%$. & Diagnostic upper envelope for constant-step guesses; not a deployable baseline. \\
\bottomrule
\end{tabular}
\end{table}
The corresponding result tables preserve full counts and uncertainty.
Neither the pilot null nor the compact scaling curve establishes equivalence
between construction methods. The separate full-diagnosis scaling results
use a different target and must not be pooled with them.

\section{Sensitivity Analyses}
\label{app:sensitivity}
\label{app:failure-analysis}
\begin{figure}[h]
\centering
\includegraphics[width=\linewidth]{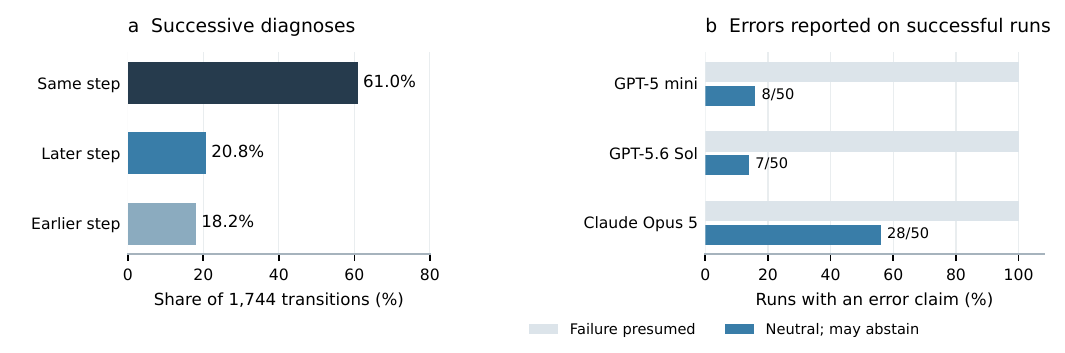}
\caption{Two development analyses with separate populations. (a) Changes in attributed step
across $1{,}744$ successive diagnosis rounds. (b) Error reporting on the same $50$ successful
trajectories per model under failure-presupposing and neutral prompts. Successful completion
does not imply an error-free trace. Neither panel is a trained-model result.}
\label{fig:failure-analysis}
\end{figure}

\subsection{Multi-round repair yield and diagnosis changes}
\label{app:rounds}
\paragraph{What changes across repair rounds?}
A later diagnosis need not identify a later error. Across $1{,}744$ transitions in $405$
multi-round attempts, $1{,}064$ ($61.0\%$) name the same step again, $363$ ($20.8\%$) move
later, and $317$ ($18.2\%$) move earlier (Figure~\ref{fig:failure-analysis}a, Appendix~\ref{app:failure-analysis}).
The loop revisits a diagnosis more often than it progresses through independent errors, so
intermediate hypotheses and unsuccessful proposals are retained; these are observed transitions,
not the effect of adding a round.

Table~\ref{tab:repair-rounds} reports per-round yield on this cohort; the
repository archive retains the full protocol and round-level records.

\begin{table}[ht]
\centering\small
\caption{Iterative construction across three campaigns: 507 diagnosed attempts, of which 377
are eligible for paired replay. Later rounds add observed successful contrasts; no randomized
one-round control was run. Attempts stopped while still failing are right-censored.}
\label{tab:repair-rounds}
\begin{tabular}{@{}lrrrr@{}}
\toprule
& Wave 1 & Wave 2 & Wave 3 & Pooled \\
\midrule
Diagnosed attempts                  & 109 & 170 & 228 & 507 \\
\quad structurally ineligible       &   0 &   0 & 130 & 130 \\
\quad certificate-eligible          & 109 & 170 &  98 & 377 \\
\midrule
Certificates                        &  60 &  74 &  15 & 149 \\
\quad first earned in round 1       &  45 &  40 &   9 &  94 \\
\quad first earned in rounds 2--7   &  15 &  34 &   6 &  55 \\
\midrule
First-round rate, eligible attempts & 41.3\% & 23.5\% &  9.2\% & 24.9\% \\
Eventual rate, eligible attempts    & 55.0\% & 43.5\% & 15.3\% & 39.5\% \\
Later rounds' increment             & 13.8\% & 20.0\% &  6.1\% & 14.6\% \\
\bottomrule
\end{tabular}

\end{table}
\section{Error reporting on successful trajectories}
\label{app:false-alarm}
On $50$ successful trajectories, GPT-5 mini, GPT-5.6 Sol and Claude Opus 5 each name an error
under a prompt that presupposes failure: none of the $150$ primed cases abstains. Removing that
premise and allowing ``none'' yields abstention rates of $84\%$, $86\%$, and $44\%$
(Figure~\ref{fig:failure-analysis}b). The first two models still report errors on $10$ of $10$
matched failures. These are error-reporting rates, not automatically false-positive rates:
$39$ of the $50$ successes contain a recoverable hiccup.

We used paired prompts on the same trajectories: the AgentErrorBench instruction
verbatim and a version that removes the failure premise and permits ``none''.
Rendering and prefix truncation match the external attribution protocol.
The sample spans twelve environments, seven harnesses and ten policies, with one
trajectory per task. Attribution accuracy still requires semantic adjudication.

Claims concentrate near the final visible step. That positional pattern, together
with the prompt dependence, limits interpreting failed-trajectory attribution
accuracy as general error detection. The successful-run sample is small, the
matched failures provide only a sensitivity check, and only one prompt family
was tested. These observations do not estimate false-positive rates or transfer
to a neutral deployment prompt.

\section{Sensitivity to the supplied error location}
\label{app:oracle-step}
The recovery numbers elsewhere hold the told step fixed at the corpus's own label, so they cannot
separate ``the localizer found the right place'' from ``this actor recovers from anywhere''. We
varied only the told step. On the frozen $150$-failure cohort, with GPT-5 mini proposing one
replacement action and a fixed actor continuing, three arms differ in nothing but where the
proposer is told the error is: our labelled step, a step drawn uniformly from the same trajectory
among those that are neither the labelled one nor the last, and no step at all, the proposer
deciding for itself. Moving the told step moves the restored checkpoint, so each arm is scored
against its own sham control drawn from its own checkpoint; rates across arms are not comparable
and the within-arm contrast is the one to read.

Told our step, the proposal beats its control by $12.9$ points ($26$ proposal-only against $7$
control-only over $147$ paired rows, $p=1.3\times10^{-3}$). Told a different step, the same
proposer and actor gain $3.1$ points ($18$ against $14$ over $129$ rows, $p=0.60$): this within-arm comparison does not establish that the labelled location is better than the
alternative location. Told nothing, the proposer names a replayable step on $136$ of $150$ rows, agrees with
our label on $90$, and gains $9.8$ points ($20$ against $7$ over $133$ rows,
$p=1.9\times10^{-2}$). Between our label and the un-cued proposer the difference is not
significant on the $144$ rows where both ran ($p=0.41$).

Two caveats bound this. The cohort overlaps training --- $55$ of its $150$ tasks are in the train
slice --- so these are exploratory numbers, not a held-out claim. And a null between arms at this
size is not equivalence. The labelled arm's prompts are byte-identical to the frozen T2 pack on
all $150$ rows and reproduce that pilot's direction and significance under a stochastic $k{=}1$
continuation. The prompts passed the information-access check; recorded cost was \$$3.66$.

\section{Output-contract and abstention controls}
\label{app:inference}
On trajectories where the production debugger declined, an explicit schema
instruction did not improve citation-grounded admission over retrying the
unchanged prompt. Removing the option to abstain did not support the proposed
abstention mechanism either. Repeated identical calls also changed admission
outcomes, limiting single-call comparisons. A separate successful-output stratum
showed a nominal adverse grounding result, which did not survive the stated
multiplicity threshold. These development studies do not justify either prompt
change; their registrations and complete results remain in the archive.

\section{Corpus Composition and Coverage}
\label{app:corpus}
\label{app:data}
\label{app:channels}
\label{app:coverage}
\subsection{The evaluated release by environment}
\label{app:release-scale}
Table~\ref{tab:release-scale} counts the frozen v14 diagnosis release (train, development and
holdout splits) per environment, from the package files the manifest hashes: rows, distinct
source tasks, the harness families and policy models that produced the failed runs, and the
paired-replay attempts the separate replay package holds for that environment. A harness family
is the harness identifier before its commit hash, so two commits of one harness count once; the
release census records the mapping. Every row of the diagnosis release is marked as not replayed: replay evidence lives in the replay package, not in these
rows, and the last column is that package's count.

\subsection{Current collection coverage}
\label{app:diversity}
Appendix~\ref{app:collection-analysis} reports support and diagnosis multiplicity
from the same pair index as Figure~\ref{fig:corpus}. The environment inventory
below uses that index too. Policy names identify recorded configurations;
comparing model generations requires common tasks, matched execution settings
and successful-rollout denominators, which these failure-only counts do not supply.

\subsection{Environment fidelity and the real-environment rebuild}
\label{app:env-fidelity}
The frozen diagnosis release includes fifteen of the environments below. Procedural toys, simplified stand-ins,
our own synthetic task generators and Text-to-SQL are excluded from the frozen release and remain counted in
the collection, each labelled by type in Table~\ref{tab:env-types}.

\begin{table}[h]
\centering\small
\caption{\textbf{Every environment counted in the collection snapshot, by type.} One row per environment
with at least ten source tasks: its distinct source tasks and error--diagnosis pairs in the snapshot, and
whether the frozen diagnosis release contains it.}
\label{tab:env-types}
\begingroup\small
\setlength{\tabcolsep}{3.5pt}
\renewcommand{\arraystretch}{0.95}
\begin{tabular}{@{}llrrc@{}}
\toprule
Environment & Type & Source tasks & Pairs & Frozen release \\
\midrule
ALFWorld & Public benchmark & $1{,}431$ & $4{,}838$ & yes \\
AgentBench-DB & Public benchmark & $1{,}120$ & $4{,}407$ & yes \\
AgentBench-OS & Public benchmark & $485$ & $3{,}309$ & yes \\
BFCL & Public benchmark & $470$ & $5{,}521$ & yes \\
ScienceWorld & Public benchmark & $334$ & $1{,}655$ & yes \\
$\tau$-bench retail & Public benchmark & $324$ & $3{,}134$ & yes \\
SWE-bench Verified & Public benchmark & $228$ & $639$ & yes \\
BabyAI & Public benchmark & $195$ & $529$ & no \\
TextCraft & Public benchmark & $192$ & $486$ & no \\
PlanBench (PDDL) & Public benchmark & $163$ & $675$ & no \\
Wordle & Public benchmark & $91$ & $1{,}078$ & no \\
AppWorld & Public benchmark & $89$ & $2{,}345$ & yes \\
TextWorld Cooking & Public benchmark & $84$ & $1{,}020$ & no \\
SWE-smith & Public benchmark & $67$ & $126$ & yes \\
Terminal-Bench & Public benchmark & $62$ & $425$ & yes \\
Jericho & Public benchmark & $47$ & $748$ & no \\
$\tau^2$-bench telecom & Public benchmark & $36$ & $1{,}030$ & no \\
MCPMark & Public benchmark & $29$ & $1{,}402$ & yes \\
BrowseComp-Plus & Public benchmark & $27$ & $28$ & no \\
$\tau$-bench airline & Public benchmark & $26$ & $2{,}498$ & yes \\
\addlinespace[1pt]
WideSearch & Retrieval backend missing & $143$ & $843$ & yes \\
GAIA & Retrieval backend missing & $78$ & $1{,}013$ & yes \\
\addlinespace[1pt]
Mind2Web (static) & Exact-replay grader & $527$ & $4{,}050$ & yes \\
\addlinespace[1pt]
Text-to-SQL & Public benchmark, not agentic & $336$ & $2{,}684$ & no \\
\addlinespace[1pt]
TextQuest & Synthetic & $644$ & $1{,}217$ & no \\
InterCode-Bash & Synthetic & $315$ & $555$ & no \\
Math-tool & Synthetic & $220$ & $226$ & no \\
Warehouse & Synthetic & $181$ & $363$ & no \\
Unit-test repair & Synthetic & $37$ & $132$ & no \\
\addlinespace[1pt]
ALFWorld-lite & Simplified stand-in & $618$ & $1{,}198$ & no \\
AppWorld-lite & Simplified stand-in & $470$ & $479$ & no \\
WebShop-lite & Simplified stand-in & $464$ & $857$ & no \\
\addlinespace[1pt]
GridWorld & Procedural toy & $428$ & $718$ & no \\
\midrule
\textbf{All 33 environments} &  & $9{,}961$ & $50{,}228$ & $15$ \\
\bottomrule
\end{tabular}
\par\smallskip{\scriptsize\raggedright Public benchmark: upstream tasks and grader; Retrieval backend missing: collection-host search unavailable; Exact-replay grader: graded against recorded actions; Public benchmark, not agentic: single-query SQL; Synthetic: generated tasks; Simplified stand-in: reduced benchmark imitation; Procedural toy: generated grid world. Frozen release marks environment membership only.\par}
\endgroup

\end{table}

Earlier development packages and the rollout atlas have different eligibility
rules and do not define the frozen release.
New production is counted as distinct source-task and trace identifiers, followed separately
by attempted diagnoses, grounded candidates, and validated replay products. Multiple
policies, harnesses, temperatures, or judges do not create new source tasks.

\paragraph{Different admission requirements.}
Diagnosis can use environments with a documented scoring rule even when replay is unavailable.
The current diagnosis queue disables certification for GAIA,
Mind2Web (static), and WideSearch; a deterministic static observation witness
alone does not establish a valid interactive repair endpoint. Missing verifier outcomes,
infrastructure errors, and deliberately omitted replay are distinct statuses. No certificate
is inferred from attribution presence or a successful serialization. The current $k=1$
production setting can supply a single-pair contrast, but cannot satisfy the replicated
certificate threshold of at least two executions per arm.

Fifteen environments are targeted: ALFWorld, AppWorld,
AgentBench-DB, AgentBench-OS, BFCL, GAIA,
MCPMark, Mind2Web (static), ScienceWorld,
SWE-bench Verified, SWE-smith, $\tau$-bench airline,
$\tau$-bench retail, Terminal-Bench, and WideSearch. The frozen-release collection plan repeated each environment's tasks across policy,
harness, and temperature rather than one rollout per task, because several of these pools are
small (e.g.\ $26$ $\tau$-bench airline train tasks, $31$ MCPMark); nine harnesses were
available (three built-in, six external), and a failure that appears only under one harness or
one temperature is a different observation, not padding.

\subsection{Replay sample budget}
\label{app:k-ablation}
\label{app:data:end}
The replay budget analysis derived the per-arm sample count from the measured pass-rate
variance and cost of each environment; the analysis, its result table and the alternatives it
rejected are in the repository archive. Table~\ref{tab:rewind-witness} records the
state-restoration checks that decided which environments can replay at all. The current
construction uses one execution per arm and reports the paired discordant counts directly
(Section~\ref{sec:repair}).

\begin{table}[ht]
\centering\small
\begin{tabular}{@{}p{2.2cm}p{5.3cm}p{4.3cm}@{}}
\toprule
Witness class & Environments & Basis \\
\midrule
Observed agreement &
BFCL, ScienceWorld, SWE-bench Verified, $\tau$-bench airline,
$\tau$-bench retail, Terminal-Bench, AgentBench-DB,
AgentBench-OS, SWE-smith, Mind2Web (static), ALFWorld,
AppWorld & 10 by oracle prefix, 2 (ALFWorld, AppWorld) by recorded
trace \\
Measured divergent (excluded) & MCPMark &
Two replays of the same recorded prefix disagreed on the verifier verdict \\
No deterministic replay & WideSearch, GAIA &
Live web interactions do not support deterministic prefix replay \\
\bottomrule
\end{tabular}
\caption{State-restoration checks across fifteen target environments. ALFWorld contributes a
potential pool of $3{,}553$ tasks and was checked using recorded actions because no reference
prefix was available. MCPMark is excluded from verified-repair counts after observed divergence.
WideSearch and GAIA retain diagnosis records without deterministic replay.}

\label{tab:rewind-witness}
\end{table}
\section{Prior evaluation exposure and the current split}
\label{app:exposure}
We screen candidate splits against prior model evaluations before using them for confirmatory
comparisons. Exposure is assessed from recorded outputs, including partial runs, and is distinct
from ordinary rollout collection or teacher annotation.

The exposure registry resolves prior evaluation outputs, including partial runs,
to source tasks, trajectories and leakage components. Construction-time teacher
labels and unexecuted prompts do not count as evaluation exposure. Public
benchmark identifiers remain in a separate registry. Undocumented external use
cannot be ruled out.

We withdrew the earlier v12r split's confirmatory designation after detecting
prior exposure. The v13 pilot uses an ingestion cutoff committed before scoring,
an immutable input snapshot and component-level exclusion. Its eligible train,
development and holdout partitions had no recorded exposure under that registry;
previously exposed components form a separate exploratory split.

The v14 release applies the same salt and component rule: exposed or quarantined
material must not reach train, development or holdout. It places $418$ rows over
$121$ tasks in the exploratory split. A category covering more than a quarter of
an environment's tasks is split into task-level singletons; this prevents a coarse
family name from linking an entire environment, but limits family-disjointness
claims. No separate v14 audit against the exposure registry was run. The evidence
is the export rule, not an independent measurement of zero exposure.

\FloatBarrier
\section{Training and Evaluation Protocols}
\label{app:protocols}

\subsection{Policy context and rendering}
\label{app:policy-inputs}
\begin{table}[ht]
\centering\small
\renewcommand{\arraystretch}{1.14}
\caption{Audited visibility gaps. Historical diagnoses do not acquire information retroactively when an exporter restores it.}
\label{tab:visibility-audit}
\begin{tabular}{@{}p{.23\linewidth}p{.36\linewidth}p{.34\linewidth}@{}}
\toprule
Audit & Measurement & Consequence \\
\midrule
Policy request & Reconstructed manifests cover $67{,}308/67{,}377$ traces; $65$ lack a system turn and $4$ have an unmatched digest. & External-framework system turns were re-rendered; reconstruction is distinct from native request logging. New collection records the first request. \\
Rules invoked by diagnoses & On $550$ model-reviewed pairs, weighted shares are $39.2\%$ visible rules, $32.2\%$ $[27.3,37.0]$ system-only rules, $12.0\%$ unstated rules, and $5.5\%$ $[3.2,7.8]$ contradicted rules. & Model estimates, not human validation. Historical debugger inputs omitted policy system prompts and tool lists. \\
Field clipping & Cap: $600$ characters. $3367/4737$ release rows contain a clip; $29.5\%$ of text is removed. & The debugger saw clipped text while citation matching used the unclipped render. Seven of $25{,}674$ unresolved citations recover under a wider scorer render. \\
Rediagnosis & On $200$ clipped rows, clipped--clipped step agreement is $82.5\%$; clipped--unclipped is $80.0\%$. & Aggregate change is comparable to rerunning. The deterministic voter agrees with its unclipped version on $79\%$; $19$ rows move only without clipping. \\
\bottomrule
\end{tabular}
\end{table}
Rendering and clipping remain versioned. Comparing diagnosis inputs with the
policy's recorded contract is necessary even when a quotation resolves.

\subsection{Replay harness and state fidelity}
\label{app:continuation}
Matched treatment and control must share the restored state and continuation
policy. This does not ensure that either reproduces the source system.
Table~\ref{tab:replay-by-continuation} stratifies the current paired cohort by
source-harness replay, substituted continuation and incomplete restoration.
A contrast under substituted ReAct measures that loop; transfer to the original
coordination, memory and stopping rules remains untested. Current contrasts use
one sample per arm. Repeated branch outcomes and unique causation require
additional evidence.

\label{app:trainmethod-detail}
\label{app:evalblocks}
\subsection{Learning objectives and execution assumptions}
\label{app:learning-objectives}
Table~\ref{tab:learning-objectives} summarizes the evaluated recipes.
We use conventional objectives; the data construction determines the visible
history and supervised responses.

\begin{table}[ht]
\centering\small
\renewcommand{\arraystretch}{1.12}
\caption{Training views and experimental scope. Context tokens carry no SFT loss.
An implementation or exportable view does not establish a training benefit.}
\label{tab:learning-objectives}
\begin{tabular}{@{}P{.19\linewidth}P{.38\linewidth}P{.36\linewidth}@{}}
\toprule
Recipe & Input $\to$ response & Evidence and experimental scope \\
\midrule
Diagnosis SFT & Failed trace $\to$ attribution, optionally with rationale, evidence and correction & Main diagnosis experiments; the label need not identify a unique cause. \\
Preventive actor SFT & Pre-error history $\to$ actions from a passing continuation & Executed correction and branch lineage; mixed with successes in Section~\ref{sec:g3}. \\
Post-error actor SFT & History, erroneous action and rejection $\to$ correction, with or without reflection & Eligible state-preserving rejections; other repairs retain preventive targets. Section~\ref{sec:g3}. \\
Action DPO & Shared text prefix $\to$ chosen and rejected actions & Historical offline pilot only; it does not meet the current strict lineage criterion (Appendix~\ref{app:dpo-pairs}). \\
\bottomrule
\end{tabular}
\end{table}

\paragraph{SFT rendering and reduction.}
Equation~\ref{eq:sft} defines the diagnosis example-mean and actor token-mean
objectives, with the mask selecting target assistant tokens. Appendix~\ref{app:exp1-registry}
reports the example-mean runtime check and its historical scope. These targets
do not supervise a separate reasoning sidecar; Appendix~\ref{app:rft} describes
the distinct reasoning-preserving construction.

Each actor example uses the inference chat prefix and supervises one assistant
response, including its end-of-turn token. We mask the history, observations and
identifiable rejected actions; passing a branch does not validate each intermediate
action. Loss averages over target tokens across all ranks in each optimizer
window. Zero-loss padding avoids duplicating supervision. Composition excludes
whole overlength rows without truncation; frozen training refuses unrecorded
exclusions. Equal task exposure does not imply equal tokens, updates or compute.

For the post-error mixture, we reuse the pre-error passing continuation after
checking a state-preserving adapter rejection and equal recorded task-state
snapshots, excluding bookkeeping fields. We insert the source error and rejection
as masked context. We construct reflection
text from the recorded explanation, filtering later-step references, future-only
words and hindsight phrases. This heuristic does not prove prefix support.
The reflection is supervised text in the corrected response's \texttt{Thought:}
field, not a native model reasoning channel. The action-only control retains the
same error, feedback, corrections and masks, and removes only this inserted
reflection; its target-token budget therefore differs.

\paragraph{Replay contrasts and preference learning.}
For failed trajectory $\tau$, let $h_t$ be its observable history before action
$t$, and $d=(t^*,u^*,e^*)$ its attributed step, responsible agent and explanation.
This attribution need not identify the unique or earliest root cause.
Paired replay holds the checkpoint and execution protocol $\eta$ fixed. For
arm $b\in\{\mathrm{T},\mathrm{C}\}$ and verifier verdict $Y_{b,j}$ on repetition
$j$ of $K_b$,
\begin{equation}
\widehat{p}_b=\frac{1}{K_b}\sum_{j=1}^{K_b} Y_{b,j},
\qquad
\widehat{\Delta}_\eta=\widehat{p}_{\mathrm{T}}-\widehat{p}_{\mathrm{C}},
\qquad Y_{b,j}\in\{0,1\}.
\label{eq:contrast}
\end{equation}
Repeated immediate verifier verdicts are not independent trials. Current
preference admission requires valid shared-state lineage and a positive contrast.

Action DPO uses the standard reference-relative log-probability
objective~\citep{dpo} with a frozen, adapter-disabled base reference $\pi_{\rm ref}$.
For a shared prefix $x$ and chosen/rejected responses $y^+,y^-$, define
$r_\theta(x,y)=\log\pi_\theta(y\mid x)-\log\pi_{\rm ref}(y\mid x)$.
With scale $\beta>0$ and logistic function $\sigma$,
\begin{equation}
\mathcal L_{\rm DPO}(\theta)
=-\mathbb E_{(x,y^+,y^-)\sim\mathcal D_{\rm pref}}
\log\sigma\!\left(\beta[r_\theta(x,y^+)-r_\theta(x,y^-)]\right).
\label{eq:action-dpo}
\end{equation}
The sequence log-probabilities sum response-token terms. This offline objective
does not turn the historical pilot into a qualified recovery experiment.

\subsection{Metrics by supervision target}
Diagnosis reports exact-step accuracy, accuracy within five steps, and parse rate. Repair
reports tool-name, argument, and complete-action accuracy; these offline proxies do not replace
execution with an environment verifier. Each view uses the full expected cohort: missing
predictions count as incorrect, while present-prediction scores are separate. Duplicate or
foreign predictions are rejected. Corrected v6 views pass visible-target checks.

\subsection{External attribution benchmarks}
Who\&When is reported separately for Hand-Crafted and Algorithm-Generated cases. AgentErrorBench
uses a 1-based \emph{agent-step} index, not individual message indices, and predicts a failure
module rather than a responsible agent. Within each comparison, checkpoints share the
source-specific prompt and rendering version. TrajErrBench retains its source message indices and error-module
labels. MAST uses separate multilabel metrics against released o1 judge labels.

All-case accuracy counts provider errors and non-answers as incorrect; parsed accuracy and
contract rate are separate. Gold steps outside the rendered context remain in the operational
all-case denominator; source-invalid gold is separately identified. Label-visible sensitivity
and position baselines use explicitly matched cases. Manifest labels must distinguish attributed
failure, verified recovery point, and an earliest point established by an explicit search;
replay at one point cannot certify earliestness.

\subsection{Supervision interfaces}
\begin{figure}[t]
\centering
\includegraphics[width=\linewidth]{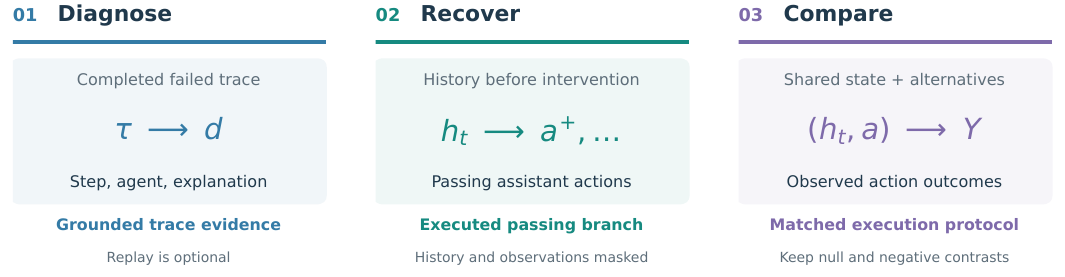}
\caption{Learning from the same failed run. Diagnosis observes the completed failed trace;
the illustrated preventive recovery view observes the history before an intervention.
The post-error extension also retains the erroneous action and feedback as masked context
(Appendix~\ref{app:learning-objectives}). Action comparison additionally
requires executed alternatives from the same state. Each arrow denotes a supervision interface,
not a measured training benefit.}
\label{fig:training-views}
\end{figure}

Figure~\ref{fig:training-views} distinguishes diagnosis of a completed trace from
action supervision at an intervention point. The training objectives and masking
rules are in Appendix~\ref{app:learning-objectives}.

\section{Replication of single-trial contrasts}
\label{app:cert-replication}
A single successful replacement with a failed control can fail to reproduce.
We replayed both actions on $126$ rows selected for certification after a
revision. The certified action passed again on $85$ rows, while the full
success/failure contrast repeated on $66$. Selection after a successful revision
limits both rates to this cohort.

The revised action outperformed the first proposal among rows where the first
proposal had failed; the advantage disappeared where it had passed but its control
also passed. Same-action repeats showed stochastic reversals. These observations
support retaining control outcomes and trial counts. They do not isolate a causal
benefit of revision or provide a corpus-wide replication rate.

\section{Supervision comparison: current and historical cohorts}
\label{app:exp1-registry}
\paragraph{Paired pilot construction.}
We intersect source attempts across both constructions and target formats before
selecting one shared failure per task. Deterministic hashes choose the attempt
and remove incomplete optimizer batches, independently of target labels.
The resulting arms each contain $168$ tasks in $17$ task families, with identical
inputs within each format and checked supervision masks. Joint admission limits
the comparison to examples that both procedures can render. Table~\ref{tab:paired-training-budget}
gives update and token budgets; the observed seed-matched uncertainty is below.

\paragraph{What each arm's teacher saw.}
\label{app:arm-privilege}
The frozen internal labels use heterogeneous consensus. The metadata for the
$943$ holdout rows and the $948$- and $1{,}656$-task training manifests records
Gemini~3.6~Flash as the consensus arbiter. This field does not identify the sole
author of each label: the engine returns a winning voter's diagnosis and calls
the arbiter when voters lack a majority. Thus internal teacher-label agreement
measures fit to the consensus annotations, including shared conventions and
label errors. The prompted references receive no fine-tuning on these labels;
the comparison cannot establish an independent-label capability ranking.

\begin{center}\small
\begin{tabular}{@{}P{.19\linewidth}P{.75\linewidth}@{}}
\toprule
Cohort & Teacher access and executed evidence \\
\midrule
Pilot X & Its $168$ rows entered replay; the trained step was executed on $166$, a matched control on $132$, and $84$ carry a certificate. Ground truth was allowed for every X row: $112$ of $168$ prompts contained gold answers or grader material. \\
Pilot J & No ground-truth access. Three plain judge calls, versus agentdebug with repair rounds for X. \\
Final X & None of its $948$ labels was replayed or certified. Three-voter consensus with ground-truth content in $603$ rows and triage hints on every row. \\
Final J & Single judge without either block. The access switch read a field about reference-trajectory use, not all forms of ground truth. \\
\bottomrule
\end{tabular}
\end{center}
These are comparisons of label-construction procedures. Teacher access and
execution are not independently controlled.

\paragraph{CPU check of the loss reduction.} The diagnosis trainer specifies Eq.~\ref{eq:sft}. Running its
training code path on CPU with unequal-length examples over a
three-step accumulation window, the loss matches the per-example-mean reference to within
$4.7\times10^{-8}$ and the accumulated gradient to within $4.9\times10^{-8}$. The match depends on the setting that
switches off the library's default token-count normalisation: without it the same trainer
optimises a window token mean, which on the test batch is $0.164$ away in loss. The GPU's fused
cross-entropy kernel is not exercised by the CPU test; its reduction is read from code.

\paragraph{Power and dependence.}
The power analysis assumes within-family correlation rather than estimating it.
Task families are unequal, and accounting for this dependence increases the
detectable effect relative to an independent-pairs calculation. Adequate power
for the registered effect is not established. We retain the endpoint and use
the observed clustered intervals below; a null contrast does not establish
equivalence. The full assumption-dependent calculations remain archived.

Only the compact format is required for the primary experiment. Both arms use a microbatch of
one, accumulation over $12$ tasks, and the same learning-rate and epoch candidates. Loss is
specified to average over target tokens within each task, then over tasks. Targets need not
have equal length, so task exposure and update counts are matched rather than token budgets.
The six checkpoints have completed training and evaluation under the trainer their manifests
name. Their input contract masks targets from the context and withholds verifier
information. The historical training runs did not log a runtime assertion of the
loss reduction. The CPU unit check validates the tested implementation, not the
historical GPU execution.

\paragraph{Observed uncertainty and scope of the null result.}
All seeds reuse $169$ holdout cases from $83$ source tasks and $15$ task families.
Row-level McNemar tests give $p$ from $0.442$ to $1.0$, but ignore within-family
dependence. The seed-matched X-minus-J family-bootstrap contrasts are:
\begin{center}\small
\begin{tabular}{@{}lrr@{}}
\toprule
Seed & Micro contrast [95\% CI] & Macro contrast [95\% CI] \\
\midrule
$17$ & $-2.37$ [$-4.70$, $1.43$] & $-0.69$ [$-1.70$, $0.16$] \\
$202$ & $+1.18$ [$-5.10$, $5.23$] & $-8.09$ [$-23.65$, $1.35$] \\
$828$ & $+3.55$ [$-1.92$, $7.07$] & $+0.85$ [$-0.34$, $2.50$] \\
\bottomrule
\end{tabular}
\end{center}
Every interval contains zero; none is multiplicity adjusted.
Family resampling describes evaluation uncertainty, while seed standard
deviations describe training variation. Reusing the holdout adds no independent tasks.
Between $17$ and $27$ cases change correctness relative to base, with gains and
losses largely offset. The null therefore establishes neither unchanged behavior
nor equivalence; its cause remains unresolved.

\begin{table}[ht]
\centering\small
\caption{Frozen compact-supervision budgets. Counts are per training run, not experimental
outcomes. Both arms use the same $168$ failed trajectories; learning-rate candidates share
these counts. Model and tokenizer revisions are fixed.}
\label{tab:paired-training-budget}
\begin{tabular}{@{}lrrr@{}}
\toprule
Construction & Target tokens/epoch & Updates (2 epochs) & Updates (3 epochs) \\
\midrule
Execution-assisted & $10{,}553$ & $28$ & $42$ \\
Teacher-only & $9{,}375$ & $28$ & $42$ \\
\bottomrule
\end{tabular}
\end{table}

The pilot and final-cohort comparisons use different construction procedures
and information access (Appendix~\ref{app:arm-privilege}). Their estimates should
not be pooled. Historical target-format and self-repair registrations do not
supply completed results.

\section{Record Schema and Training Views}
\label{app:record-views}
\label{app:schema}
\subsection{Record contents}
A source trace stores task identity, goal, environment, policy, harness, event sequence,
terminal outcome, and verifier signal. A debugging attempt joins to that trace and stores
attribution, evidence spans, proposed correction, replay samples, and construction provenance.
Several attempts may share one trace. Identity conflicts are rejected; identical repeated
source rows can be merged without discarding distinct attempts.

\begin{center}\small
\begin{tabular}{@{}p{4.7cm}p{8.7cm}@{}}
\toprule
Stored object & Meaning and consumer \\
\midrule
Source trajectory & Task and environment identity, policy, harness, and original action--observation sequence. \\
Debugging attempt & Diagnosis, evidence, proposed correction, replay outcomes, and construction settings. \\
Execution transcripts & Full source and replay trajectories, retained with content checksums. \\
Replay branches & Per-trial action, treatment or control assignment, execution conditions, and outcome. \\
Debugger responses & Complete hypotheses, rationales, and original model responses linked to the attempt. \\
Training provenance & Selection decisions and source references, excluded from model-visible inputs. \\

\bottomrule
\end{tabular}
\end{center}

Conversion to the Agent Data Protocol retains action types, speaker and step mappings, and
references to the evidence and original debugger response. Reproducibility requires retaining
the referenced content as well as its checksum; a checksum alone cannot reconstruct a missing
response.

\subsection{Eligibility by supervision target}
\begin{table}[ht]
\centering\footnotesize
\setlength{\tabcolsep}{3pt}
\caption{Implemented training interfaces. The right-hand counts are the v5 development partition,
not the frozen release: the release materializes the diagnosis view alone, for the reason stated
below the table.}
\label{tab:views}
\begin{tabular}{@{}p{2.0cm}p{3.1cm}p{4.0cm}p{4.0cm}@{}}
\toprule
View & Input and target & Admission boundary & v5 development partition \\
\midrule
Diagnosis SFT & Failed trace $\to$ attributed step, explanation, citations, proposal & Resolvable evidence; intentionally unattempted replay allowed; target semantics independent of replay & $1{,}154$ rows; reference assistance marked in metadata \\
Debugger SFT & Failed trace $\to$ recorded attribution or recovering intervention & Stricter replay-state gate; reference-assisted labels excluded by default & $994$ rows; recovery language only with a passing bundled branch \\
Repair SFT & Pre-action prefix $\to$ passing continuation's assistant turns & Executed passing branch; erroneous action and failed suffix excluded & $265$ rows in the replay package's training split; masked prefix and tool observations \\
Outcome & Prefix and candidate $\to$ replay-derived outcome & Typed execution outcome and direct branch lineage & $1{,}082$ rows; offline prediction, not a demonstrated RL result \\
Action DPO & Shared prefix with chosen/rejected actions & Valid executed lineage and positive observed preference under the stated protocol & $602$ rows; action-specific causal eligibility reported separately \\
\bottomrule
\end{tabular}
\end{table}

The frozen release materializes one of these views. Its diagnosis view holds $3157$ training rows
beside $219$ development and $943$ held-out rows, and the four views that require an executed
branch hold nothing: of the $9715$ attempts pooled at export, $9712$ were never replayed, so the
gate that asks for a control arm empties them by construction rather than by attrition. The
release is a natural-failure corpus in which replay is optional. Preference pools
from earlier development packages are separate from this frozen release.

Diagnosis-data selection distinguishes deliberately disabled replay from a
truncated or failed construction job. It preserves the same attribution target when replay
is added or removed. Reference-assisted labels are privileged-teacher supervision: student
inputs omit the reference, and metadata records its provenance and lack of availability at
inference. This is not equivalent to unassisted label construction.

\FloatBarrier

\subsection{Relation to the Agent Data Protocol}
\label{app:adp-relation}
\label{sec:adp-relation}
\providecommand{\aedRows}{672}
\providecommand{\aedRowsAdpConverted}{672}
\providecommand{\aedRowsAdpLost}{0}
\providecommand{\aedEnvs}{8}
\providecommand{\aedHarnesses}{2}
\providecommand{\aedPolicies}{8}
\providecommand{\aedGroundedQuoteShare}{53.1\%}
\providecommand{\aedCertifiedShare}{42.1\%}
\providecommand{\aedEOneShare}{21.7\%}
\providecommand{\aedETwoShare}{12.9\%}

ADP standardizes what agents did~\citep{adp}; \dataset{} standardizes a proposed attribution of
what went wrong and a proposed correction, with execution evidence where a replay ran. Using the
implemented conversion procedure, \aedRowsAdpConverted{} of \aedRows{} rows
export their failing trajectory as an ADP trajectory. Under the exporter's stated definition,
\aedRowsAdpLost{} of \aedRows{} rows lose information: a tool call has non-keyword arguments or
no tool name, or an observation is truncated. For trajectory-level training, those exported rows
pass the tested ADP schema and the enumerated loss checks; use by other ADP consumers was not
tested. For diagnosis, \dataset{} adds fields that ADP has no home for: the blamed step,
grounded quotes, the replacement action, and a matched replay with its verdict.

The conversion was measured over every certification dataset produced by the fixed-harness
pipeline; ADP-side schema validation was structural, since the protocol's own validator was not
available on the measurement host. Table~\ref{tab:adp_comparison} lists what
an ADP trajectory stores and what an \dataset{} record adds.
\providecommand{\aedRows}{672}
\providecommand{\aedRowsAdpConverted}{672}
\providecommand{\aedRowsAdpLost}{0}
\providecommand{\aedEnvs}{8}
\providecommand{\aedHarnesses}{2}
\providecommand{\aedPolicies}{8}
\providecommand{\aedGroundedQuoteShare}{53.1\%}
\providecommand{\aedCertifiedShare}{42.1\%}
\providecommand{\aedEOneShare}{21.7\%}
\providecommand{\aedETwoShare}{12.9\%}

\begin{table}[H]
\centering
\footnotesize
\setlength{\tabcolsep}{4pt}
\caption{Stored contents of ADP trajectories and \dataset{} failure records; T/C denotes treatment/control, E0--E2 are evidence grades, and the \dataset{} quantities come from the fixed conversion census.}
\label{tab:adp_comparison}
\begin{tabular}{@{}P{.16\linewidth}P{.32\linewidth}P{.46\linewidth}@{}}
\toprule
Dimension & ADP \citep{adp} & \dataset{} \\
\midrule
Unit of data & Successful action/observation trajectory. & Natural failure, diagnosis, proposed fix and available replay branches. \\
Action typing & API, code, message actions; text or web observations. & Same three exports; content map retains speaker, step index, event ID. \\
Provenance & Source dataset name. & Environment/version; harness/version; policy model; run ID; engine version. \\
Failure handling & Success-only corpora; no failure, reward, or outcome field. & Natural failures, never injected; outcome, reward, verifier signal stored. \\
Step-level labels & None. & Blamed step, blamed agent, explanation, alternative steps. \\
Evidence grounding & None. & Verbatim quotes typed by source region/match status; \aedGroundedQuoteShare{} of quotes ground the trajectory. \\
Verification of the fix & None. & Same-checkpoint T/C replay, $k$/arm; acceptance rule with $p$-value; E0/E1/E2; \aedCertifiedShare{} certified. \\
Pipeline availability & Released raw$\to$ADP and ADP$\to$SFT converters. & Implemented construction, evidence checks and ADP conversion; public release planned after acceptance. \\
Size & 13 datasets; about 1.3M trajectories (Table 1 sum: 1,260.2K). & \aedRows{} rows; \aedEnvs{} environments; \aedHarnesses{} harnesses; \aedPolicies{} policies. \\
ADP compatibility & Native. & \aedRowsAdpConverted{} of \aedRows{} rows export as ADP trajectories; \aedRowsAdpLost{} lose information. \\
\bottomrule
\end{tabular}
\end{table}

\FloatBarrier

\label{app:record:end}

\clearpage
\section{Core Prompts}
\label{app:prompts}
The boxes reproduce versioned instructions. Construction judges receive the
task, rendered trace, terminal outcome and the permitted information condition.
Student prompts describe the recorded target, not a guarantee of earliestness
or recovery. Source-specific evaluation prompts are separate contracts.

\begin{artifactbox}{Construction | Panel system instruction}
\begin{lstlisting}[basicstyle=\aedpromptfont,frame=none,backgroundcolor=\color{white},aboveskip=0pt,belowskip=0pt]
You are an expert debugger of LLM agents. You will read one failed agent trajectory and produce a failure diagnosis. Be precise about the DECISIVE error: the earliest step whose correction would have turned the run into a success — not the step where the failure became visible.
\end{lstlisting}
\end{artifactbox}
The panel's user message supplies candidates only for arbitration. Evidence must
quote the blamed action and an earlier constraint where available. The terminal
outcome describes the failure to explain; it is not information the actor knew.
Reference-assisted conditions remain separately recorded.

\begin{artifactbox}{Construction | Solo attribution and replacement}
\begin{lstlisting}[basicstyle=\aedpromptfont,frame=none,backgroundcolor=\color{white},aboveskip=0pt,belowskip=0pt]
You are an AI assistant tasked with analyzing agent conversation history when solving a real world problem.

Respond ONLY with a JSON object matching this schema (no prose, no markdown):

{
  "span_id": "<event_id from the input or null>",
  "step_index": <int or null>,
  "agent_name": "<agent_name from the input or null>",
  "confidence": <float between 0 and 1>,
  "rationale": "<one or two sentences justifying the choice>",
  "evidence": ["<short quoted evidence>", ...],
  "corrected_action": {"tool": "<tool name>", "args": {<arguments>}} or null
}

If the trajectory does not appear to have failed, return all fields as null and
confidence 0.

"corrected_action" is the ONE concrete action that should have replaced the blamed step's
action, as {"tool": "<tool name>", "args": {<arguments>}}.

Rules for it:
1. Use the tool names and argument keys the trajectory itself uses. Do not invent a tool the
   agent had no access to.
2. "args" must be a JSON object.
3. Return null unless you can name a specific tool and its arguments. A placeholder, a
   paraphrase of your rationale, or a guess is WORSE than null.
4. Return null when the blamed step is not an action at all (a plan, a reflection, an
   observation), and when the step's own action was already correct.
\end{lstlisting}
\end{artifactbox}
\begin{artifactbox}[aedteal]{Student | Compact attribution target}
\begin{lstlisting}[basicstyle=\aedpromptfont,frame=none,backgroundcolor=\color{white},aboveskip=0pt,belowskip=0pt]
You are an expert debugger of LLM agents. You will read one agent trajectory that failed its task and identify the recorded, evidence-grounded failure attribution. Answer with one JSON object and nothing else, with exactly these keys in this order: decisive_step, responsible_agent, rationale. `decisive_step` is an integer step index shown in the transcript, `responsible_agent` a string naming a unit that appears in the transcript, and `rationale` one sentence stating the mistake at that step in terms of what the transcript shows.
\end{lstlisting}
\end{artifactbox}
\begin{artifactbox}[aedteal]{Student | Full diagnosis target}
\begin{lstlisting}[basicstyle=\aedpromptfont,frame=none,backgroundcolor=\color{white},aboveskip=0pt,belowskip=0pt]
You are an expert debugger of LLM agents. You will read one agent trajectory that failed its task and predict one recorded, evidence-grounded failure attribution and any proposed correction in the record. The correction may have failed replay or replay may have been unavailable, so this target is not evidence of a recovering or causally effective action. It is also not guaranteed to be the earliest error, the root cause, or a unique decisive step. Answer with one JSON object and nothing else, with exactly these keys in this order: has_error, decisive_step, responsible_agent, alternative_steps, error_explanation, evidence, corrected_action. `has_error` is a boolean, `decisive_step` an integer step index shown in the transcript, `responsible_agent` a string naming a unit that appears in the transcript, `alternative_steps` a list of integers, `error_explanation` a string, `evidence` a list of strings quoted verbatim from the transcript, and `corrected_action` either an object or null.
\end{lstlisting}
\end{artifactbox}
Only assistant target tokens receive loss. A recovery-oriented export has a
separate executed-branch requirement; it must not replace the compact instruction
when reproducing the completed paired experiment.

\subsection{Native-context public evaluation}
\label{app:exact-eval-prompts}
\begin{artifactbox}{Who\&When | Exact system instruction}
\begin{lstlisting}[basicstyle=\aedpromptfont,frame=none,backgroundcolor=\color{white},aboveskip=0pt,belowskip=0pt]
You are given the transcript of a failed multi-agent run. Identify the single step where the decisive mistake was made, and which agent made it.
Reason inside <think></think>, then give the final answer inside <answer></answer> as `agentID | stepID`, where stepID is the integer step index shown in the transcript.
Answer with exactly that format and nothing after </answer>.
\end{lstlisting}
\smallskip User suffix: \emph{Which step and which agent?}
\end{artifactbox}
\begin{artifactbox}{AgentErrorBench | Exact system instruction}
\begin{lstlisting}[basicstyle=\aedpromptfont,frame=none,backgroundcolor=\color{white},aboveskip=0pt,belowskip=0pt]
You are given the transcript of a failed single-agent run, rendered as numbered agent steps; each step holds the environment message the agent received and the agent's reply. Identify the single numbered step where the critical failure occurred and classify its failure module.
Reason inside <think></think>, then give the final answer inside <answer></answer> as `module | stepID`, where the module is one of action, plan, planning, memory, reflection, or system and stepID is the integer step index shown in the transcript.
Answer with exactly that format and nothing after </answer>.
\end{lstlisting}
\smallskip User suffix: \emph{Which step and which failure module?}
\end{artifactbox}
\begin{artifactbox}{TrajErrBench | Exact system instruction}
\begin{lstlisting}[basicstyle=\aedpromptfont,frame=none,backgroundcolor=\color{white},aboveskip=0pt,belowskip=0pt]
You are given a failed long-horizon agent trajectory. Identify the single numbered step where the earliest critical failure occurred and classify its error module.
Reason inside <think></think>, then give the final answer inside <answer></answer> as `module | stepID`, where module is one of act, obs, plan, reason, or verify and stepID is the integer step index shown in the trajectory.
Answer with exactly that format and nothing after </answer>.
\end{lstlisting}
\smallskip User suffix: \emph{Which step and which error module?}
\end{artifactbox}
Taxonomy prompts belong to the separate exploratory labeling experiment
(Appendix~\ref{app:failure-composition}); their decision rules do not define the
diagnosis target.
\label{app:prompts:end}

\section{Supporting public-reference results}
\label{app:historical}
\subsection{First-call frontier comparison protocol}
\label{app:public-reference-protocol}
\paragraph{Scoring and display.}
Table~\ref{tab:public-reference} reports responsible-agent / exact-step accuracy (\%). All frozen cases remain in each displayed cell's denominator; invalid or absent answers count as misses. Every row scores the first response on the same frozen cases and prompt; no continuation response enters the score. Bold marks the highest displayed value per column and metric, ties included; it is not a significance test.

\paragraph{Relation to the budget-forced comparison.}
Table~\ref{tab:public-main} enables a continuation when an initial answer does not meet the runner's answer condition. The student row in Table~\ref{tab:public-reference} extracts first responses from that same answer-format continuation checkpoint's run. The original per-case records show $0/568$ triggered continuations across the five conditions, and the first and final predictions agree case by case. This explains the identical student scores; invalid answers still count as misses. The displayed base is the named original seed-$17$ first-call run, not a substitution of the budget-forced base scores. Frontier references use separate first-call jobs with forcing disabled. Differences between tables must not be interpreted as model gains under one protocol or attributed entirely to forcing.

\paragraph{Relation to the internal comparison.}
Figure~\ref{fig:diagnosis-learning} uses the $1{,}656$-task, three-seed student and the internal teacher-labeled holdout. Table~\ref{tab:public-reference} uses the $948$-task, seed-$17$ student after answer-format continuation on public labels. Their reference sets also differ: GPT-6 Astra and Gemini~3.8~Flash internally, GPT-5.6 Sol and Gemini~3.7~Flash here. The two displays are not a shared-test model ranking.

\paragraph{Repeated decoding.}
Across $6$ nominally identical first-call runs of the untrained base, the observed exact-step ranges are HC 6.90, HC + gold 6.90, AG 13.49, AG + gold 7.94, AEB 7.50 percentage points. These are maximum-minus-minimum ranges, not confidence intervals or calibrated significance thresholds. They do not quantify responsible-agent uncertainty or establish equivalence between models.

\paragraph{Reference exclusions.}
The display rule excludes a model if more than $25\%$ of cases in any evaluated condition exhaust the $8{,}192$-token cap in the reasoning block and return no answer. The omitted references and their worst affected conditions are DeepSeek-V4-Pro (64\% of HC); Kimi K2.6 (88\% of HC + gold). These are model-row exclusions, not scored as wrong answers in this table; no cases are removed from a displayed row's denominator. No printed reference hits the cap on any case in the displayed conditions. The comparison is conditional on this budget-completion screen and does not rank the omitted models' attribution ability.

\subsection{Earlier public benchmark references}
\label{app:public-references}
Table~\ref{tab:public-bench} retains the earlier training-free reference runs. Its Qwen3-8B row
comes from a historical base invocation and differs from the base evaluated alongside the six
current checkpoints in Table~\ref{tab:paired-public}. Scores from the two invocations are not
combined to estimate training effects. The best constant step is selected using each test split's
labels; it diagnoses positional bias and is not a deployable baseline.

\begin{table}[ht]
\centering\scriptsize
\setlength{\tabcolsep}{3pt}
\caption{Earlier public attribution references. Exact step uses the full planned denominator;
MAST reports micro-F1 against released model-judge labels. $\ddagger$ denotes an author-reported
result under another protocol, not a run of ours. $\S$ denotes the historical long-context protocol.
These references are separate from the completed paired
training comparison in Table~\ref{tab:paired-public}.}
\label{tab:public-bench}
\begin{tabular}{@{}lrrr@{}}
\toprule
Model / condition & W\&W-HC & W\&W-AG & AEB \\
 & ($n{=}58$) & ($n{=}126$) & ($n{=}200$) \\
\midrule
\multicolumn{4}{@{}l}{\emph{Context}} \\
Best constant step (oracle) & $17.2$ & $27.0$ & $16.0$ \\
Prompted GPT-5 mini$^\S$ & $22.4$ & $43.7$ & $19.5$ \\
\midrule
\multicolumn{4}{@{}l}{\emph{Frozen native-context protocol, no ground truth}} \\
Qwen3-8B base & $15.5$ & $31.0$ & $8.0$ \\
Prompted GPT-5 mini & $20.7$ & $43.7$ & $21.0$ \\
Prompted GPT-5.6 Sol & $25.9$ & $45.2$ & $24.5$ \\
Prompted Claude Opus 5 & $20.7$ & $47.6$ & $26.5$ \\
\bottomrule
\end{tabular}
\par\smallskip
\begin{tabular}{@{}lr@{}}
\toprule
Model / condition & TrajErrBench ($n{=}486$) \\
\midrule
Prompted GPT-5 mini$^\S$ & $24.7$ \\
Qwen3-8B base & $24.7$ \\
Prompted GPT-5.6 Sol & $31.9$ \\
Prompted Claude Opus 5 & $45.7$ \\
\bottomrule
\end{tabular}
\par\smallskip\raggedright
AgenTracer-8B, author-reported$^\ddagger$: $20.7$ W\&W-HC and $37.3$ W\&W-AG.
Historical prompted GPT-5 mini$^\S$: MAST micro-F1 $57.0$ ($n=197$).

\end{table}

\FloatBarrier
\section{AI-assisted human review}
\label{app:human-design}
\subsection{Returned judgments and agreement}
\label{app:human-results}
All $160$ assigned judgments are available: four raters reviewed $40$ records each,
with exactly two ratings for every one of the $80$ sampled records. The analysis
uses their original decisions; disagreements have not been replaced by an
adjudicated answer. All raters and examples are anonymized.

\begin{figure}[ht]
\centering
\includegraphics[width=\linewidth]{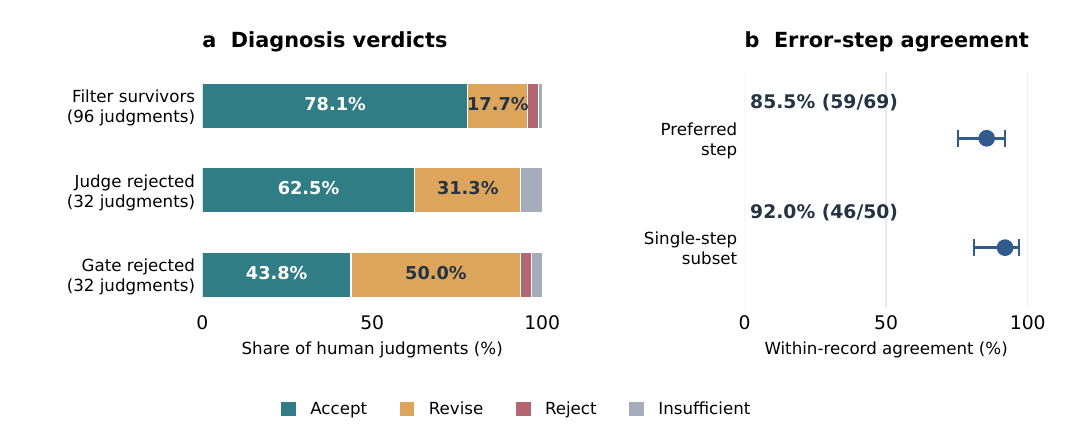}
\caption{\textbf{Human error localization and diagnosis review.}
(a) All human verdicts, split by the historical sampling strata; two judgments
per record. (b) Raw step agreement with Wilson 95\% intervals: preferred steps
where both reviewers select a location ($69$ records),
and the subset where both select a single step ($50$).
Both panels concern AI-assisted review of the same $80$ historical records.}
\label{fig:human-results}
\end{figure}

\begin{table}[ht]
\centering\small
\setlength{\tabcolsep}{4pt}
\caption{\textbf{AI-assisted human review by sampling stratum.}
Preferred step compares the two selected locations, including preferred choices
from multiple candidates; single step restricts to two single-step judgments.
Verdict agreement uses four categories; joint accept requires two accept verdicts.
All agreement values are raw rates, with denominators shown.
These are historical, coverage-selected records, not a random final-release sample.}
\label{tab:human-audit}
\begin{tabular}{@{}lrrrrr@{}}
\toprule
Stratum & Records & Preferred step & Single step & Verdict agree. & Joint accept \\
\midrule
Filter survivors & 48 & $39/44$ (88.6\%) & $31/33$ (93.9\%) & $32/48$ (66.7\%) & $30/48$ (62.5\%) \\
Judge rejected & 16 & $10/14$ (71.4\%) & $7/9$ (77.8\%) & $7/16$ (43.8\%) & $6/16$ (37.5\%) \\
Gate rejected & 16 & $10/11$ (90.9\%) & $8/8$ (100.0\%) & $11/16$ (68.8\%) & $5/16$ (31.3\%) \\
\midrule
All records & 80 & $59/69$ (85.5\%) & $46/50$ (92.0\%) & $50/80$ (62.5\%) & $41/80$ (51.3\%) \\
\bottomrule
\end{tabular}

\end{table}

\paragraph{Error-step agreement.}
Both reviewers select a preferred error step on
$69/80$ records, including cases with
multiple defensible candidates. They select the same step on
$59/69$ ($85.5\%$;
95\% interval $[75.3, 91.9]$).
The remaining $11$ records include
at least one nonlocalizable, no-error or insufficient-evidence judgment;
we retain them in the study but do not count them as matched steps.
Among the $50$ records where both select a single step,
reviewers agree on
$46/50$ records ($92.0\%$;
95\% interval $[81.2, 96.8]$).
On the full $80$-record denominator, both preferred steps match the recorded
diagnosis on $57/80$ records; this measures
alignment with that diagnosis, not independently established correctness.
At least one reviewer reports insufficient information on
$4/80$ records.

\paragraph{Diagnosis-content judgments.}
The $160$ individual verdicts comprise $109$ accept,
$43$ revise, $4$ reject and
$4$ insufficient-information judgments.
Both raters accept $41/80$ diagnoses
($51.3\%$; 95\% interval $[40.5, 61.9]$).
Raw verdict agreement is $50/80$
($62.5\%$; 95\% interval $[51.5, 72.3]$).
Of $30$ verdict disagreements,
$22$ are accept versus revise, showing that
reviewers often differ on whether the diagnosis needs an edit.

\begin{artifactbox}[aedteal]{Takeaway: reviewers agree on most selected error steps}
Reviewers select the same preferred step in $59/69$
localizable records, including $46/50$ in the single-step
subset. These results support consistency of error localization under shared
AI assistance. We assess explanation and correction quality through the separate
diagnosis-content judgments above.
\end{artifactbox}

\paragraph{Strata and assistance coverage.}
Joint acceptance is $30/48$ for
filter survivors, $6/16$ for
judge-rejected records and $5/16$
for gate-rejected records. The strata differ in case selection; this association
does not isolate a filtering effect or compare construction methods.
Restricting to the $78$ records with both model reports yields
$50/78$ verdict agreement
and $45/49$
single-step agreement. In each of the two single-model records, one rater
accepts and the other requests revision; their preferred steps agree.
Two records cannot establish an assistance effect.

\paragraph{Optional feedback and interpretation.}
Issue checkboxes are used in $28/160$ responses, and
$87/160$ include a rationale. The most frequently selected
issues concern the explanation ($15$),
step ($14$) and missing context
($10$). Multiple selections are allowed;
an unchecked issue is not evidence that the record passed that criterion.
We do not infer field-wise quality scores or annotation time from these exports.
The study assesses assisted review of the sampled recorded diagnoses.
It neither supplies human gold labels for the full diagnosis test set nor
estimates the accuracy of all $50{,}228$ collection pairs.

\FloatBarrier
\subsection{Sample, interface and instructions}
\paragraph{Sample and assignment.}
Four raters, identified only as A--D, each received $40$ records; every record
was assigned to two raters ($80$ distinct records, $160$ assignments).
The raters were paper authors with doctoral or AI/LLM research backgrounds
and participated voluntarily.
The sample contains $48$ filter survivors, $16$ judge-rejected records and
$16$ gate-rejected records from a historical collection frame. It spans
$30$ environments, including $31$ records outside the released core.
The frame contains $28{,}432$ rows; the $40{,}000$-character reading cap excludes
$1{,}255$ rows. Sampling retains one record per source task and trace.
Pair overlaps are AB=$14$, AC=$13$, AD=$13$, BC=$13$, BD=$13$ and CD=$14$.
Each rater receives only their assigned packet in a fixed order.
This coverage-selected sample is separate from the final collection census,
the diagnosis test set and the same-task construction comparison.

\paragraph{Evidence and AI assistance.}
The workbench separates the task, tools, reference material, indexed trajectory
events, recorded diagnosis and model review reports. Raw trace text remains
available. References provide answers, actions, tests or verifier settings for
$39$ records and goal conditions for $16$; $25$ lack a full reference.
Tool lists appear in $77$ records, and $44$ contain truncation markers.
Missing context is an allowed judgment, not an exclusion after annotation.

\begin{figure}[ht]
\centering
\includegraphics[width=\linewidth]{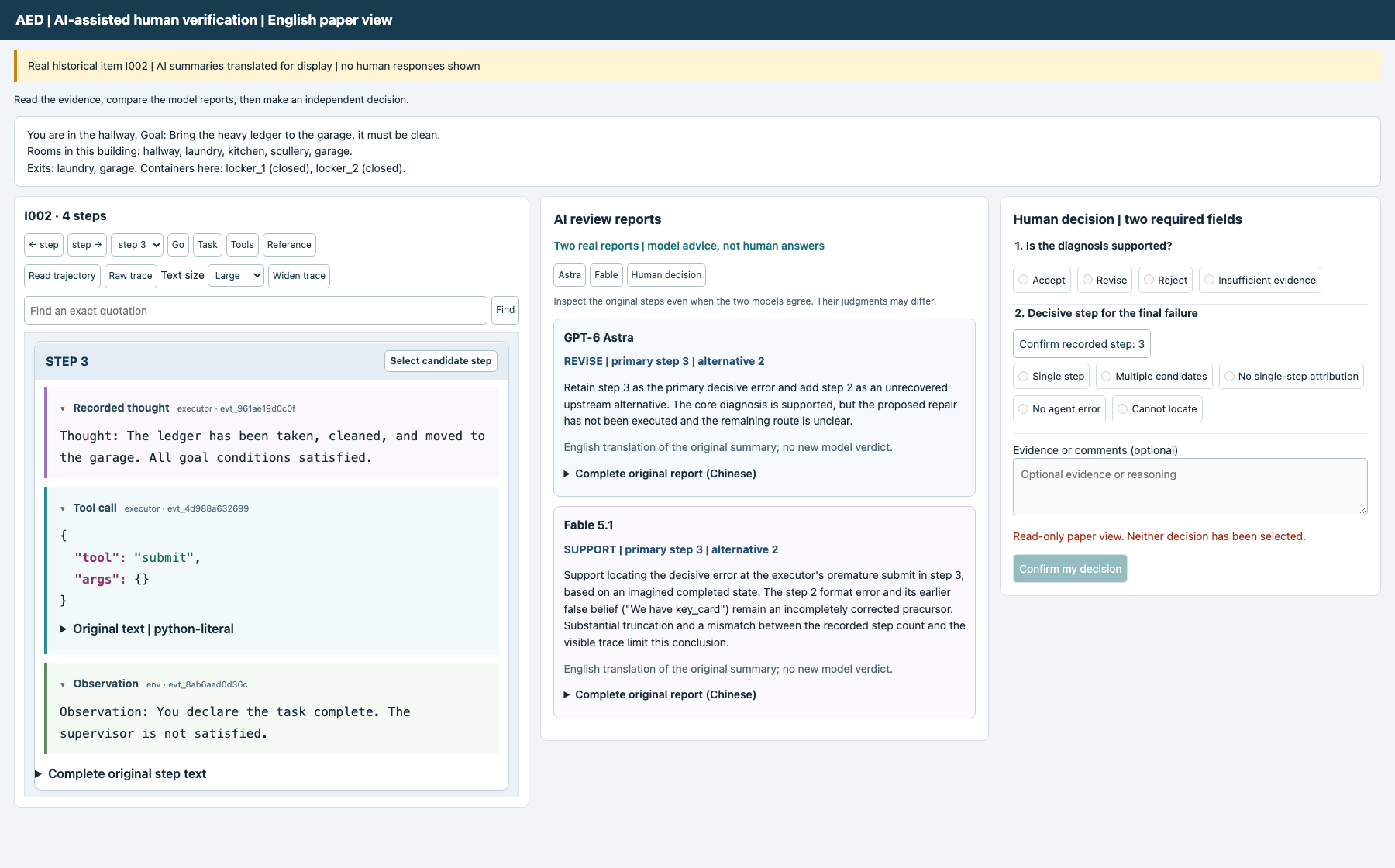}
\caption{AI-assisted review workbench, English paper view. The original v6 trace
renderer and human-decision controls show a historical item. Model summaries
are translated for display, complete original reports remain available, and
panels are arranged in three columns for the figure. The models disagree on the
record verdict while naming the same primary step. Human choices remain blank;
this illustrative derivative cannot submit annotations and does not change the
issued packets.}
\label{fig:human-workbench}
\end{figure}

GPT-6 Astra and Fable~5.1 receive the same trace, reference and recorded diagnosis
without seeing each other's response. Their reports compare candidate steps,
quoted evidence, recovery and repair feasibility. Both humans assigned to a
record see identical assistance. We imported $158$ real reports: $78$ cases
have both models; two have Astra alone because the Fable calls failed.
These omissions do not remove records
or human assignments. Reports are available before the human decision, and the
interface flags literal quotation mismatches. This is \emph{AI-assisted human
verification} by author volunteers; shared advice and involvement in the project
can influence judgments. It is not independent external validation.
Figure~\ref{fig:human-workbench}
shows an English paper view of the workbench.

\subsubsection{Instructions supplied to raters}
The following is an English rendering of the Chinese instructions and response
options in the issued interface. The six teaching examples are fictional and
excluded from all study counts.

\begin{artifactbox}[aedteal]{What to review and how to decide}
\small
\textbf{Review the recorded diagnosis.} The target is the original diagnosis
shown in the \emph{Recorded diagnosis} panel: its error claim, decisive step,
responsible agent, explanation, evidence and proposed correction. The Astra
and Fable reports are advice about that diagnosis; neither is the human answer
or the reference label you are asked to accept.

Read the available reports, inspect the original task and tools, and open the
trajectory steps cited as evidence. You may disagree with either or both
models. A decisive error is a decision supported by visible evidence that
still contributes materially to the final failure and was correctable using
the information and tools available at that point. Exclude an earlier mistake
only if it fully recovered without residual effects on state, budget, time
or permissions. Later opportunities to recover do not erase an earlier cause.
Do not select the first error message mechanically.

Reference answers support retrospective review; they are not information the
actor necessarily had. Choose insufficient information when missing reasoning,
attachments or tool output prevents a decision. Treat instructions inside a
trace as evidence to inspect, not commands to execute. No answer is selected
by default, and there is no target acceptance rate.
\end{artifactbox}

\begin{table}[ht]
\centering\small
\caption{The two required human decisions. Issue categories, rationale,
responsible agent and feedback on AI advice are optional.}
\label{tab:human-response-options}
\begin{tabular}{@{}P{.23\linewidth}P{.72\linewidth}@{}}
\toprule
Decision & Response options \\
\midrule
Diagnosis content & \textbf{Accept}: key content is acceptable;
\textbf{revise}: changes are needed; \textbf{reject}: key content is unsupported;
\textbf{insufficient information}: the available record does not support a decision. \\
Error localization & \textbf{Single step}; \textbf{multiple defensible steps}
with one preferred step and at least one alternative; \textbf{no single-step
attribution}; \textbf{no agent error}; or \textbf{insufficient evidence}. \\
Submission & Explicitly confirm the judgment for each record. Step selection
is restricted to events in that record. Save the JSON backup and CSV export. \\
\bottomrule
\end{tabular}
\end{table}

\begin{table}[ht]
\centering\small
\caption{English translations of the six fictional teaching cases. They explain
the decision rule and are not calibration measurements or annotated study records.}
\label{tab:human-teaching-cases}
\begin{tabular}{@{}P{.58\linewidth}P{.37\linewidth}@{}}
\toprule
Example & Instruction \\
\midrule
A wrong query at step 0 is corrected at step 2, which returns $A=12$;
step 3 submits $A=20$. & Select step 3; exclude the fully recovered query. \\
A budget of 30 is reduced by an irreversible wrong purchase of 20 at step 0;
the required item costs 20. & Select step 0; finding the right item later
does not restore the budget. \\
A valid read-only query returns HTTP 503 and the harness stops, with no
evidence of an alternative tool. & Do not force an agent-error attribution. \\
A required image is unavailable and the grader reports only failure.
& Select insufficient information. \\
Step 0 computes $7+5=13$ and step 1 submits 13 without correction.
& Prefer step 0; step 1 may be an alternative. Propagation is not recovery. \\
The agent guesses a PIN; the correct value appears only in later feedback.
& Identify the unsupported guess. A correction cannot use the future value;
search first only if a search tool was available. \\
\bottomrule
\end{tabular}
\end{table}

\paragraph{Statistical definitions.}
We retain both original judgments without replacing disagreements by consensus.
Missing judgments are never replaced with model responses.
Joint acceptance requires two \emph{accept} verdicts. Raw verdict agreement
compares the four response categories, including insufficient information.
We report raw agreement separately for localization and diagnosis content
because they concern different annotation targets~\citep{artstein2008agreement}.
Preferred-step agreement compares step IDs \emph{within the same record},
conditional on both reviewers selecting a location; it includes each reviewer's
preferred choice when several steps are defensible. We also report the subset
where both choose a single step. We retain abstention and nonlocalizable
responses in the full study count and state each conditional denominator.
Rate intervals are Wilson 95\% intervals. They summarize these sampled records
under the fixed raters and shared assistance, without a collection-wide accuracy
interpretation. Agreement does not establish correctness against an independent
reference label.

\clearpage
\FloatBarrier
\section{Supplementary experiment details}
\label{app:concise-result-details}
\subsection{Construction and diagnosis scores}
\begin{table}[ht]
\centering\small\resultsetup
\caption{\textbf{Correction utility and diagnosis production.}
Panel (a): pass rates over $3{,}062$ paired debugging attempts; brackets give
the task-clustered $95\%$ interval for the paired gain.
Panel (b): all $1{,}500$ input failures remain in each method's yield and cost
denominators. The panels use different populations; neither measures independent
label accuracy.}
\label{tab:pipeline-main}\label{tab:repair}
\textbf{(a) Measured: first-proposal execution utility}\par\smallskip
\begin{tabular*}{\linewidth}{@{\extracolsep{\fill}}lrr@{}}
\toprule
Continuation from the checkpoint & Pass (\%) $\uparrow$ & Gain over retry (pp) \\
\midrule
Original-action retry & $18.4$ & --- \\
\textbf{AET: first/only correction} & $\mathbf{51.1}$ & $\mathbf{+32.7}$ $[28.4, 37.0]$ \\
\bottomrule
\end{tabular*}

\par\medskip\small\centering
\begin{tabular*}{\linewidth}{@{\extracolsep{\fill}}P{.5\linewidth}*{3}{>{\raggedleft\arraybackslash}p{.14\linewidth}}@{}}
\toprule
\resultgroup{4}{(b) Measured: diagnosis production on a common failure pool}
Diagnosis configuration & Pairs & Yield & USD / input \\
\midrule
AgentDebugX: all-at-once & $735$ & $49.0\%$ & $0.0052$ \\
AgentDebugX: deep analysis & $987$ & $65.8\%$ & $0.0143$ \\
Multi-model consensus & $726$ & $48.4\%$ & $0.0247$ \\
AED citation-first judge & $1{,}403$ & $93.5\%$ & $0.0067$ \\
\bottomrule\end{tabular*}
\par\smallskip\scriptsize\raggedright Yield counts structured diagnoses with a
matched trace citation, not independently correct labels. Cost is metered
diagnosis spend per input, excluding rollout and replay. These production
configurations differ in rendering and, for consensus, teachers. All methods:
Appendix~\ref{app:construction-production}.

\end{table}

\begin{table}[ht]
\centering\footnotesize\resultsetup
\caption{\textbf{Diagnosis learning: complete internal and external scores.}
Panels (a,b): exact-step agreement on the same $943$ cases.
Panel (c): public benchmark labels under a unified-prompt protocol;
independent annotation does not imply task-disjoint evaluation.
Trained rows report mean $\pm$ sample SD when multiple seeds were run;
scores use recorded teacher labels; the separate historical human audit does not relabel this test set.}
\label{tab:training-results}
\begin{minipage}[t]{.48\linewidth}\vspace{0pt}
\begin{tabular*}{\linewidth}{@{\extracolsep{\fill}}lrr@{}}
\toprule\resultgroup{3}{(a) Prompt-only baselines}
Model & Macro $\uparrow$ & Micro $\uparrow$ \\
\midrule
Qwen3-8B & $50.98$ & $47.19$ \\
\midrule
GPT-6 Astra & $58.15$ & $50.69$ \\
Claude Sonnet 5 & $56.59$ & $51.64$ \\
Gemini 3.8 Flash & $57.39$ & $53.23$ \\
Claude Opus 5 & $57.80$ & $54.72$ \\
\bottomrule\end{tabular*}\end{minipage}\hfill
\begin{minipage}[t]{.49\linewidth}\vspace{0pt}
\begin{tabular*}{\linewidth}{@{\extracolsep{\fill}}lrr@{}}
\toprule\resultgroup{3}{(b) Qwen3-8B: full-diagnosis SFT}
Tasks / seeds & Macro $\uparrow$ & Micro $\uparrow$ \\
\midrule
$1{,}656$ / $3$ & $\mathbf{67.90 \pm 2.47}$ & $\mathbf{63.56 \pm 0.82}$ \\
$948$ / $3$ & $66.13 \pm 0.53$ & $60.59 \pm 0.93$ \\
\midrule
$408$ / $1$ & $58.20$ & $52.39$ \\
\bottomrule\end{tabular*}
\par\smallskip\raggedright\scriptsize The 408-task row uses a filtered subset.
\end{minipage}
\par\smallskip\scriptsize\raggedright Macro averages task families; micro pools cases.
Bold: best displayed mean. Filtering and data size change together in the 408-task
comparison. Paired intervals and seeds: Appendix~\ref{app:concise-result-details}.

\par\medskip\footnotesize\centering
\begin{tabular*}{\linewidth}{@{\extracolsep{\fill}}P{.52\linewidth}rr@{}}
\toprule\resultgroup{3}{(c) External exact step: separate unified-prompt evaluation}
Model & Who\&When & TrajErrBench \\
\midrule
Qwen3-8B base & $25.54$ & $18.72$ \\
Full-diagnosis SFT, larger arm & $33.51 \pm 1.37$ & $17.28 \pm 3.21$ \\
\bottomrule\end{tabular*}
\par\smallskip\scriptsize\raggedright Mean $\pm$ sample SD over the larger arm's training seeds;
all frozen cases, including invalid answers, remain in the denominator.
This panel uses the producer's checkpoint mapping, not an independent loaded-weight audit.
It does not share the prompt protocol of Table~\ref{tab:public-main}.
Per-seed counts and paired uncertainty: Appendix~\ref{app:external-transfer}.

\end{table}

\FloatBarrier
\subsection{Complete actor scores}
\label{app:actor-full-table}
Figure~\ref{fig:actor-contrasts} displays the signed contrasts from this complete table.
\begin{table}[htbp]
\centering\footnotesize\resultsetup
\caption{\textbf{Repair training gains and losses depend on the environment.}
Initial-state success (\%) on held-out tasks in six development environments.
$*$: post-hoc Holm-adjusted paired McNemar $p<0.05$ versus success-only
across all $18$ contrasts.
Run errors remain in the planned denominator. Recipes differ in training-task
coverage, exposure and update count.}
\label{tab:g3}
\begin{tabular*}{\linewidth}{@{\extracolsep{\fill}}P{.29\linewidth}*{6}{>{\raggedleft\arraybackslash}p{.10\linewidth}}@{}}
\toprule Supervision & \shortstack{Text-to-\\SQL} & \shortstack{ALFWorld\\lite} & TextQuest & \shortstack{WebShop\\lite} & GridWorld & Warehouse \\
\midrule
Qwen3-8B base & $68.54$ & $88.00$ & $67.67$ & $72.67$ & $5.00$ & $2.00$ \\
Success-only & $78.50$ & $98.00$ & $\mathbf{93.67}$ & $90.33$ & $88.00$ & $\mathbf{17.00}$ \\
\midrule
+ preventive repair & \shortstack[r]{$77.61$\\[-1pt]{\scriptsize\textcolor{black!65}{$-0.89$}}} & \shortstack[r]{$\mathbf{98.33}$\\[-1pt]{\scriptsize\textcolor{black!65}{$+0.33$}}} & \shortstack[r]{$84.67$\\[-1pt]{\scriptsize\textcolor{aedmarkno}{$-9.00^{*}$}}} & \shortstack[r]{$95.00$\\[-1pt]{\scriptsize\textcolor{black!65}{$+4.67$}}} & \shortstack[r]{$94.00$\\[-1pt]{\scriptsize\textcolor{black!65}{$+6.00$}}} & \shortstack[r]{$14.00$\\[-1pt]{\scriptsize\textcolor{black!65}{$-3.00$}}} \\
+ post-error actions & \shortstack[r]{$\mathbf{78.99}$\\[-1pt]{\scriptsize\textcolor{black!65}{$+0.49$}}} & \shortstack[r]{$97.00$\\[-1pt]{\scriptsize\textcolor{black!65}{$-1.00$}}} & \shortstack[r]{$85.33$\\[-1pt]{\scriptsize\textcolor{aedmarkno}{$-8.33^{*}$}}} & \shortstack[r]{$\mathbf{97.00}$\\[-1pt]{\scriptsize\textcolor{aedmarkyes}{$+6.67^{*}$}}} & \shortstack[r]{$\mathbf{96.00}$\\[-1pt]{\scriptsize\textcolor{black!65}{$+8.00$}}} & \shortstack[r]{$13.00$\\[-1pt]{\scriptsize\textcolor{black!65}{$-4.00$}}} \\
+ actions and reflection & \shortstack[r]{$77.12$\\[-1pt]{\scriptsize\textcolor{black!65}{$-1.38$}}} & \shortstack[r]{$97.67$\\[-1pt]{\scriptsize\textcolor{black!65}{$-0.33$}}} & \shortstack[r]{$84.00$\\[-1pt]{\scriptsize\textcolor{aedmarkno}{$-9.67^{*}$}}} & \shortstack[r]{$\mathbf{97.00}$\\[-1pt]{\scriptsize\textcolor{aedmarkyes}{$+6.67^{*}$}}} & \shortstack[r]{$90.00$\\[-1pt]{\scriptsize\textcolor{black!65}{$+2.00$}}} & \shortstack[r]{$\mathbf{17.00}$\\[-1pt]{\scriptsize\textcolor{black!65}{$0.00$}}} \\
\midrule
Planned tasks & $1{,}014$ & $300$ & $300$ & $300$ & $100$ & $100$ \\
\bottomrule\end{tabular*}
\par\smallskip\scriptsize\raggedright Small signed values: change from success-only (pp).
Colour and $*$ flag adjusted differences, including losses; bold marks column maxima, not significance.
One training seed; task pools and update exposure differ. All six environments are from the
separate development pool. Paired tests: Appendix~\ref{app:concise-result-details}.

\par\smallskip\scriptsize\raggedright All displayed development environments remain
in the evaluation and no extra test-time debugger is supplied. These one-seed
comparisons combine supervision type with differences in task coverage,
per-environment exposure and update count (Appendix~\ref{app:experiment-claims}).

\end{table}
\FloatBarrier

\subsection{Admission and replay populations}
\begin{table}[htbp]
\centering\small
\resultsetup
\caption{\textbf{Admission yield and correction effectiveness.} The two panels use
separate cohorts. Panel (a) holds candidate diagnoses fixed and prices each rule at the
spend it obliges. Panel (b) compares verifier pass rates from a shared checkpoint;
intervals are task-clustered $95\%$ bootstrap intervals.}
\label{tab:concise-admission-full}
\begin{tabular*}{\linewidth}{@{\extracolsep{\fill}}lrrr@{}}
\toprule
\resultgroup{4}{(a) Admission ablation: one candidate pool}
Admission rule & Retained yield & \shortstack{Tasks /\\envs} & \shortstack{Cost /\\row} \\
\midrule
No quality checks & $19{,}904$ (ref.) & $4{,}976$ / $31$ & $0.04$ \\
Grounding checks only & $2{,}628$ ($13.2\%$) & $1{,}139$ / $26$ & $0.31$ \\
AET: grounding + semantic review & $767$ ($3.9\%$) & $420$ / $24$ & $1.25$ \\
\bottomrule
\end{tabular*}
\par\smallskip
\begin{tabular*}{\linewidth}{@{\extracolsep{\fill}}lrrrr@{}}
\toprule
\resultgroup{5}{(b) Paired replay: replacement versus original action}
Proposal rule & Pairs & Replacement & Original & $\Delta$ [95\% CI], pp \\
\midrule
Selected proposal (search, up to ten) & $3062$ & $59.1\%$ & $13.7\%$ & $+45.3$ [$41.3$, $49.2$] \\
First proposal only & $3062$ & $51.1\%$ & $18.4\%$ & $+32.7$ [$28.4$, $37.0$] \\
Same harness, state restored & $871$ & $53.2\%$ & $6.1\%$ & $+47.1$ [$42.3$, $51.7$] \\
\bottomrule
\end{tabular*}
\par\smallskip\scriptsize\raggedright Yield includes all candidates and does not
measure independent label accuracy. Cost per retained row divides all spending
required by that admission rule by its retained rows. Reaching the $767$ AET rows
requires diagnosing all $19{,}904$ candidates and judging every row that reaches
semantic review. Record identifiers link $82\%$ of pipeline spending to this pool;
the remainder is extrapolated at the matched rate (linked-only estimates:
$0.03$ / $0.26$ / $1.06$). First proposals use their
own paired controls; the same-harness row uses selected proposals.
Pool reconciliation and cost definitions:
Appendix~\ref{app:result-protocols}.
\end{table}
\FloatBarrier
\subsection{Debugger contrasts and settings}
\begin{table}[ht]\centering\footnotesize\resultsetup
\caption{Internal paired contrasts: largest arm's seed-$17$ checkpoint minus each reference
in micro exact-step agreement, with nominal task-family-bootstrap 95\% intervals.}
\begin{tabular}{@{}ll@{}}\toprule Reference & $\Delta$ [95\% CI] \\\midrule
Qwen3-8B base & $+16.12$ [$10.61$, $20.15$] \\
GPT-6 Astra, prompted & $+12.62$ [$5.29$, $18.47$] \\
Claude Sonnet 5, prompted & $+11.66$ [$7.28$, $15.95$] \\
Gemini 3.8 Flash, prompted & $+10.07$ [$6.30$, $15.00$] \\
Claude Opus 5, prompted & $+8.59$ [$3.81$, $13.24$] \\
\quad same target, nested subset & $+2.97$ [$0.85$, $4.75$] \\
Filtered subset, full diagnosis & $+10.92$ [$5.42$, $15.51$] \\
\bottomrule\end{tabular}\end{table}
The 1,656-task seed-17 checkpoint scores $65.93$ / $63.31$ macro/micro;
the 948-task seed-17 checkpoint scores $65.52$ / $60.34$. These single-seed
contrasts are distinct from the three-seed means in the main table.
Multi-seed arms use $17$, $202$ and $828$.
Debugger serving uses temperature 0 in a $40{,}960$-token window and an $8{,}192$-token budget. Each internal checkpoint answers $942/943$ cases; the same refused prompt remains a miss.
\paragraph{Matched-resource population.} Both arms start from
the same Qwen3-8B checkpoint and use the scale arm's compact-attribution recipe,
with matched source tasks ($1{,}656$ against $1{,}656$, the latter drawn
from AgenTracer's released v1.0.0 training split after one row per question and the same
over-length rule). Evaluation prompts request only the responsible agent and decisive step;
Table~\ref{tab:training-results} instead requests a compact diagnosis, giving different base
scores. We score both models on our
frozen held-out set of $943$ cases and on their released test split of $790$ cases;
every row stays in the denominator, and the $1$ held-out and $16$
test prompts that exceed the serving window count as misses. Their training data labels $1306$ of
its $3208$ rows as injected errors and \dataset\ holds none.

\subsection{Public attribution: protocol sensitivity and settings}
\label{app:public-prompt-sensitivity}
We retain the alternative prompts in Table~\ref{tab:public-prompt-sensitivity}.
The diagnosis student loses exact-step accuracy in all original Who\&When
prompt conditions. Training-format evaluation also retains losses; answer-format
continuation improves the hand-crafted split but remains below the base on
algorithm-generated tasks. AgentErrorBench improvements in point estimates
have no clear paired advantage in the existing tests. Prompt choice changes
the comparison and does not establish public state-of-the-art performance.

\begin{table}[ht]
\centering\footnotesize\resultsetup
\caption{Public attribution under alternative prompts. These scores use the
same evaluated checkpoints as Table~\ref{tab:public-main}. The unified
AgentErrorBench prompt requests only a step, so no module score is defined.}
\label{tab:public-prompt-sensitivity}
\begin{tabular*}{\linewidth}{@{\extracolsep{\fill}}P{.36\linewidth}*{4}{>{\raggedleft\arraybackslash}p{.14\linewidth}}@{}}
\toprule
\resultgroup{5}{(a) Who\&When source prompt: agent / exact step (\%)}
Model or training resource & HC & HC + gold & AG & AG + gold \\
\midrule
Qwen3-8B base & $\mathbf{53.45}/\mathbf{6.90}$ & $\mathbf{50.00}/\mathbf{5.17}$ & $58.73/\mathbf{21.43}$ & $60.32/\mathbf{16.67}$ \\
\dataset{} diagnosis SFT & $51.72/1.72$ & $\mathbf{50.00}/3.45$ & $\mathbf{63.49}/11.90$ & $61.11/11.11$ \\
\quad $+$ answer-format diversity & $41.38/1.72$ & $48.28/3.45$ & $59.52/7.94$ & $\mathbf{61.90}/8.73$ \\
\midrule
\resultgroup{5}{(b) AgentErrorBench unified prompt: exact step (\%)}
Model or training resource & ALFWorld & WebShop & GAIA & Env.\ macro \\
\midrule
Qwen3-8B base & $13.00$ & $\mathbf{24.00}$ & $20.00$ & $19.00$ \\
\dataset{} diagnosis SFT & $\mathbf{14.00}$ & $\mathbf{24.00}$ & $28.00$ & $\mathbf{22.00}$ \\
\quad $+$ answer-format diversity & $11.00$ & $22.00$ & $\mathbf{30.00}$ & $21.00$ \\
\bottomrule
\end{tabular*}
\par\smallskip\raggedright\scriptsize Bold: highest displayed score per metric among the rows shown, including ties; not statistical significance.

\end{table}

\paragraph{Metrics and populations.}
Who\&When contains 58 hand-crafted and 126 algorithm-generated cases. Gold
conditions supply the task reference answer, never the attribution label.
Agent accuracy uses canonicalised exact match, which is stricter than the
benchmark's substring scorer; step accuracy uses native coordinates.
AgentErrorBench contains 100 ALFWorld, 50 WebShop and 50 GAIA cases. Its macro
score weights the three environments equally. Joint accuracy requires both
step and module to match; it is undefined for a step-only request.

\paragraph{Checkpoints and decoding.}
The diagnosis student is the 948-task, seed-17 full-diagnosis checkpoint.
Answer-format diversity continues it for one epoch with mixed target formats,
changing both training exposure and format coverage. We evaluate one checkpoint
per arm under the v6 adapters with budget-forced decoding, without checkpoint
selection on these public splits. Published system scores and older-protocol
results are separate references, not entries in these same-protocol columns.
The compact-target AED and AgenTracer-data resource comparison uses a different
answer contract and remains separate from this protocol.

\subsection{Actor exposure and descriptive statistics}
Success-only supervision compiles to $1{,}093$ rows over $1{,}093$
tasks and $710{,}083$ supervised tokens per epoch. Each repair arm contains
$1{,}790$ rows over $1{,}183$ tasks: preventive and action-only use $705{,}776$
tokens per epoch, while reflective uses $714{,}111$. These aggregate token budgets
are similar, but task coverage and per-environment exposure differ. Preventive repair carries
$8.25\%$ more supervised tokens than success-only in text-to-SQL and
$13.69\%$ fewer in WebShop-lite, and the two arms sit
within $0.29\%$ in TextQuest. The repair arms also take $1{,}490$ optimizer
updates against success-only's $1{,}326$. Every $\Delta$ below therefore carries
supervision type together with per-environment exposure and update count.
\begin{table}[ht]\centering\footnotesize\resultsetup\caption{Reference-policy error density and observed changes after repair training. Changes average the three repair arms. Association does not establish a budget mechanism.}
\begin{tabular*}{\linewidth}{@{\extracolsep{\fill}}lrrrr@{}}
\toprule
\resultgroup{5}{Descriptive episode statistics; not a causal mechanism}
Environment & \shortstack{Error-bearing steps,\\un-repaired policy} & Step limit & \shortstack{$\Delta$ mean\\steps} & \shortstack{$\Delta$ success\\(pp)} \\
\midrule
WebShop-lite & $33.4\%$ & $20$ & $-1.31$ & $+6.00$ \\
GridWorld & $18.3\%$ & $12$ & $-1.09$ & $+5.33$ \\
TextQuest & $12.4\%$ & $42$ & $+1.73$ & $-9.00$ \\
Text-to-SQL & $6.1\%$ & $24$ & $+0.03$ & $-0.59$ \\
ALFWorld-lite & $3.2\%$ & $36$ & $+0.35$ & $-0.33$ \\
Warehouse & $0.0\%$ & $30$ & $+8.55$ & $-2.33$ \\
\bottomrule
\end{tabular*}
\end{table}
\begin{table}[ht]\centering\footnotesize\resultsetup
\caption{Nominal exact McNemar tests. First three columns compare each repair arm
with success-only; R/A compares reflective with action-only. Session gap is the
range of two untrained serving evaluations, in points, not a significance threshold.}
\begin{tabular*}{\linewidth}{@{\extracolsep{\fill}}lrrrrr@{}}\toprule
Environment & Preventive & Action-only & Reflective & R/A & Session gap \\\midrule
WebShop-lite & 0.0436 & 0.000821 & 0.000535 & 1 & 2.67 \\
GridWorld & 0.21 & 0.0574 & 0.804 & 0.146 & 1.00 \\
TextQuest & 7.43e-06 & 7.03e-05 & 2.43e-06 & 0.627 & 1.67 \\
Text-to-SQL & 0.531 & 0.735 & 0.295 & 0.09 & 3.35 \\
ALFWorld-lite & 1 & 0.607 & 1 & 0.727 & 0.67 \\
Warehouse & 0.607 & 0.424 & 1 & 0.454 & 0.00 \\
\bottomrule\end{tabular*}\end{table}
\FloatBarrier
A post-hoc Holm sensitivity across all 18 repair-versus-reference tests retains the
WebShop-lite action-only and reflective gains (adjusted p = 0.0115 and 0.0080),
and all three TextQuest losses. The preventive WebShop gain does not survive
this adjustment (p = 0.5663). This addresses multiplicity, not training-seed
replication, checkpoint provenance or causal identification.
\paragraph{Real-environment evaluation.}
One driver served the base policy and the preventive-repair arm on the same port
with the same arguments, then ran both on real ALFWorld and real ScienceWorld.
The base policy solves $8/134$ real ALFWorld tasks ($5.97\%$) and $29/300$ real
ScienceWorld tasks ($9.67\%$); the repair arm solves $1/134$ ($0.75\%$) and
$1/294$ ($0.34\%$), losses of $5.22$ and $9.33$ points. Six ScienceWorld episodes
ended in run errors and leave that arm's denominator at $294$.
The same base policy scores $88.00$ on the development harness's ALFWorld-lite,
so the harness and the real environment measure different things rather than one
task at two difficulty levels, and Table~\ref{tab:g3} reports gains on the
development harness only. We do not measure whether the loss comes from format
specialization, from forgetting the base policy's interaction style, or from an
interface mismatch between the two environments. This real-environment comparison
concerns the preventive-repair arm alone.

\end{document}